\documentclass[dvipsnames,x11names,svgnames,HTML]{article}

\usepackage[pagenumbers]{cvpr} % To force page numbers, e.g. for an arXiv version

\usepackage[utf8]{inputenc} % allow utf-8 input
\usepackage[T1]{fontenc}    % use 8-bit T1 fonts
\usepackage{url}            % simple URL typesetting
\usepackage{booktabs}       % professional-quality tables
\usepackage{amsfonts}       % blackboard math symbols
\usepackage{nicefrac}       % compact symbols for 1/2, etc.
\usepackage[x11names, svgnames]{xcolor}
\usepackage[most]{tcolorbox}
\usepackage{xcolor}
\usepackage[most]{tcolorbox}
\usepackage[HTML]{xcolor} 
\usepackage{microtype}      % microtypography
\usepackage{xcolor}         % colors
\usepackage{xspace}
\usepackage{amsmath}        % math
\newtheorem{definition}{Definition}
\usepackage{array}
\usepackage{graphicx}
\usepackage{wrapfig}
\usepackage{subcaption}
\usepackage{longtable}
\usepackage{array}
\usepackage{algorithm}
\usepackage{algpseudocode}
\usepackage{multirow}
\usepackage{bibunits}
\usepackage{silence}

\newcommand{\xhdr}[1]{\vspace{1mm}\noindent{{\bf #1.}}}
\newcommand{\method}{\textbf{PGF}\xspace}

\newcommand{\circnum}[1]{\textcircled{\small #1}}

\definecolor{cvprblue}{rgb}{0.21,0.49,0.74}
\usepackage[pagebackref,breaklinks,colorlinks,allcolors=cvprblue]{hyperref}

\def\paperID{XXXXX} % *** Enter the Paper ID here
\def\confName{CVPR}
\def\confYear{2026}

\title{Understanding Task Transfer in Vision-Language Models}

\makeatletter
\renewcommand{\thefootnote}{\fnsymbol{footnote}}
\makeatother

\author{%
Bhuvan Sachdeva\thanks{Equal contribution. Corresponding Author: \texttt{vineeth.nb@microsoft.com}}
\quad
Karan Uppal\footnotemark[1]
\quad
Abhinav Java\footnotemark[1]
\quad
Vineeth N. Balasubramanian \\
Microsoft Research India \\
\small Project page: \url{https://aka.ms/task-transfer-vlms}
}

\tcbset{
    takeaways/.style={
        colback=cyan!5,          % light background
        colframe=cyan!60!black,  % lighter border (blue mixed with black 30%)
        boxrule=0.8pt,
        arc=4pt,
        left=4pt,
        right=4pt,
        top=2pt,
        bottom=2pt,
        fonttitle=\bfseries,
        coltitle=black
    }
}

\begin{document}

\newcommand{\maketitlewithvisual}[1]{
\twocolumn[{%
\renewcommand\twocolumn[1][]{##1}%
\vspace{-50pt}
\begin{center}
\maketitle
% \vspace{-6mm}
\includegraphics[width=0.99\linewidth]{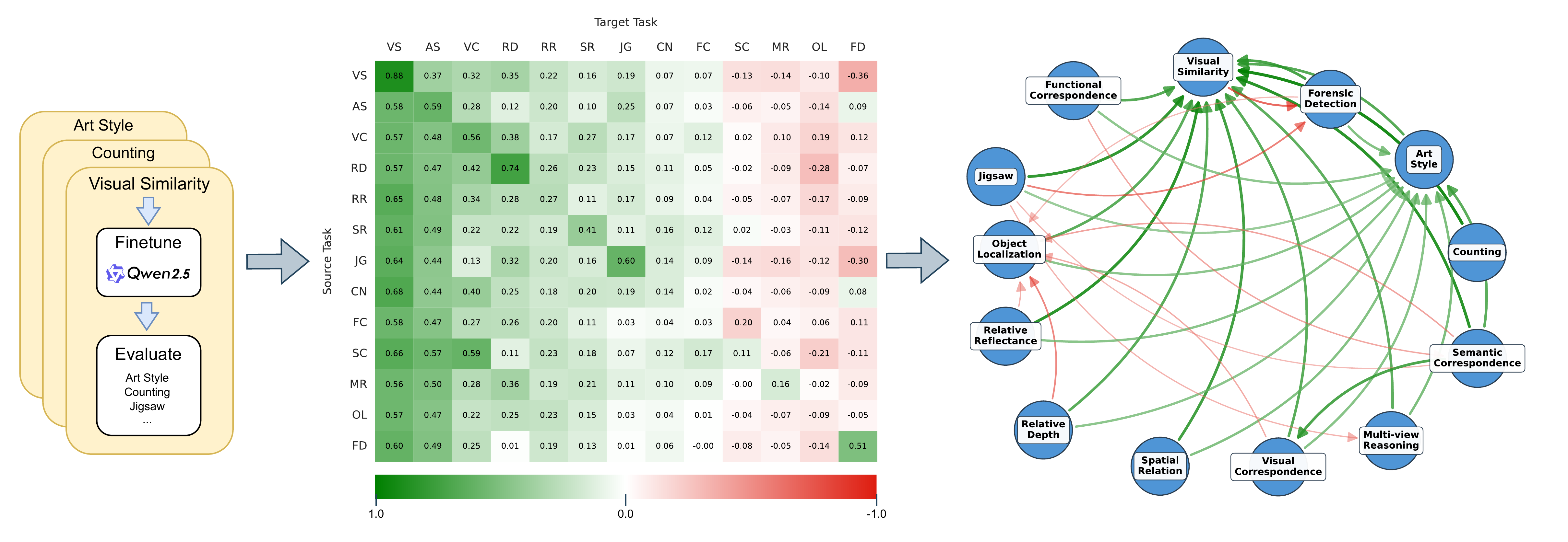}
% \vspace{-8pt}
\captionsetup[figure]{hypcap=false}
\captionof{figure}{One finetune, many fates: Finetuning Qwen-2.5-VL 32B on perception tasks creates a structured map of transfer capabilities. (The list of perception tasks considered can be found in Table~\ref{tab:task_classification}.)}
\label{fig:teaser}
\end{center}
}]
}

\maketitlewithvisual

\begin{abstract}

Vision-Language Models (VLMs) perform well on multimodal benchmarks but lag behind humans and specialized models on visual perception tasks like depth estimation or object counting. Finetuning on one task can unpredictably affect performance on others, making task-specific finetuning challenging. In this paper, we address this challenge through a systematic study of task transferability. We examine how finetuning a VLM on one perception task affects its zero-shot performance on others. We introduce Perfection Gap Factor (PGF), a normalized metric that measures change in performance as a result of task transfer. We utilize PGF to compute Task Transferability, which captures both the breadth and the magnitude of transfer induced by a source task. Using three open-weight VLMs evaluated across 13 perception tasks, we construct a task transfer graph that reveals previously unobserved relationships among perception tasks. Our analysis uncovers patterns of positive and negative transfer, identifies groups of tasks that mutually influence each other, organizes tasks into personas based on their transfer behavior and demonstrates how PGF can guide data selection for more efficient training. These findings highlight both opportunities for positive transfer and risks of negative interference, offering actionable guidance for advancing VLMs.

% OLD
% Vision-Language Models (VLMs) perform well on multimodal benchmarks but lag behind humans and specialized models on visual perception tasks like depth estimation or object counting. Finetuning on one task can unpredictably affect performance on others, making task-specific finetuning challenging. In this paper, we address this challenge through a systematic study of task transferability. We examine how finetuning a VLM on one perception task affects its zero-shot performance on others. To quantify these effects, we introduce Perfection Gap Factor (PGF), a metric that captures both the breadth and magnitude of transfer. Using three open-weight VLMs evaluated across 13 perception tasks, we construct a task-transfer graph that reveals previously unobserved relationships among perception tasks. Our analysis uncovers patterns of positive and negative transfer, identifies groups of tasks that mutually influence each other, organizes tasks into personas based on their transfer behavior and demonstrates how PGF can guide data selection for more efficient training. These findings highlight both opportunities for positive transfer \& risks of negative interference, offering actionable guidance for advancing VLMs.

\end{abstract}    
\section{Introduction}

% \begin{figure*}[!t]
%     \centering
%     \begin{subfigure}[b]{0.48\textwidth}
%         \centering
%         \includegraphics[width=\textwidth]{figures/task_graph_32B.pdf}
%         \caption{Qwen-2.5-VL 32B}
%         \label{fig:32Btaskgraph}
%     \end{subfigure}
%     \hfill
%     \begin{subfigure}[b]{0.48\textwidth}
%         \centering
%         \includegraphics[width=\textwidth]{figures/transferability_qwen-2.5-72b.pdf}
%         \caption{Qwen-2.5-VL 32B}
%         \label{fig:32Bheatmap}
%     \end{subfigure}
%     \caption{One finetune, many fates: Finetuning Qwen-2.5-VL (32B) on a single perception task creates a structured map of transfer (PGF). Node color shows donor/pirate role; edge width = PGF magnitude; right: representative task pairs and before/after zero-shot answers.}
%     \label{fig:teaser}
% \end{figure*}

% Para 1: VLMs are powerful, but we don't know how perception tasks transfer after finetuning -- a critical gap for robust multimodal learning
Vision Language Models~\citep{llava1.5, llava_improved2024, li2024llavaonevisioneasyvisualtask, gpt4, li2024llavanextinterleavetacklingmultiimagevideo, qwen2vl} have demonstrated significant progress in understanding visual information in recent years, as reflected in their performance across well-known benchmarks such as MMMU~\citep{yue2024mmmu}, DocVQA~\citep{mathew2021docvqadatasetvqadocument}, InfoVQA~\citep{mathew2021infographicvqa}, and TextVQA~\citep{singh2019vqamodelsread}. Visual instruction tuning~\citep{llava1.5} has helped adapt Large Language Models (LLMs) to parse visual input by finetuning a small number of parameters to align a visual encoder (e.g., CLIP~\citep{radford2021learning}) with a given LLM backbone. Despite this progress, careful analysis has shown that VLMs fall short in many visual understanding tasks, most often due to their limitations in visual perception~\cite{fu2024blink, tong2024eyes}. Understanding the limits of VLMs on visual perception tasks, especially ones that are natural to humans and serve as building blocks that scaffold on to more complex visual tasks remains an urgent need, in order to provide foundational solutions to robust visual processing. 

\begin{figure*}[!t]
    \centering
    \includegraphics[width=\linewidth]{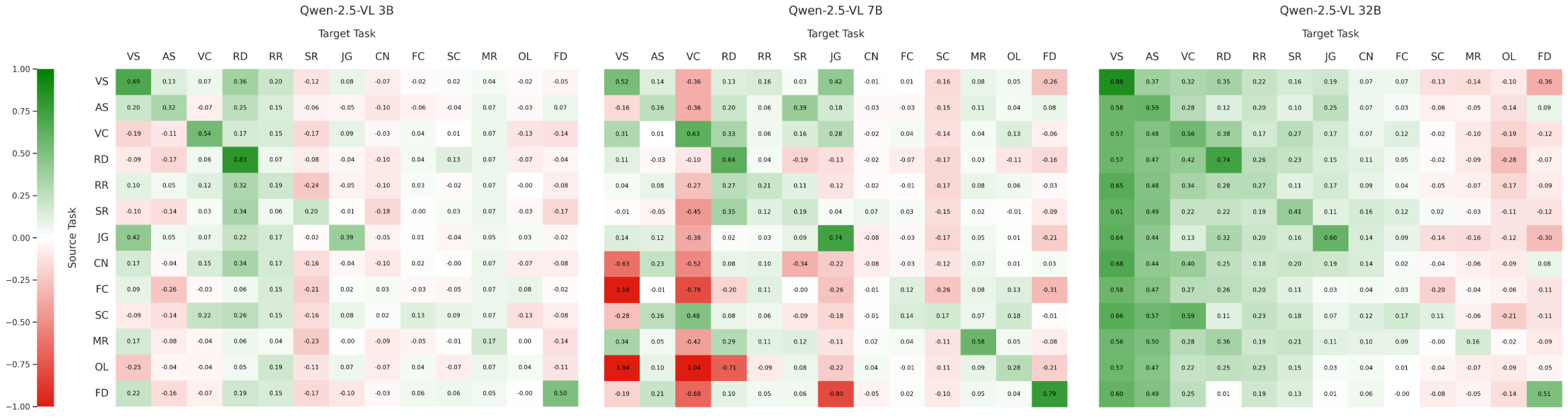}
    \caption{PGF Heatmaps for Qwen-2.5-VL model family (3B, 7B, 32B).}
    \label{fig:pgf}
\end{figure*}

% \footnotetext{*Equal contribution. Corresponding Author: \texttt{vineeth.nb@microsoft.com}}

\footnotetext[1]{Equal contribution. Corresponding Author: \texttt{vineeth.nb@microsoft.com}}

\makeatletter
\renewcommand{\thefootnote}{\arabic{footnote}}
\setcounter{footnote}{0}
\makeatother

VLMs lag behind humans and specialist models on basic perception tasks such as depth estimation, object detection, and counting. For example, on the BLINK~\citep{fu2024blink} leaderboard, the top performing models (GPT-4o at 60.04\% and GPT-4V at 51.14\%) achieve modest average performance compared to humans (95\%). This has motivated practitioners to finetune VLMs using parameter-efficient methods such as LoRA~\citep{hu2022lora} on curated, task-specific datasets to improve performance on perception tasks. 

However, little is known about how such finetuning affects a model’s other perception capabilities, particularly in modern foundation models used predominantly in zero-shot settings. Understanding this phenomenon is essential for both improving robustness and characterizing how VLMs adapt and generalize. In this work, we address this gap by investigating the following question, which to our knowledge has not been previously studied:
\begin{quote}
    \textbf{How does finetuning a VLM on one perception task affect its zero-shot performance on other perception tasks?}
    % \textbf{How does finetuning a VLM on one perception task transfer to its zero-shot performance on other perception tasks?}
\end{quote} 

% Para 3: Positioning
Prior work on task relationships in vision has largely focused on transfer learning involving finetuning on both source and target tasks~\citep{zamir2018taskonomy, chen2024learning}. Other efforts have explored pretraining strategies and their downstream effects~\citep{shariatnia2022well, tiong2024we}. In contrast, we study zero-shot cross-task transfer in VLMs: how a single-task finetuning intervention reshapes performance across a diverse set of perception tasks.

% Recent model-level efforts such as LLaVA-OneVision~\citep{li2024llavaonevisioneasyvisualtask} have enhanced transfer across modalities (e.g., video and images), while ~\cite{ivison2022data} explored adaptive data mixtures for text models to improve finetuning efficiency.

% Para 4: Short Summary of our key findings, metric
To quantify these effects, we introduce the notion of \textit{transferability} and \textit{malleability}, where transferability captures the effect a source task induces on other tasks through finetuning, and malleability captures how susceptible a target task is to being affected by finetuning on other tasks. These metrics encompass both the \textbf{breadth} (how many tasks are influenced) and the \textbf{magnitude} (the average strength of that influence).
The metrics, \textit{transferability} and \textit{malleability}, are computed using \textit{Perfection Gap Factor}, which is a normalized measure of the extent to which finetuning on a source task improves or degrades zero-shot performance on a target task.
Our analysis reveals several properties of tasks, including scale-dependent sharpening of transfer, distinct behaviors across perception granularity levels, and natural clustering among mutually beneficial tasks. We further extend our study to video models and observe similar trends across model sizes. Finally, we demonstrate that PGF can guide principled selection of training data subsets to improve finetuning efficiency and reduce performance regressions.

\xhdr{Key Contributions} Our contributions are as follows: 
\begin{itemize}
\item \textbf{Systematic Study}: We are the first to analyze how finetuning on one visual perception task affects zero-shot performance on a broad suite of other perception tasks.
% \item \textbf{Perfection Gap Factor}: PGF measures both the magnitude and breadth of transfer effects while normalizing for heterogeneous task difficulties and model baselines.
\item \textbf{Perfection Gap Factor}: PGF provides a transfer measure that normalizes for heterogeneous task difficulties and model baselines.
\item \textbf{Task Properties}: We uncover consistent structural properties of transfer, including scale-dependent sharpening, task type dependent transfer (granularity and perceptual), and the emergence of mutually beneficial task clusters. 
\item \textbf{Evaluation beyond images}: We evaluate cross-task transfer in multimodal models trained on video, showing that our key findings generalize to the temporal domain. 
\item \textbf{Downstream use of PGF}: We demonstrate that PGF can be used to identify beneficial source tasks and construct data subsets that improve finetuning efficiency while mitigating negative transfer.
\end{itemize}

%     \item We formally define task transferability in the context of modern Multimodal Language Models and introduce a robust metric, the Perfection Gap Factor (PGF), to measure cross-task transfer. Using PGF, we conduct the first systematic study of perception task transfer, mapping transferability across 13 tasks and 3 model scales.
%     \item We identify distinct transfer personas; namely donors, pirates, sponges, and sieves. This taxonomy provides actionable insights into which tasks are most beneficial for finetuning and which may hinder transfer.
%     \item We extend the notion of transferability to groups of tasks, analyzing semantic and hierarchical relationships. This group-level perspective offers practical guidance on how task families interact under finetuning.
%     \item We demonstrate that PGF can be predicted using interpretable classifiers, revealing which task-level features drive transferability. This opens up new directions for building models that can proactively estimate transfer potential.
% 2. Related Work
%     * Benchmarks for visual perception in VLMs (BLINK, plus others like MMVP, SEED-Bench).
%     * Transfer studies in NLP/CV — mostly narrow task pairs, not broad perception maps
%         - Task transfer & Taskonomy (CVPR’18) — precedent and analogy.
%     * LoRA finetuning & adaptation in VLMs.

% Gap to highlight across the above subsections: no systematic transfer map across perception tasks for VLMs.

\section{Related Work}
\label{sec:related}

% In this section, we present a detailed description of adjacent literature.

\xhdr{Benchmarks for Visual Perception} Several benchmarks~\citep{lin2015microsoftcococommonobjects, gupta2019lvisdatasetlargevocabulary, mathew2021docvqadatasetvqadocument, mathew2021infographicvqa} have been introduced to evaluate the progress of VLMs on visual understanding. Notably, MMMU~\citep{yue2024mmmu} is a large-scale benchmark assessing model capabilities across 30 subjects spanning technology, engineering, art, medicine, and more. Despite its broad coverage, MMMU largely focuses on examination-style question answering and lacks core perception tasks. Other benchmarks, such as DocVQA~\citep{mathew2021docvqadatasetvqadocument} and InfoGraphicsVQA~\citep{mathew2021infographicvqa}, combine OCR capabilities with visual understanding, while ChartQA~\citep{masry2022chartqabenchmarkquestionanswering} evaluates a model’s ability to parse complex charts and draw meaningful inferences. Visual Commonsense Reasoning~\citep{zellers2019vcr} presents a challenging task in which models must reason about the intention or consequences of actions depicted in a scene. MathVista~\citep{lu2023mathvista} is a multiple-choice mathematics dataset that requires interpreting figures or diagrams to answer questions, and NLVR~\citep{wu2024surprisingfailuremultimodalllms} tests a broad range of linguistic phenomena through image captioning tasks. While each of these datasets highlights specific aspects of multimodal understanding, most do not explicitly measure the visual perception capabilities that humans perform naturally. BLINK~\citep{fu2024blink}, in contrast, aggregates over 14 datasets spanning diverse tasks and serves as a central benchmark for our experiments.

\xhdr{Transferability Studies} The Taskonomy framework~\citep{zamir2018taskonomy} first introduced a framework for modeling the structure of computer vision tasks through transfer learning. ~\citet{zamir2018taskonomy} pretrain an encoder on a source task and then perform transfer learning on a task-specific decoder to estimate transferability between tasks. However, this study is restricted to the pre foundation model era and is primarily conducted using CNN-based vision encoders and small decoders. \cite{bao2022informationtheoreticapproachtransferabilitytask} introduce a new metric to estimate the performance of transfer-learned representations from source to target task. Unlike prior works that investigate transfer-learning, we focus our study on understanding the \textit{zero-shot task transfer} in VLMs providing novel ways to quantify it and provide actionable \textit{finetuning insights}. \citet{shariatnia2022well} compare various pretraining techniques by evaluating their zero-shot capabilities. In contrast, our work focuses on task finetuning, rather than pretraining strategies, to study how perception tasks transfer in modern foundation models. Closer to our work, ~\cite{tiong2024we} conduct experiments with VLMs from an evaluation perspective. Though similar, ~\cite{tiong2024we}'s study encompasses tasks like OCR, VQA, Captioning, Visual Reasoning, etc, on transfer learning providing insights about common biases such as length of output. \citep{chen2024learning} study the impact of various factors like dataset size, pretraining strategy on transfer in vision language models. In Natural Language Processing, \cite{huan2025does} examine how finetuning models on mathematical reasoning tasks affects their performance on both general reasoning and non-reasoning tasks. They also introduce a task transferability index, that is the accuracy gain relative to baseline scores. We present the details of our metric in Section~\ref{sec:method}, which is different from standard accuracy gain based metrics that do not account for variance in task difficulty.   

\xhdr{Finetuning VLMs} Adapting VLMs to unseen domains and tasks is an active area of research. A variety of strategies have been proposed to improve the efficiency and effectiveness of finetuning, particularly for large-scale models where full-parameter updates are computationally expensive. Parameter-efficient finetuning methods such as LoRA~\citep{hu2022lora} and related techniques~\citep{peft} have been widely adopted, allowing small subsets of model parameters to be updated while keeping the majority of the network frozen. These approaches have been shown to maintain or even improve downstream performance across a range of tasks, making them especially attractive for finetuning. QLoRA~\citep{dettmers2023qlora} backpropagates gradients through a frozen, 4-bit quantized model into Low Rank adapters, making the finetuning more efficient while preserving 16-bit performance.  Despite these advancements, systematic studies on how task-specific finetuning impacts the transferability of perception capabilities in VLMs remain limited, motivating our work.
% 3. Problem Formulation and Metric
%     * Problem setup: Finetune on one source task, evaluate on all others.
%     * Metric: Perfection Gap Factor (PGF)
%         - Intuition (how much of the remaining gap is closed or opened); bounded, interpretable across tasks with different ceilings
%         - Formal definition; equation, definition + illustrative toy example.
%         - Behavior and boundary cases (positive vs negative PGF)
%         - Why not just delta accuracy? Problems of ceiling effects and comparability.
%         (Keep long derivations/PGF boundary math in Appendix, we can keep the main explanation intuitive)
%     * Training setup: LoRA, data splits, compute.
%         - Point to Appendix for full hyperparams and BLINK task reconstruction details.

\section{Problem Formulation and Metric}
\label{sec:method}

We present our framework for characterizing the behavior of Vision-Language Models (VLMs) across diverse perception tasks. Our goal is to understand how finetuning on one task influences performance on others. We begin by discussing preliminaries like notation and problem setup, followed by quantifying task transferability using our proposed metric \textbf{Perfection Gap Factor (\method)}, which provides a robust way to account for differences in task difficulty and performance ceilings.

\xhdr{Preliminaries} We consider the setting where a VLM $\mathcal{M}$ is finetuned on a source task $T_\text{S}$ using a source dataset $\mathcal{D}_{\text{S}}^{\text{ train}}$ and subsequently evaluated on a set of $N$ target tasks $\{T_{j}\}_{j=1}^{N}$ using target datasets $\{\mathcal{D}_{j}^{\text{ eval}}\}_{j=1}^{N}$. We represent a VLM $\mathcal{M}$ finetuned on a dataset $\mathcal{D}_i$ for task $T_i$ as $\mathcal{M}(T_{i})$. The central question we study is on how finetuning on a task $T_{S}$ affects zero-shot performance on tasks $\{T_{j}\}_{j=1}^{N}$, and how one can quantify such inter-task relationships. We begin by formally defining task transferability.

\begin{definition}[Task Transferability]
% Let $\mu_{i \to j}$ denote a transferability score from source task $T_i$ to target task $T_j$. 
Let $\mu_{i \to j}$ denote a metric that captures change in performance on a target task $T_j$ as a result of finetuning on source task $T_i$.
Define $p = |\{j : \mu_{i \to j} > 0 \}|$ as the number of positive scores, and 
$n = |\{j : \mu_{i \to j} < 0 \}|$ as the number of negative scores. 
The \textbf{positive} and \textbf{negative} task transferability of $T_i$ to a set of target tasks $\{T_j\}_{j=1}^N$ are given by:
\begin{small}
\begin{equation}
\label{eq:transferability}
\begin{aligned}
\Delta(i)^{+} &= 
\Bigl(\tfrac{1 - e^{-\tfrac{p}{N}}}{p}\Bigr)
\sum_{j=1}^{N} \mu_{i \to j}\,\mathbf{1}_{\{\mu_{i \to j} > 0\}}, \\
\Delta(i)^{-} &= 
\Bigl(\tfrac{1 - e^{-\tfrac{n}{N}}}{n}\Bigr)
\sum_{j=1}^{N} \mu_{i \to j}\,\mathbf{1}_{\{\mu_{i \to j} < 0\}}.
\end{aligned}
\end{equation}
\end{small}
where $\Delta(i)^{+}$ and $\Delta(i)^{-}$ denote positive and negative task transferability, respectively.
\end{definition}

% Intuition for transferability
% Our definition of task transferability is not tied to any specific transferability score $\mu$ and can be applied to any choice of scoring function. 
$\Delta(i)^{\pm}$ captures both the \textit{magnitude} and the \textit{breadth} of influence of a source task $T_i$. The summation term helps measure the average performance gain or degradation across the affected tasks. The exponential weighting adjusts for the number of tasks, penalizing cases where positive or negative effects occur only on a small fraction of tasks. In other words, a task with large but isolated transfer effects will be scored lower than one that provides consistent improvements across many targets. However, this metric only describes the behavior of a source task on other tasks.

To characterize how sensitive a target task is to finetuning on other tasks, we introduce the notion of malleability. A target task can be considered highly malleable if finetuning on different source tasks leads to significant change (positive or negative) in the performance on that task.  To quantify this value, we aggregate the positive and negative PGF scores induced on that task by other source tasks.

\begin{definition}[Malleability]
Let $\mu_{i \to j}$ denote a metric that captures change in performance on a target task $T_j$ as a result of finetuning on source task $T_i$. 
Define $p = |\{i : \mu_{i \to j} > 0 \}|$ as the number of positive scores, and 
$n = |\{i : \mu_{i \to j} < 0 \}|$ as the number of negative scores. 
The \textbf{positive} and \textbf{negative} malleability of  $T_j$ to a set of source tasks $\{T_i\}_{i=1}^N$ are given by:
\begin{small}
\begin{equation}
\label{eq:malleability}
\begin{aligned}
\Theta(j)^{+} &= 
\Bigl(\tfrac{1 - e^{-\tfrac{p}{N}}}{p}\Bigr)
\sum_{i=1}^{N} \mu_{i \to j}\,\mathbf{1}_{\{\mu_{i \to j} > 0\}}, \\
\Theta(j)^{-} &= 
\Bigl(\tfrac{1 - e^{-\tfrac{n}{N}}}{n}\Bigr)
\sum_{i=1}^{N} \mu_{i \to j}\,\mathbf{1}_{\{\mu_{i \to j} < 0\}}.
\end{aligned}
\end{equation}
\end{small}
where $\Theta(j)^{+}$ and $\Theta(j)^{-}$ denote positive and negative malleability, respectively.
\end{definition}

Note: These definitions are agnostic to the choice of $\mu_{i \to j}$. For our analysis, we use \textit{Perfection Gap Factor} (defined below) as the default choice, and include transferability analysis using relative gain in the supplementary.

\xhdr{Perfection Gap Factor}  A central challenge in quantifying task transferability and malleability is designing a metric that is comparable across tasks with very different difficulty levels and performance ceilings. Reporting accuracy gains after finetuning can be misleading. For example, a +2\% improvement on a task where the model is already near the ceiling is much more significant than the same +2\% improvement on a task where the model starts very low. Conversely, small drops in accuracy near the floor may not reflect meaningful transfer failure. To address this, we introduce the \textit{Perfection Gap Factor} (PGF), which measures how much of the remaining gap to the ceiling is closed (or opened) by finetuning on a source task. Mathematically, we define the \textbf{PGF} between a source task $T_{i}$ and a target task $T_{j}$ as the \textit{ratio of performance gain to the performance gap}, i.e.,
\begin{equation}
    \mu_{i\to j} = \frac{\mathrm{Acc}\!\left(\mathcal{M}(T_i),\, T_j\right) - \mathrm{Acc}\!\left(\mathcal{M},\,T_j\right)}{U_{j} - \mathrm{Acc}\!\left(\mathcal{M},\,T_j\right) + \epsilon}
\end{equation}
where $\mathrm{Acc}(\mathcal{M},T_j)$ represents the accuracy of model $\mathcal{M}$ on task $T_j$, $U_j$ is the upper-bound (ceiling) performance of the target task and $\epsilon=10^{-6}$ is added to the denominator for numerical stability. By normalizing the gain relative to the remaining gap, PGF becomes both bounded and interpretable, making it comparable across tasks. Values above zero indicate positive transfer, while negative values indicate degradation. Intuitively, PGF incorporates the following questions:

\begin{enumerate}[label=\circnum{\arabic*}]%, leftmargin=*]
    \item \textit{How much does finetuning on a source task improve the target task?}
    \item \textit{What is the model's zero-shot performance on the target task before finetuning?}
    \item \textit{What is the ceiling performance of the target task?}
\end{enumerate}

We illustrate PGF with the help of a toy example and later discuss its properties. %~\ref{}.

\xhdr{Toy Example} Consider three target tasks with different baselines and ceiling performance, as shown in Table~\ref{tab:pgf_example}. 

\begin{table}[t!]
\centering
\resizebox{\linewidth}{!}{
\begin{tabular}{ccccccc}
\toprule
\textbf{Task} & Baseline (\%) & After FT (\%) & Ceiling (\%) & Raw Gain & PGF \\
\midrule
A & 90 & 93 & 95 & +3 & 0.60 \\
B & 40 & 50 & 95 & +10 & 0.18 \\
C & 98 & 97 & 99 & $-1$ & $-0.50$ \\
\bottomrule
\end{tabular}}
\caption{Illustration of the Perfection Gap Factor (PGF) across three target tasks.}
\label{tab:pgf_example}
\end{table}

\begin{table}[!t]
\centering
\resizebox{\linewidth}{!}{%
\begin{tabular}{lccc}
\toprule
\textbf{Task Name} & \textbf{Abbreviation} & \textbf{Perceptual Level} & \textbf{Granularity} \\
\midrule
Art style & AS & Mid-level & Image-level \\
Counting & CN & High-level & Image-level \\
Forensics detection & FD & High-level & Image-level \\
Functional corr. & FC & High-level & Pixel-level \\
Jigsaw & JG & Mid-level & Crop-level \\
Multi-view reasoning & MR & Mid-level & Image-level \\
Object localization & OL & High-level & Crop-level \\
Relative depth & RD & Low-level & Pixel-level \\
Relative reflectance & RR & Low-level & Pixel-level \\
Semantic corr. & SC & High-level & Pixel-level \\
Spatial reasoning & SR & Mid-level & Image-level \\
Visual corr. & VC & Low-level & Pixel-level \\
Visual similarity & VS & High-level & Image-level \\
\bottomrule
\end{tabular}%
}
\caption{BLINK tasks with abbreviation and classification by Perceptual Level and Granularity.}
\label{tab:task_classification}
\end{table}

\noindent Although Task~B shows the largest raw gain (+10), it closes only 18\% of its remaining gap to perfection. In contrast, Task~A, despite a smaller +3 gain, closes 60\% of its available headroom. Task~C illustrates the opposite case: a small drop from 98\% to 97\% yields a PGF of $-0.5$, reflecting a complete loss of the narrow headroom. This example illustrates how PGF provides a normalized and interpretable view of task transferability, enabling comparison across tasks with varying difficulty levels and performance ceilings.

% We discuss the detailed behavior of PGF in the Appendix~\ref{sec:pg_behavior}.
\section{Results and Analysis}
% - add seeds, stat tests, etc
\xhdr{Experimental Setup} % can make \xhdr to save space if required 
We consider a diverse set of 13 multimodal perception tasks\footnote{We exclude the “IQ Test” task from our analysis because it was manually constructed and does not have a corresponding training set.}, from the widely followed BLINK Benchmark~\cite{fu2024blink}, listed in Table~\ref{tab:task_classification}. A detailed description of these tasks can be found in the supplementary material. We employ three variants (3B, 7B, and 32B) from the open-weight Qwen-2.5-VL lineup~\cite{qwen2vl} as base models for our experiments. These models are finetuned independently on each task using LoRA~\cite{hu2022lora}, and evaluation is carried out on the validation splits of all tasks. Since BLINK itself only provides validation and test splits, we construct training data by retrieving the original datasets used in BLINK, adhering to the same task definitions and response formats. To assess robustness, all experiments are performed with four different random seeds. Unless noted otherwise, we set the ceiling performance $U_j=100$ for calculating PGF.
% \begin{figure*}[t]
%     \centering
%     \begin{subfigure}[b]{0.48\textwidth}
%         \centering
%         \includegraphics[width=\linewidth]{figures/transferability_qwen-2.5-32b.pdf}
%         \caption{Perfection Gap Factor heatmap for Qwen-2.5-VL 32B.}
%         \label{fig:pgf_32b}
%     \end{subfigure}
%     \hfill
%     \begin{subfigure}[b]{0.48\textwidth}
%         \centering
%         \includegraphics[width=\linewidth]{figures/clique_graphs/qwen-2.5vl-32B-instruct/cliq1_task_graph_positive.png}
%         \caption{Example of a positive clique found in Qwen-2.5-VL 32B analysis.}
%         \label{fig:cliq_eg}
%     \end{subfigure}
%     \caption{(a) Perfection Gap Factor heatmap and (b) example positive clique for Qwen-2.5-VL 32B.}
%     \label{fig:pgf_clique}
% \end{figure*}

% \subsection{Main results}
We first visualize the cross-task transfer matrix for each model using PGF. Each row corresponds to a source (finetuning) task, and each column denotes the target (evaluated) task (Figure~\ref{fig:pgf}). Figure~\ref{fig:teaser} shows the PGF heatmap for Qwen-2.5-VL 32B, revealing a structured pattern of both positive and negative transfer that persists across random seeds. To highlight salient transfer relationships, we construct task transfer graphs by retaining the top 20\% of strongest positive and negative PGF edges (Figure~\ref{fig:teaser}).
To further understand the broad dynamics of task transferability, we investigate :
\begin{enumerate}[label=\circnum{\arabic*}]%, leftmargin=*]
    \item How does transferability vary with task perception level (Low, Mid, High)?
    \item How does transferability vary with task granularity (Pixel, Crop, Image)?
    \item How does transferability scale with model size?
\end{enumerate}
% \subsection{Task graphs}

\begin{figure}[!t]
    \centering
    \includegraphics[width=\linewidth]{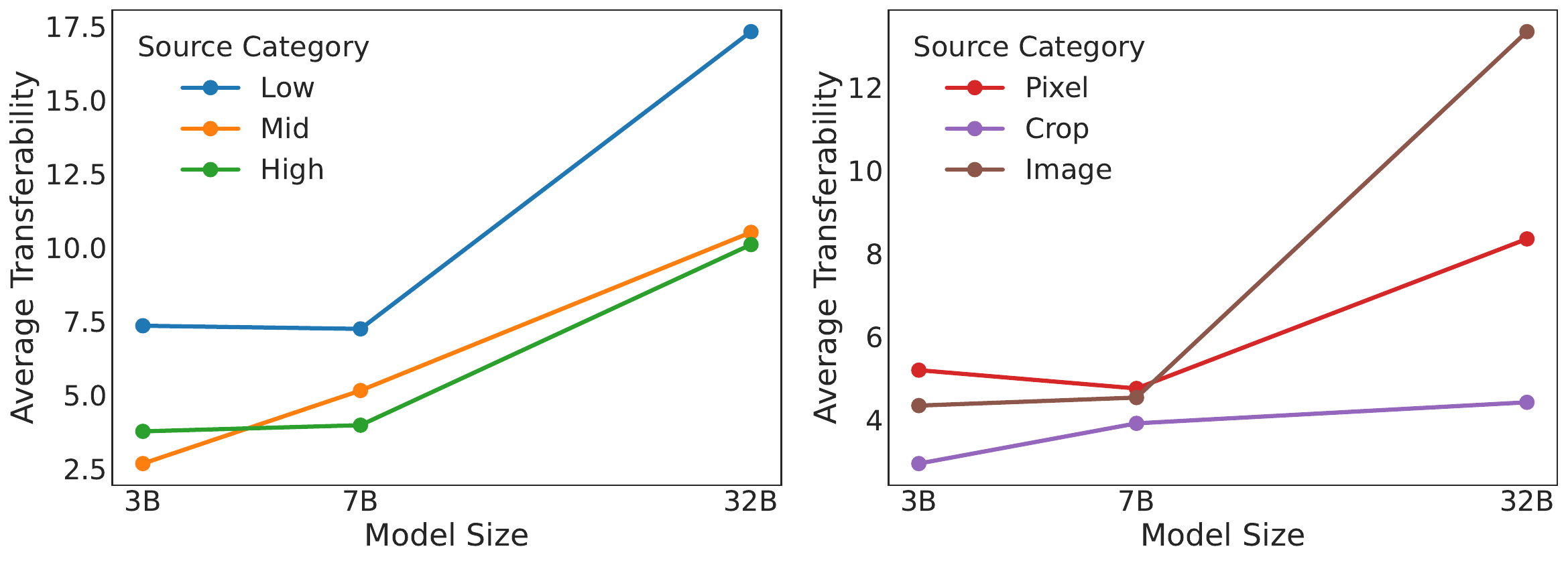}
    \caption{Average positive transferability trends across granular and perceptual levels. We observe that positive transferability increases with model size and generally low-level and image-level are highly transferable. Detailed category-wise heatmaps are provided in the supplementary material.}
    \label{fig:group-transfer}
\end{figure}

\begin{figure}[!t]
    \centering
    \includegraphics[width=\linewidth]{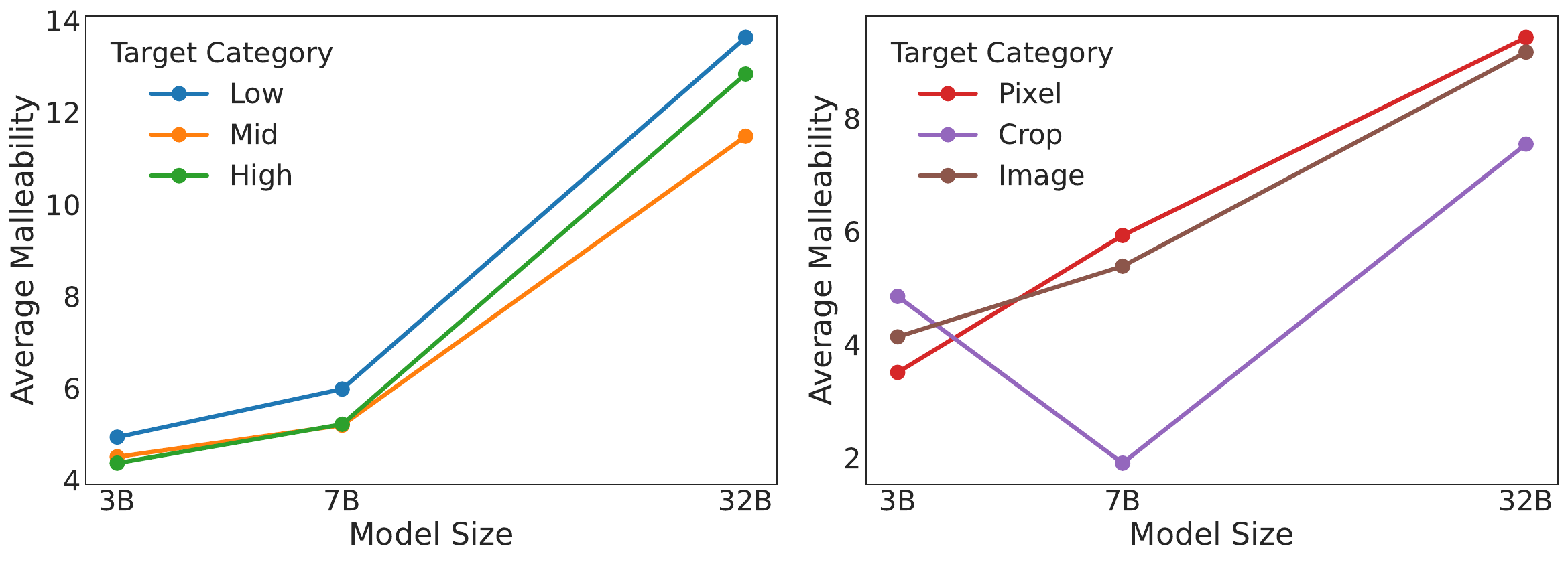}
    \caption{Average positive malleability trends across granular and perceptual levels. We observe that positive malleability increases with model size and generally low-level benefit the most from finetuning. Detailed category-wise heatmaps are provided in the supplementary material.}
    \label{fig:group-malleability}
\end{figure}

\subsection{Task Transfer across Categories}

The BLINK benchmark organizes tasks into 2 types of categories, as shown in Table~\ref{tab:task_classification}. To study transfer dynamics at this broad level, we examine both positive and negative transferability between these categories across all model sizes. For each ordered pair of source and target categories, we compute the category-level transferability by averaging the positive and negative transfer effects between every task in the source category and every task in the target category, aggregated over four random seeds.

\xhdr{Perceptual Level vs Transfer} In the task categorization, $\{\text{low-level},\text{mid-level},\text{high-level}\}$, we find that finetuning on low-level tasks (Relative Depth, Relative Reflectance, Visual Correspondence) has the highest average magnitude of positive task transferability across categories, for all model sizes. In addition, low-level tasks also benefit the most on average from finetuning, achieving the highest positive malleability in all models. We present the average transferability and average malleability in Figure ~\ref{fig:group-transfer} and Figure ~\ref{fig:group-malleability} respectively.

\begin{tcolorbox}[takeaways, title=Key Takeaway]
Low-level tasks (Relative Depth, Relative Reflectance, Visual Correspondence) are highly positively transferable and malleable. Finetuning on low-level tasks is beneficial compared to mid and high-level tasks.
\end{tcolorbox}

\xhdr{Granularity vs Transfer} In the task categorization, $\{\text{pixel-level},\;\text{crop-level},\;\text{image-level}\}$, we find that finetuning on image level tasks on average results in the highest positive transferability (Figure~\ref{fig:group-transfer}). In Figure~\ref{fig:group-malleability}, we observe that both pixel and image level tasks are malleable across model sizes.

\begin{tcolorbox}[takeaways, title=Key Takeaway]
Image-level tasks (Art Style, Counting, Forensic Detection, etc.) exhibit the higher positive transferability compared to pixel- and crop-level tasks. Both image-level and pixel-level tasks show higher malleability than crop-level tasks.
\end{tcolorbox}

 % Figures~\ref{fig:category_32B},~\ref{fig:category_7B},~\ref{fig:category_3B} (c, d) illustrates the positive and negative transferability between semantic categories. On average, finetuning on \textit{pixel-level} tasks yields the highest magnitude of both positive and negative transferability across all three models, with one exception: for Qwen2.5-VL-3B, the positive transferability of pixel-level tasks is a close second. This may be attributed to the fact that pixel-level tasks—often correspondence tasks—are generally harder to learn, and thus exert stronger influence on crop- and image-level tasks such as object localization and similarity. We further observe that, when finetuned on pixel-level tasks, the magnitude of positive transferability is consistently higher than that of negative transferability. We also find that in two out of three cases pixel-level tasks benefit the most on average from finetuning.
 % Figures~\ref{fig:category_32B},~\ref{fig:category_7B},~\ref{fig:category_3B} (a, b) illustrates the positive and negative transferability between hierarchical categories. Across model sizes, we observe that finetuning on low-level tasks like relative depth, relative reflectance, and visual correspondence has the highest average magnitude of positive task transferability. This is similar to the trend observed in semantic task transferability because all low-level tasks are pixel-level tasks~\ref{tab:categories}.

\subsection{Model Scale vs Transfer}

\begin{figure}[!t]
    \centering
    \begin{subfigure}{0.48\linewidth}
        \centering
        \includegraphics[width=\linewidth]{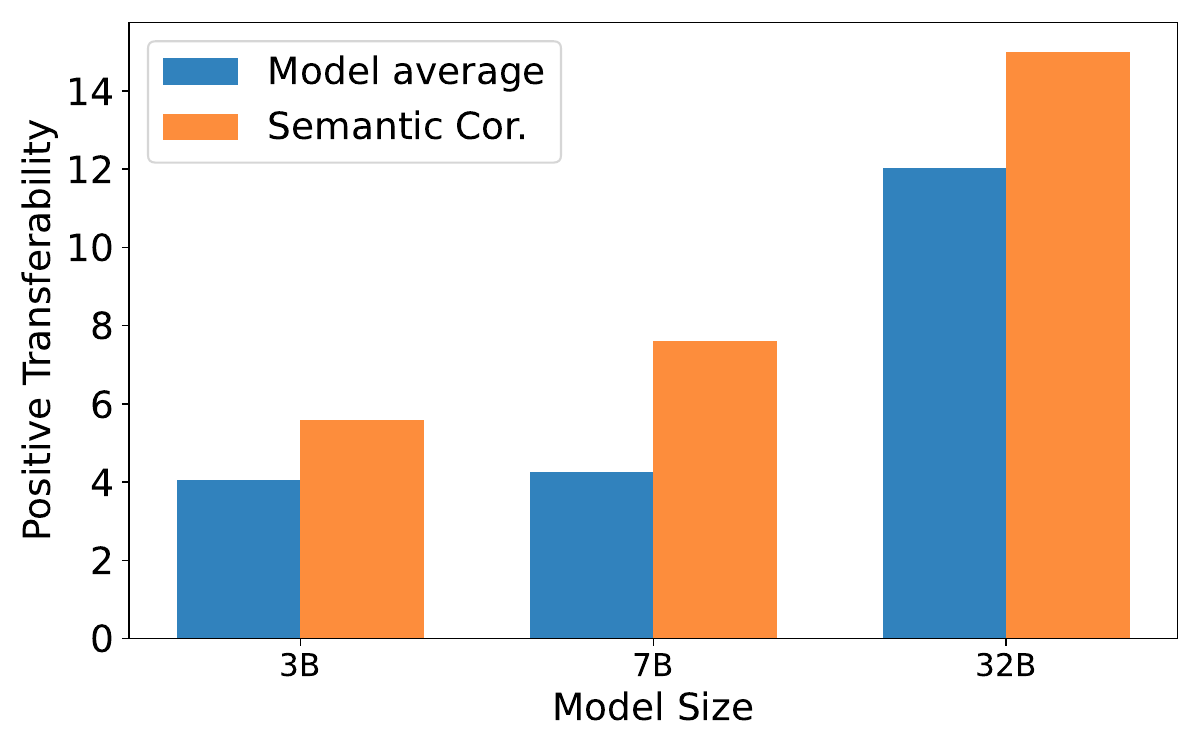}
        \caption{Positive task transferability across model sizes.}
        \label{fig:pos_trans_32}
    \end{subfigure}
    \hfill
    \begin{subfigure}{0.48\linewidth}
        \centering
        \includegraphics[width=\linewidth]{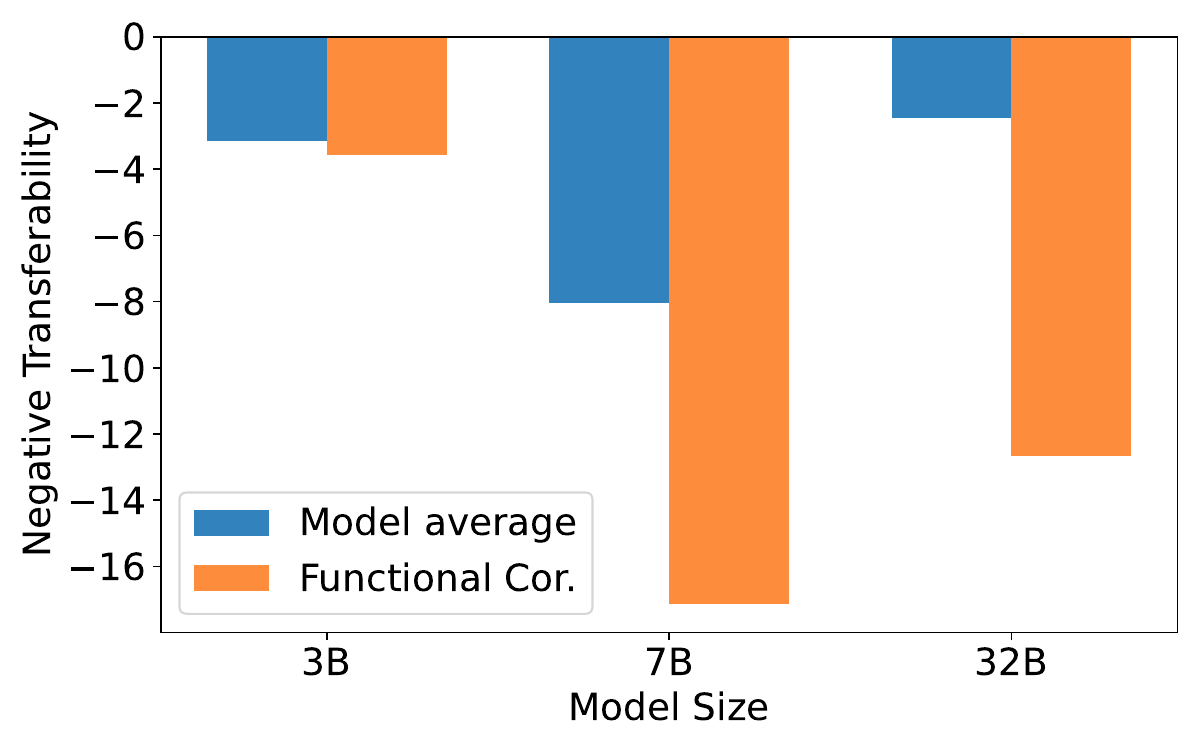}
        \caption{Negative task transferability across model sizes.}
        \label{fig:neg_trans_32}
    \end{subfigure}
    \caption{Task transferability trends across model sizes in Qwen-2.5-VL. As expected, as model size increases, the average positive transferability increases.}
    \label{fig:transferability_32B}
\end{figure}

To understand how task transferability varies with increasing model size, we analyze the average positive and negative task transferability across all tasks for each model in Figure~\ref{fig:transferability_32B}. As expected, as model size increases, the average positive transferability also increases. This finding aligns with the intuition that increased representational capacity allows models to capture more generalizable features, leading to better transfer of beneficial knowledge across diverse tasks. However, there is no consistent trend with the average negative transferability. We provide detailed PGF heatmaps for all model sizes in the supplementary material.

\begin{tcolorbox}[takeaways, title=Key Takeaway]
The magnitude of positive transferability and malleability increases with model size. 
\end{tcolorbox}

\section{Additional Results}

In the previous sections, we examined task transfer from a broader perspective. Here, we shift to a more granular view and analyze how small clusters of mutually influential tasks emerge within the broader transfer landscape, and how distinct categories of transfer tendencies (\textit{task personas}) characterize the roles that individual tasks play during finetuning.

\subsection{Cliques of Cooperation}

The improvements across tasks are not uniformly distributed and instead exhibit structured clusters of mutual influence. 
To formalize this observation, we define the notion of a \textit{task clique} within the transfer graph.

\begin{definition}[Task Clique]
Let $\{T_{k}\}_{k=1}^{N}$ denote the set of all tasks, and let $\mu_{i \to j}$ denote a transferability score from source task $T_i$ to target task $T_j$. A subset of tasks $C \subseteq \{T_{k}\}_{k=1}^{N}$ is said to form a \textbf{task clique} if, for all ordered pairs of distinct tasks $(T_i, T_j)$ with $T_i, T_j \in C$ and $i \neq j$, the induced transfer values $\mu_{i \to j}$ exhibit consistent sign (all positive or all negative).
Tasks that mutually induce positive transfer form a \textit{Positive Clique}, while those that mutually induce negative transfer form a \textit{Negative Clique}.
\end{definition}

To assess whether the extracted cliques are stable across seeds, we perform Wilcoxon tests and identify multiple statistically significant cliques of varying sizes across models. In the smaller variants (3B and 7B), the largest cliques comprise 3–4 tasks, while in Qwen-2.5-VL 32B, we observe a maximal positive clique of size 9, as illustrated in Figure~\ref{fig:cliq}. Detailed clique statistics and additional examples are provided in the supplementary material.

\subsection{Task Personas}

To uncover characteristic transfer behaviors, we group tasks into distinct \textit{personas}: source tasks that consistently help or hinder others (\textit{Donors} and \textit{Pirates}), and target tasks that readily absorb or resist transfer (\textit{Sponges} and \textit{Sieves}).

\xhdr{Donors and Pirates} Tasks which consistently exhibit a higher magnitude of positive transferability than the model average positive transferability across the model sizes are called \textit{Donor Tasks}. Similarly, tasks that consistently induce a higher magnitude of negative transfer than the model average negative transferability across the model sizes are called \textit{Pirate Tasks}. We identify \textit{Semantic Correspondence} as a donor task and \textit{Functional Correspondence} as a pirate task. Unpaired t-tests over transferability values across seeds validate that \textit{Semantic Correspondence} is a statistically significant donor task across all models~($p<0.01$ across models), whereas \textit{Functional Correspondence} is a significant pirate task in both 3B and 7B variants~($p<0.05$).

\begin{figure}[!t]
    \centering
    \includegraphics[width=\linewidth]{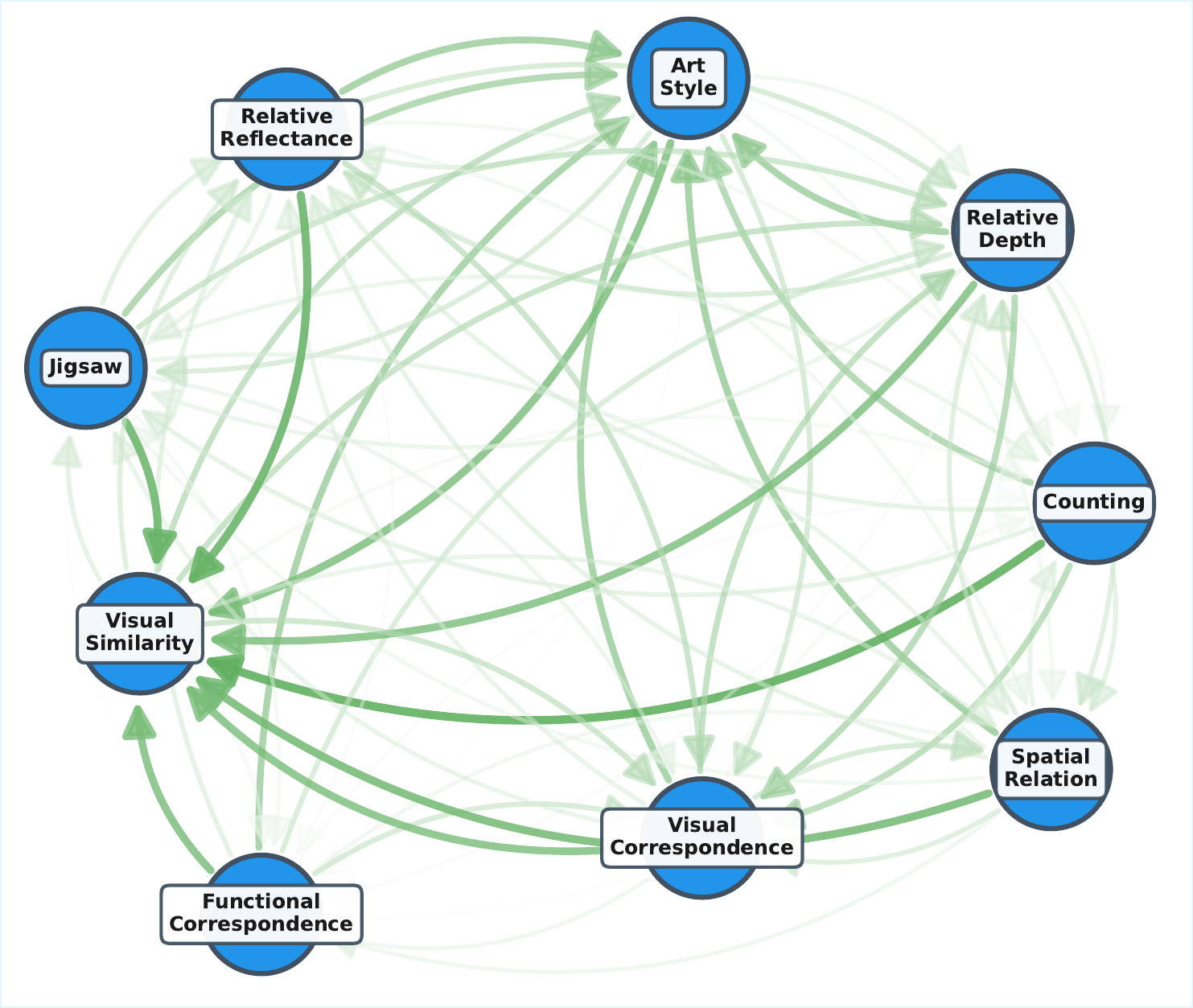}
    \caption{Positive clique of size 9 from Qwen-2.5-VL 32B.}
    \label{fig:cliq}
\end{figure}

% \xhdr{Malleability} To characterize how sensitive a target task is to finetuning on other tasks, we introduce the notion of malleability. A target task can be considered highly malleable if finetuning on different source tasks leads to significant change (positive or negative) in the performance on that task. To quantify this value, we compute the average of the positive and negative PGF scores induced on that task by other source tasks. Each score is scaled with logistic weighting, following the same scheme used for transferability. This weighting emphasizes tasks that benefit from many others (positive malleability), and conversely highlights tasks that are consistently impaired (negative malleability).

\xhdr{Sponges and Sieves} Tasks that consistently exhibit above average positive malleability across model sizes are classified as \textit{Sponge Tasks}. On the other hand, tasks that consistently exhibit a higher magnitude of negative malleability than the model average across model sizes are classified as \textit{Sieve Tasks}. We identify multiple Sponge tasks: \textit{Visual Similarity}, \textit{Relative Depth} and \textit{Relative Reflectance}, whereas \textit{Forensic Detection} emerges as the sole Sieve task. We conduct unpaired t-tests across seeds and find \textit{Visual Similarity} and \textit{Relative Depth} to be statistically significant Sponge tasks across models~($p < 0.001$). \textit{Forensic Detection} is a significant Sieve in both 3B and 32B variants~($p<0.005$).

% FD: Significant across 3B and 32B with p < 0.005
% VS: 3, 7, 32 p < 0.001
% RD: 
% RR:

% To find \textit{Sponges and Sieves} effectively, we Relative Depth demonstrates high positive malleability, acting as a Sponge, as shown in Figures~\ref{} and Figure~\ref{} in Appendix. Visual Correspondence emerges as a Sieve showing deterioration in accuracy across tasks and models.

\subsection{Transfer to Spatio-Temporal Tasks}

\begin{figure}[!t]
    \centering
    \includegraphics[width=\linewidth]{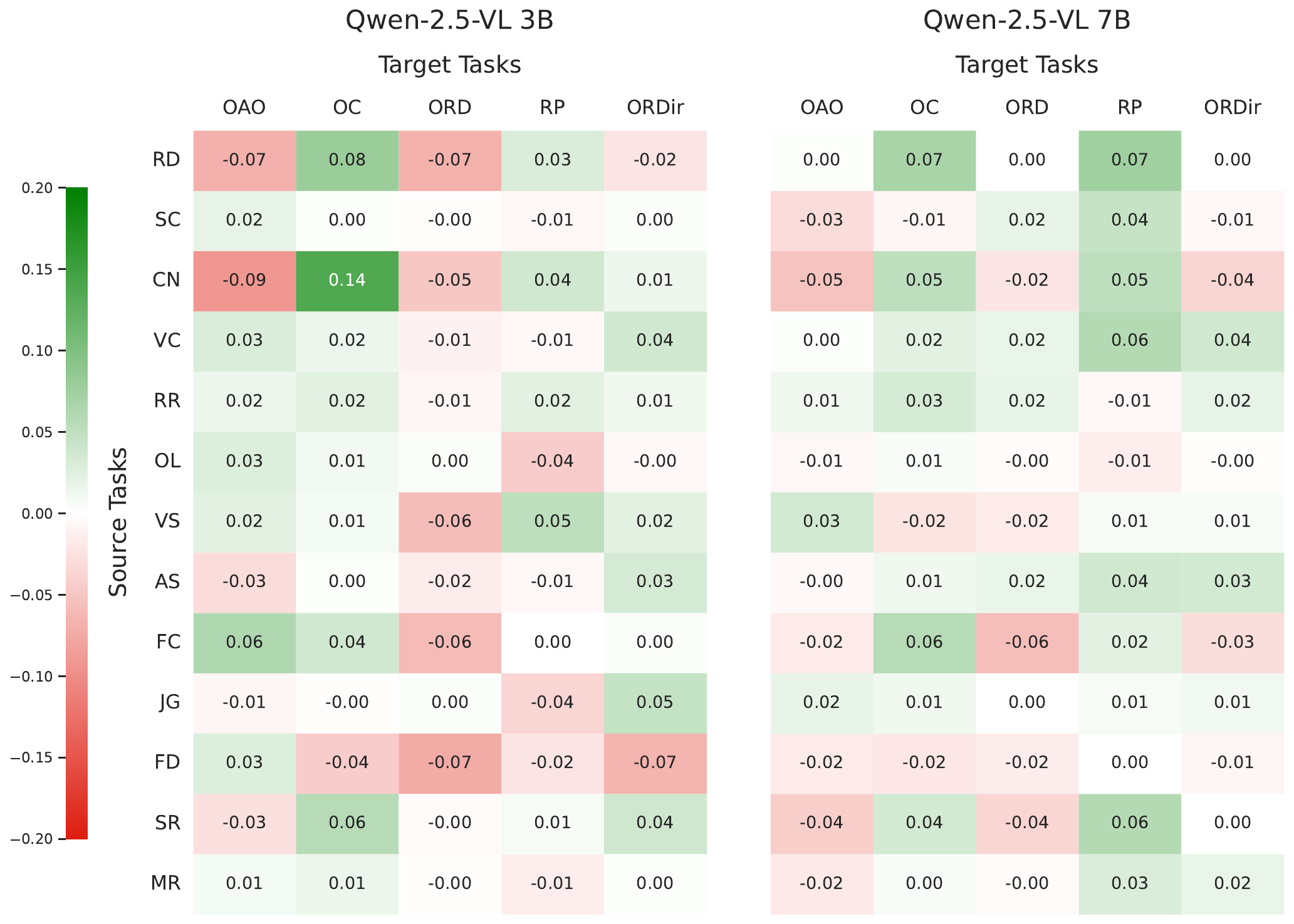}
    \caption{PGF heatmaps for Qwen-2.5-VL 3B (left) and 7B (right) models across the VSI benchmark. Consistent with previous findings, \textit{Relative Reflectance} and \textit{Forensic Detection} emerge as donor task and pirate task, respectively.}
    \label{fig:vsi_pgf}
\end{figure}

To study the effects of perception task transfer to video-based tasks, we evaluate the finetuned checkpoints on VSI Bench~\cite{yang2025thinking} which contains a series of spatio-temporal tasks\footnote{We do not include size estimation tasks from VSI-Bench in our analysis.}, such as Object Counting, Object Appearance Order and Route Planning. The benchmark comprises over 5,000 question-answer pairs derived from nearly 288 egocentric indoor videos drawn from public 3D-scene datasets. It evaluates how well multimodal models perceive, recall, and reason about spatial layouts from egocentric video.
We study this cross-modal transfer in the Qwen-2.5-VL 3B and 7B variants and limit the analysis to the following tasks: Object Appearance Order (OAO), Object Counting (OC), Object Relative Distance (ORD), Route planning (RP), Object Relative Direction (ORDir). The results are shown in Figure~\ref{fig:vsi_pgf}. Consistent with our previous findings, we note that \textit{Relative Reflectance} emerges as a donor task and \textit{Forensic Detection} acts as a pirate task. Moreover, we identify \textit{Object Counting} as a sponge task while \textit{Object Appearance Order} and \textit{Object Relative Distance} act as sieve tasks.

\begin{tcolorbox}[takeaways, title=Key Takeaway]
Image-level perception tasks induce positive transfer to video-based tasks as well, demonstrating consistent trends in task transfer.
\end{tcolorbox}

\subsection{Data Selection with PGF}

\begin{figure}[t]
    \centering
    \includegraphics[width=\linewidth]{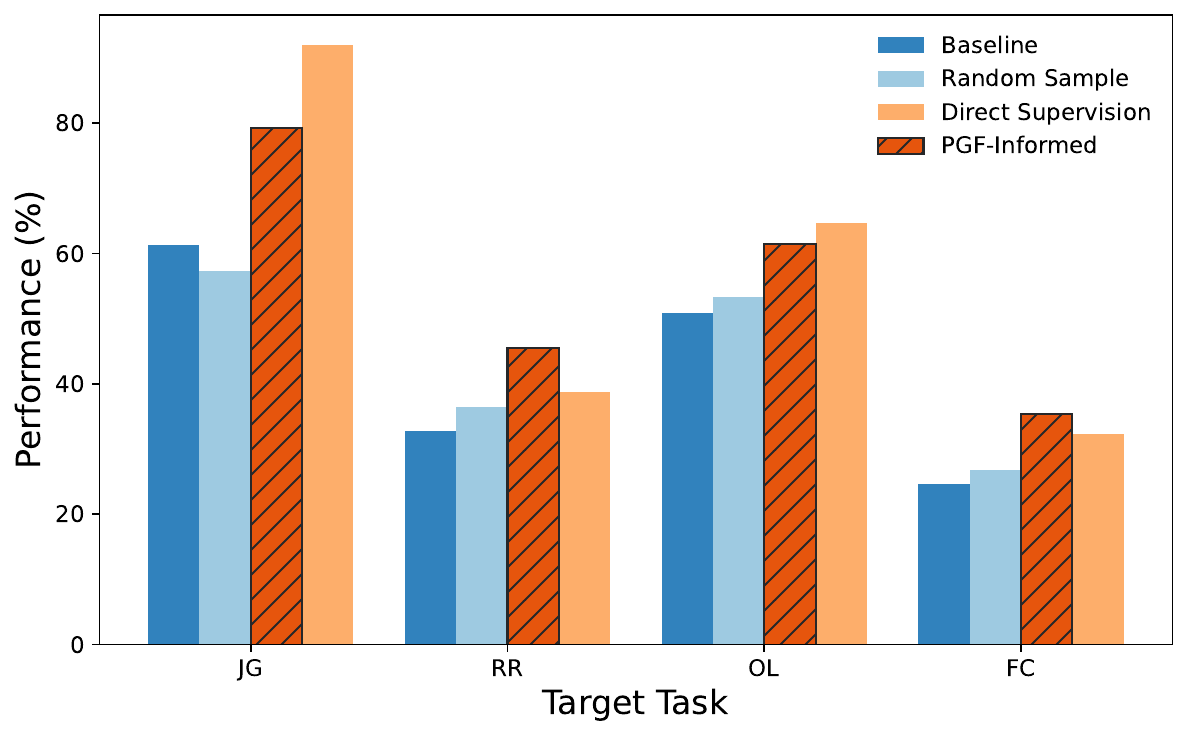}
    \caption{Performance comparison under different dataset selection strategies. PGF-informed mixtures consistently outperform random mixtures and even surpass direct supervision in two cases.}
    \label{fig:pgf_selection}
\end{figure}

Lastly, we demonstrate how Perfection Gap Factor can guide dataset selection in the absence of task-specific training data. We consider a setting where we aim to optimize performance on some task $T$ for which no training data is available. Instead, we have access to datasets from several related tasks and seek an optimal mixture to improve performance on $T$. We propose using the most transferable tasks to $T$ (above a certain threshold) using the PGF metric. We compare the PGF-informed dataset mixtures against randomly sampled mixture trained to optimize $T$. The baseline model and $\mathcal{M}(T)$ are considered the lower and upper bound respectively. The results are presented in Figure~\ref{fig:pgf_selection}. Although we limit our experiments to Qwen-2.5-VL 7B, we consistently find that PGF-informed data selection leads to better performance across multiple target tasks, demonstrating its effectiveness in guiding data selection. In case of Jigsaw and Object Localisation, PGF-informed data selection even outperforms finetuning directly on the target task itself. We note that this experiment is a preliminary finding and is included only to illustrate the potential of PGF for dataset selection. A comprehensive study of PGF-guided dataset mixtures is out of scope for this work and will be pursued in future research.

\begin{tcolorbox}[takeaways, title=Key Takeaway]
When lacking supervised data, PGF-informed data selection can inform alternative dataset designs which can match and even exceed the performance compared to direct finetuning. 
\end{tcolorbox}

% \textcolor{red}{Write that this is out of scope and is only an initial finding to avoid reviewer madness.}

% \begin{table}
% \centering
% \resizebox{\linewidth}{!}{
% \begin{tabular}{ccccc}
% \toprule
% \multicolumn{1}{c}{\begin{tabular}[c]{@{}c@{}}\textbf{Target} \\\textbf{Task} \end{tabular}} &
% \multicolumn{1}{c}{\begin{tabular}[c]{@{}c@{}}\textbf{Baseline}\\ $\mathcal{M}$\end{tabular}} &
% \multicolumn{1}{c}{\begin{tabular}[c]{@{}c@{}}\textbf{Random Sample}\\ \textbf{Data Mixture}\end{tabular}} &
% \multicolumn{1}{c}{\begin{tabular}[c]{@{}c@{}}\textbf{PGF-Informed}\\ \textbf{Data Mixture}\end{tabular}} &
% \multicolumn{1}{c}{\begin{tabular}[c]{@{}c@{}}\textbf{Target}\\ $\mathcal{M}(T)$\end{tabular}} \\ \midrule
% JG                    & 61.33\% & 57.33\% & 79.33\% & 92.00\% \\
% RR      & 32.84\% & 36.42\% & 45.52\% & 38.81\% \\
% OL      & 50.82\% & 53.28\% & 61.48\% & 64.75\% \\
% FC      & 24.62\% & 26.77\% & 35.38\% & 32.31\% \\
% \bottomrule
% \end{tabular}
% }
% \caption{Performance comparison under different dataset selection strategies.}
% \label{tab:pgf_selection}
% \end{table}

\section{Discussion}

    % * Implications: What the transfer graph tells us about how VLMs internalize perception tasks, designing finetuning curricula, avoiding harmful transfers, Practical guidance: which tasks are safe donors, which to avoid
    % * Limitations: BLINK’s multiple-choice setup (BLINK tasks are multiple-choice; transfer patterns could change for open-ended tasks), single VLM family, no open-ended tasks
    % * Future directions: Task design rules, cross-architecture generalization, probing representations

Our analysis of task transferability in Vision–Language Models (VLMs) reveals a rich structure in how perception capabilities interact under finetuning. Below, we unpack the broader implications of these findings, acknowledge key limitations, and propose promising directions for future work.

\begin{figure}[t]
    \centering
    \includegraphics[width=0.8\linewidth]{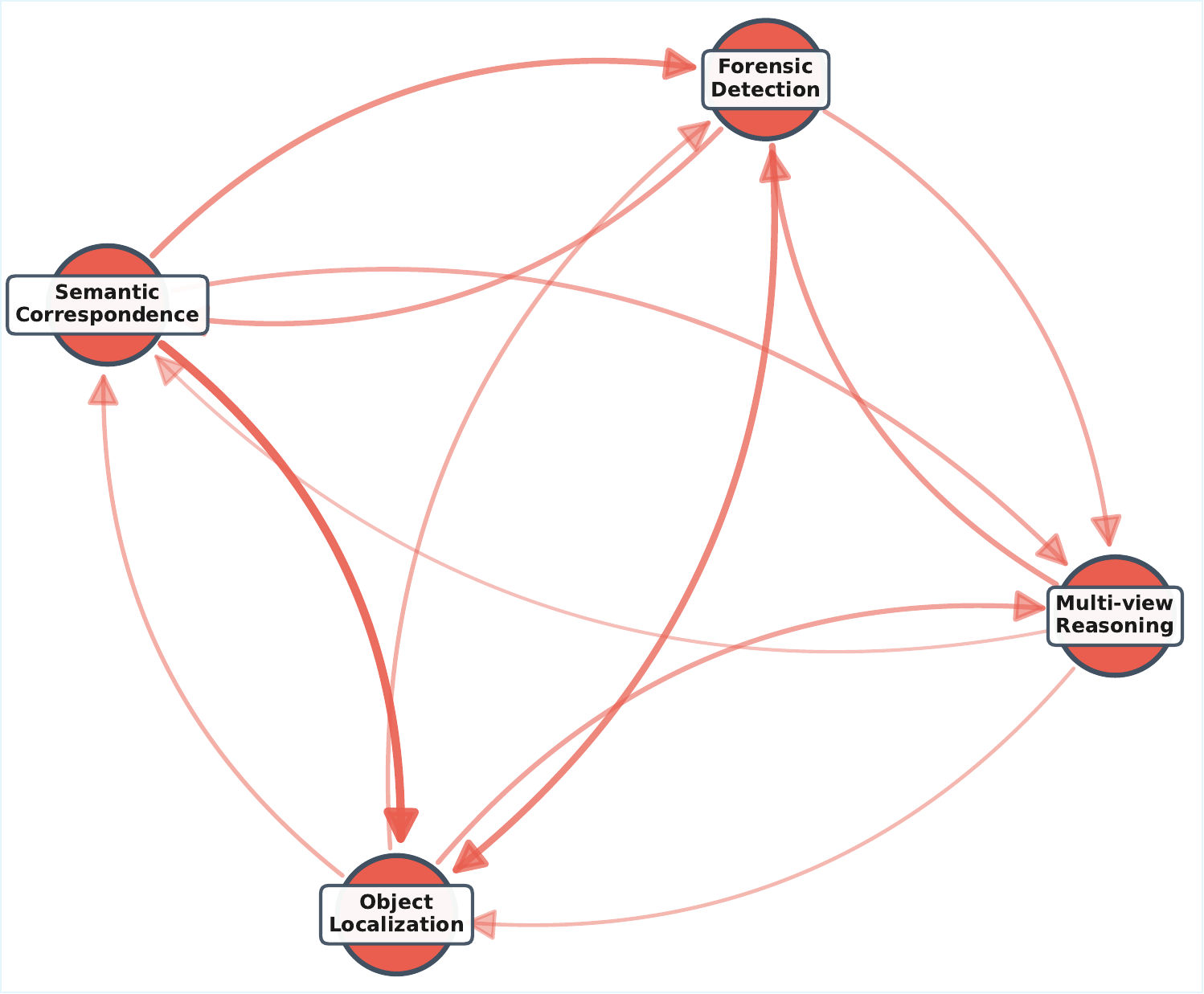}
    \caption{A negative clique of size 4 from Qwen-2.5-VL 32B.}
    \label{fig:neg_cliq}
\end{figure}

\xhdr{Implications} The emergence of a structured task transfer graph, characterized by cliques, personas, and scale-dependent patterns, suggests that VLMs do not treat perception tasks as independent learnings, but rather internalize them through shared or competing representational substructures. For instance, the consistent identification of low-level tasks (e.g., Relative Depth, Relative Reflectance) as strong sponges implies that early-stage visual features are highly reusable and adaptable across a wide range of downstream perception tasks. This supports the hypothesis that VLMs can benefit from hierarchical visual processing pathways. From a practical standpoint, these insights directly inform finetuning dataset design. We provide an early example of such a practical use-case. Notably, the fact that PGF-guided data selection can surpass direct target-task finetuning underscores the practical utility of our framework. Lastly, we note that our framework allows the discovery of negative cliques. This provides a unique understanding on the nature of deteriorative relationships between the considered tasks. An example of a negative clique is presented in Figure~\ref{fig:neg_cliq}.

\xhdr{Behavior of Perfection Gap Factor (PGF)} We further analyze the behavior of PGF. Figure~\ref{fig:pgf_behaviour} illustrates how PGF varies with baseline performance $x$ and accuracy change $k$ after finetuning. Several numerical properties emerge:

\begin{figure}[t!]
    \centering
    \includegraphics[width=\linewidth]{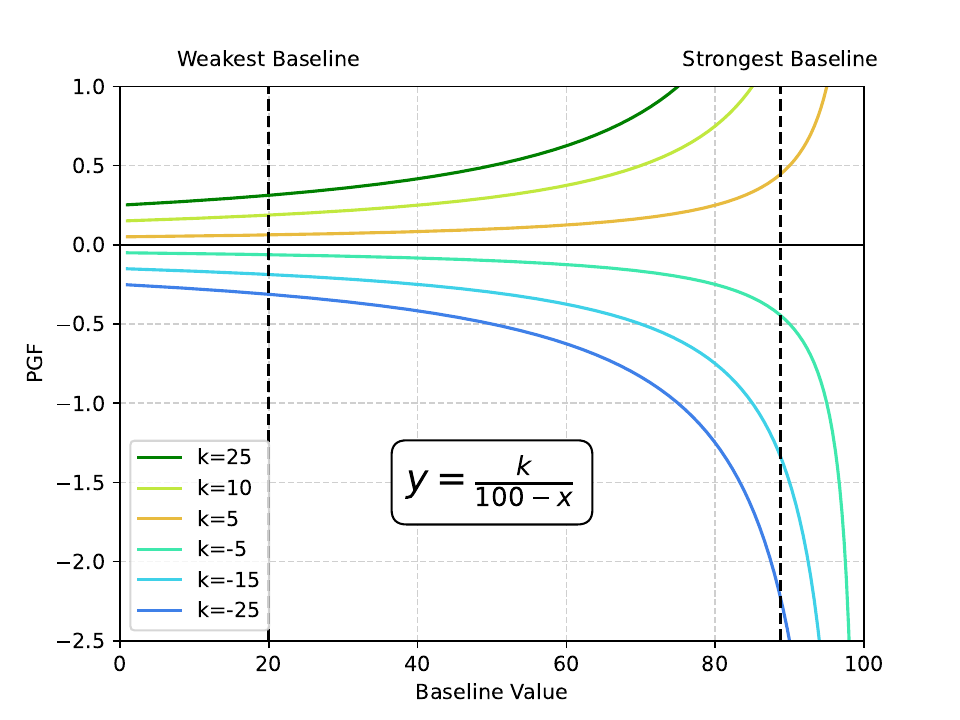}
    \caption{Behavior of PGF as a function of baseline accuracy ($x$) and change after finetuning ($k$).}
    \label{fig:pgf_behaviour}
\end{figure}

\begin{itemize}
    \item \textbf{Positive Bound}: For improvements ($k > 0$), PGF is capped at $1$, achieved when finetuning fully closes the gap to perfection ($k = 100 - x$).

    \item \textbf{Negative Bound}: For deterioration ($k < 0$), PGF admits a finite lower bound due to accuracy discreteness. With $m$ evaluation questions, the highest baseline strictly below $100\%$ is $x = 100\,(1 - \tfrac{1}{m})$. The worst deterioration is $k = -x$ (accuracy drops to zero), yielding
    \[
    \text{PGF}_{\min} \;=\; \frac{-x}{100 - x} 
    \;=\; \frac{-100(1 - \tfrac{1}{m})}{100/m} 
    \;=\; -(m-1).
    \]
    For instance, with $m = 200$ qns, $\text{PGF}_{\min} = -199$. The worst-case deterioration therefore grows linearly with $m$.

    \item \textbf{Asymmetry}: Since positive PGF is capped at $1$ but negative PGF can reach $-(m-1)$, PGF is inherently asymmetric, motivating our separate study of positive vs. negative transferability.

    \item \textbf{Ceiling Sensitivity}: Near-perfect baselines amplify PGF: small accuracy shifts yield disproportionately large values. This highlights ceiling-level improvements while penalizing degradations more harshly.
\end{itemize}

\xhdr{Limitations} The observations in this work come from comprehensive empirical analysis; however, it has certain limitations which can be interesting directions of future work. Our analysis is based mainly on benchmarks that model tasks in terms of multiple choice questions. This format can restrict the output space and suppress failure modes (or transfer patterns) that emerge in open-ended generation. Exploring open-ended generation for visual tasks would be a promising future direction. Besides, extending the studies to newer models will help understand the generalizability as well as evolution of architectures as pertains to their capabilities. %we study only one VLM family (Qwen-2.5-VL), albeit across three scales. While this allows controlled analysis of scaling effects, it leaves open whether our findings generalize across architectures.

\section{Conclusion}

In this work, we present the first systematic analysis of perception task transfer in vision-language models. To facilitate this analysis, we introduce a new metric called Perfection Gap Factor, which helps us quantify perception task transfer in VLMs. Through experiments with three state-of-the-art VLMs, we study how finetuning on a source task impacts zero-shot performance on other tasks. Our analysis reveals several key insights. Firstly, we note that positive task transferability increases with model size. Secondly, we identify distinct cliques of mutually beneficial and mutually detrimental tasks. Lastly, we investigate inter-task interactions and characterize them as task personas. This analysis provides actionable insights into how task interactions shape model behavior, guiding the development of finetuning strategies to enhance general-purpose VLMs.
{
    \small
    \bibliographystyle{ieeenat_fullname}
    \bibliography{main}
}

\clearpage
\setcounter{page}{1}
\setcounter{section}{0}

\onecolumn

\begin{center}
\Large
\textbf{Understanding Task Transfer in Vision-Language Models}\\
\vspace{0.5em}Supplementary Material \\
\vspace{1.0em}
\end{center}

\renewcommand{\thesection}{A.\arabic{section}}
\renewcommand{\thetable}{A.\arabic{table}}
\renewcommand{\thefigure}{A.\arabic{figure}}
\renewcommand{\theequation}{A.\arabic{equation}}

\newcommand{\ToCEntry}[3]{%
  \ifcase#1
    \noindent\ref{#3}. #2\hspace{1.5em}\dotfill\hspace{1.5em}\pageref{#3} \\ % Level 0
  \or
    \noindent\hspace*{2em}\ref{#3}. #2\hspace{1.5em}\dotfill\hspace{1.5em}\pageref{#3} \\ % Level 1
  \else
    \noindent\ref{#3}. #2\hspace{1.5em}\dotfill\hspace{1.5em}\pageref{#3} \\ % fallback
  \fi
}

\section*{Table of Contents}
\ToCEntry{0}{PGF Calculation and Heatmaps}{sec:pgf_calc}
\ToCEntry{0}{Accuracy Heatmaps}{sec:acc_maps}
\ToCEntry{0}{Task Category Trends}{sec:task_cat}
\ToCEntry{0}{Cliques across Model Sizes}{sec:cliques}
\ToCEntry{0}{Implementation Details}{sec:implement-det}
\ToCEntry{0}{Effect of Training Steps on PGF}{sec:step}
\ToCEntry{0}{LoRA Weights Analysis}{sec:lora}
\ToCEntry{0}{Generalization to Other Models}{sec:llava}
\ToCEntry{0}{Task Graph Visualizations}{sec:task_graph}
\ToCEntry{0}{PGF with Best Performance Ceiling}{sec:best_perf}
\ToCEntry{0}{Broader Impact}{sec:broader_impact}
\hrule

\section{PGF Calculation and Heatmaps}
\label{sec:pgf_calc}

We provide a pseudo-code to compute the Perfection Gap Factor in Algorithm~\ref{alg:pgf_code}. The goal of the metric is to quantify how much of the remaining achievable performance a model recovers through finetuning.

\begin{algorithm}[!htbp]
\begin{algorithmic}[1]
\Require Baseline accuracy $A_{\text{base}}$, finetuned accuracy $A_{\text{ft}}$, ceiling $U$, small constant $\epsilon$
\Ensure PGF value $\mu$

\State $\Delta \gets A_{\text{ft}} - A_{\text{base}}$   \Comment{accuracy change}
\State $\text{gap} \gets U - A_{\text{base}} + \epsilon$  \Comment{remaining room to improve}
\State $\mu \gets \Delta / \text{gap}$  \Comment{PGF definition}

\State \Return $\mu$
\end{algorithmic}
\caption{Pseduo-code to compute Perfection Gap Factor.}
\label{alg:pgf_code}
\end{algorithm}

\begin{figure*}[!htbp]
    \centering
    \includegraphics[width=0.85\linewidth]{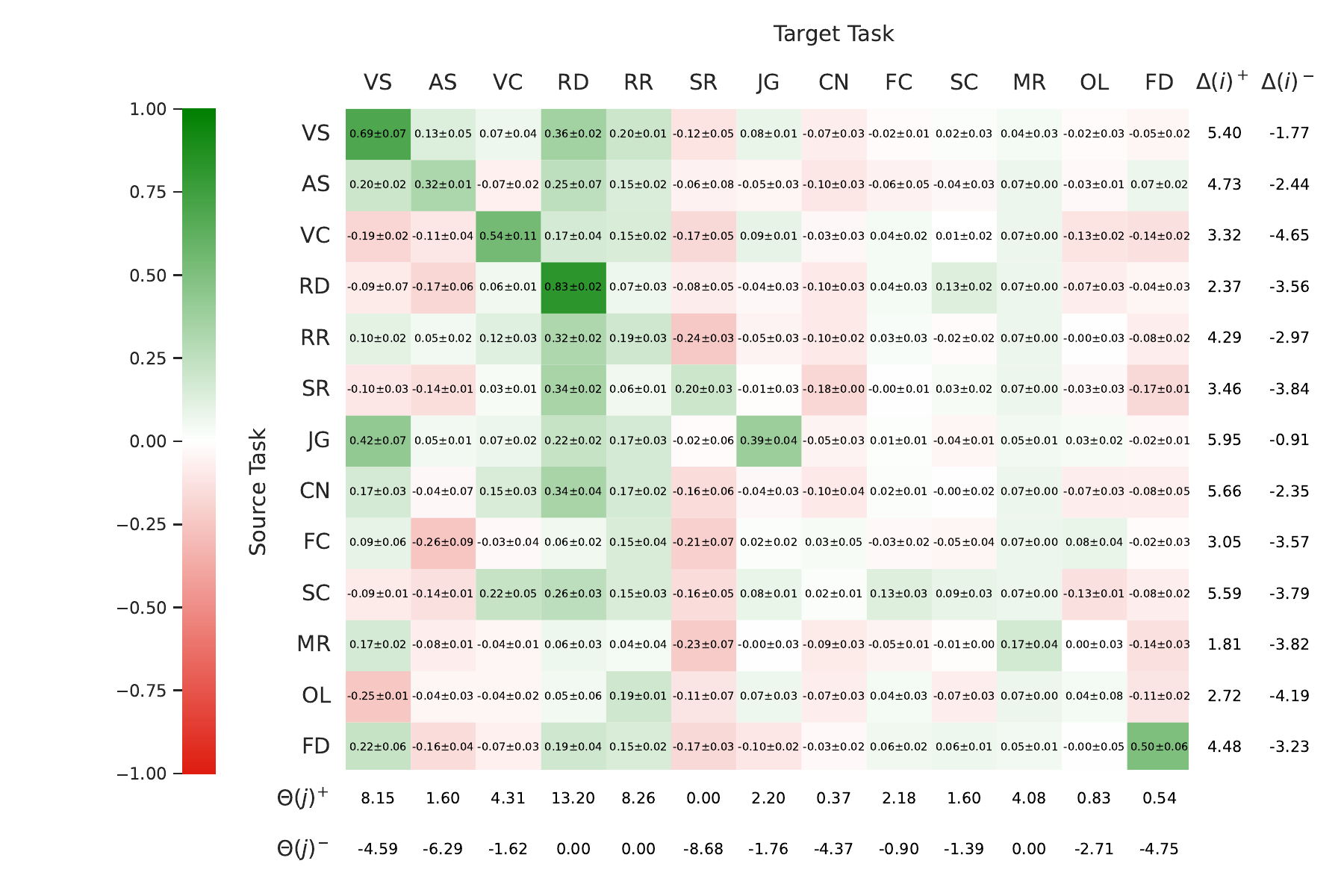}
    \caption{PGF Heatmap for Qwen-2.5-VL 3B.}
    \label{fig:pgf_detail_3}
\end{figure*}

\begin{figure*}[!htbp]
    \centering
    \includegraphics[width=0.85\linewidth]{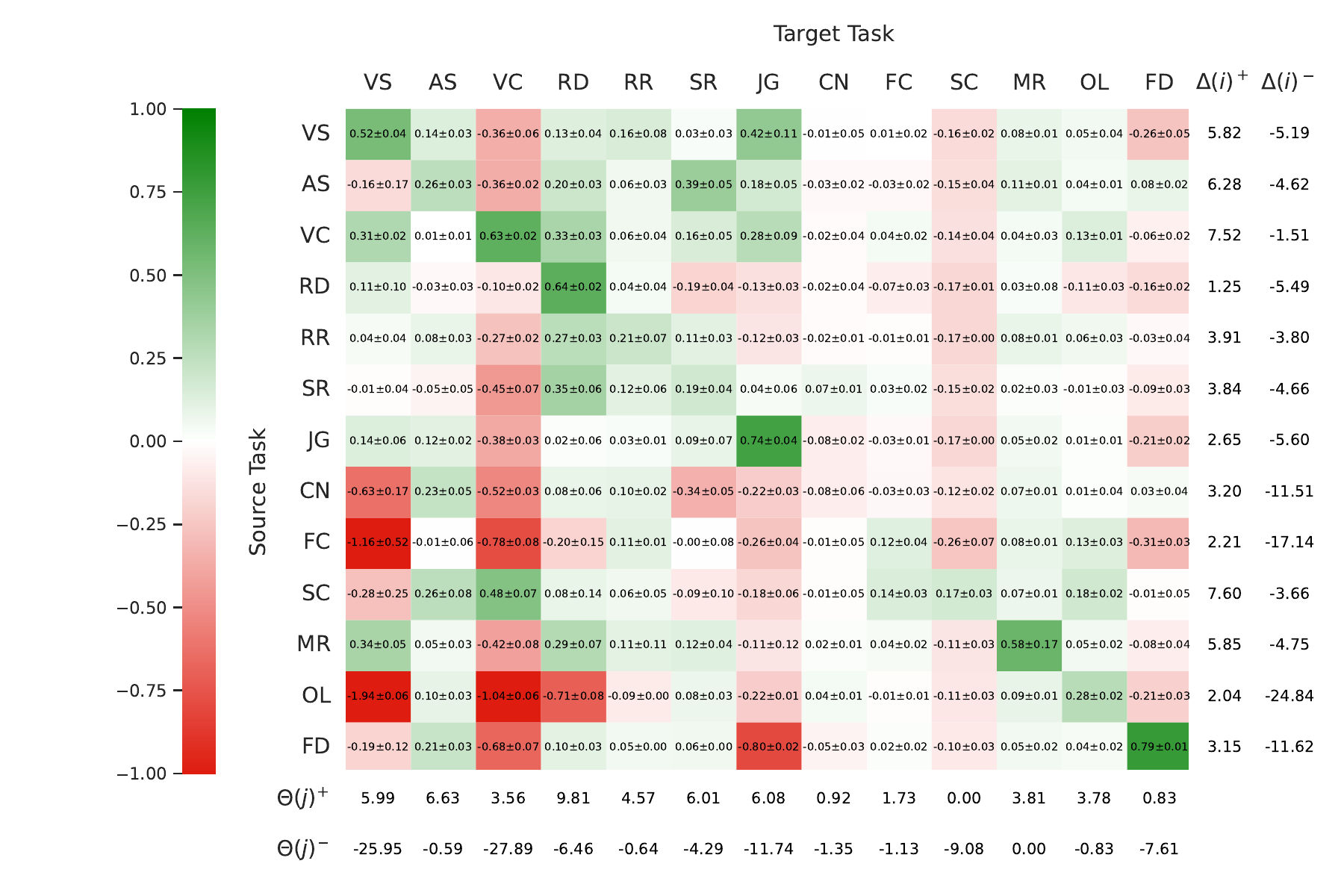}
    \caption{PGF Heatmap for Qwen-2.5-VL 7B.}
    \label{fig:pgf_detail_7}
\end{figure*}

\begin{figure*}[!htbp]
    \centering
    \includegraphics[width=0.85\linewidth]{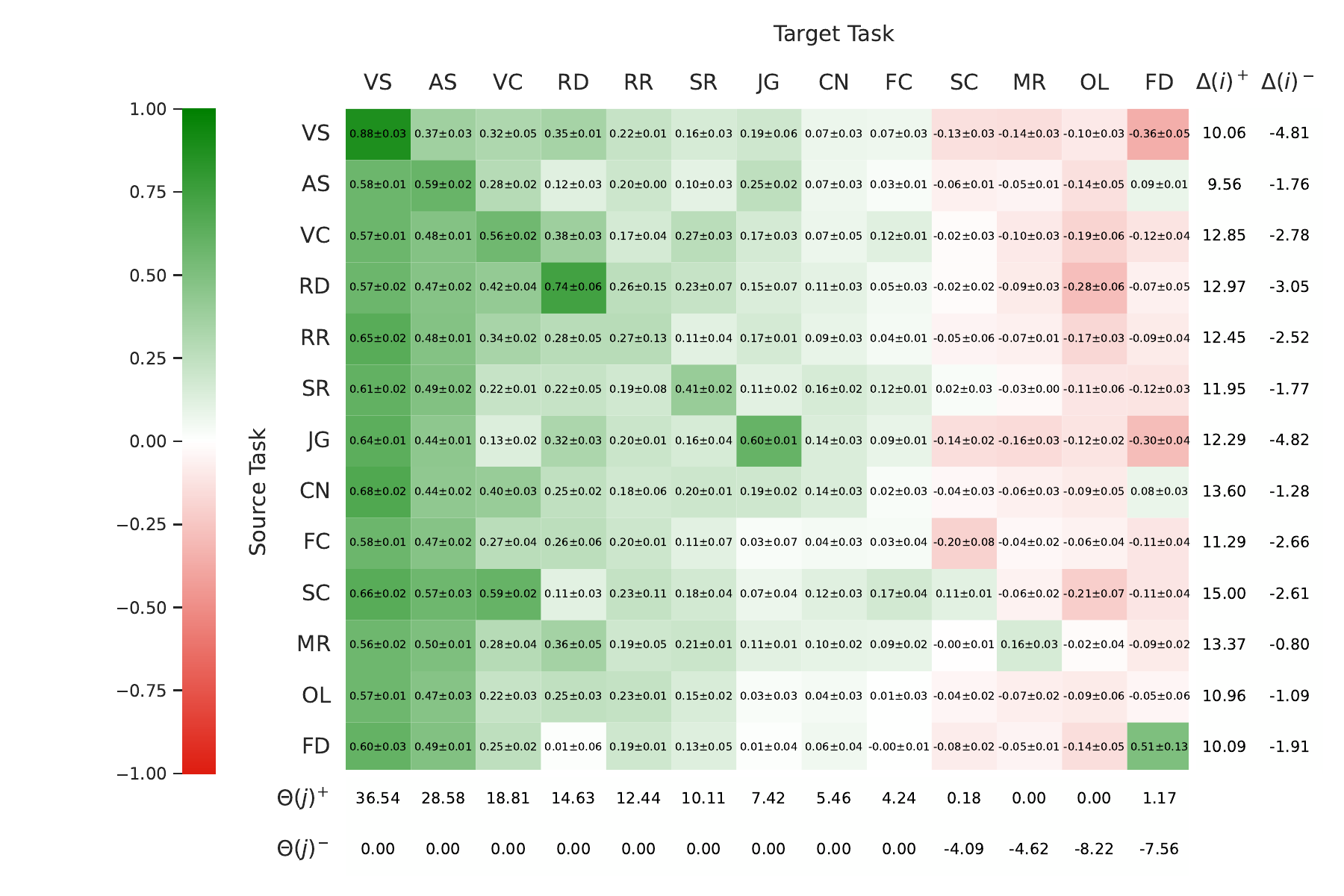}
    \caption{PGF Heatmap for Qwen-2.5-VL 32B.}
    \label{fig:pgf_detail_32}
\end{figure*}

We also provide PGF heatmap for the 13 tasks with mean PGF and standard deviation, alongside transferability and malleability in Figure~\ref{fig:pgf_detail_3}, Figure~\ref{fig:pgf_detail_7}, and Figure~\ref{fig:pgf_detail_32}, for 3B, 7B and 32B, respectively. We note that the standard deviation remains consistently small, suggesting that the results are stable rather than driven by noise.

\section{Accuracy Heatmaps}
\label{sec:acc_maps}

\begin{figure*}[!htbp]
    \centering
    \includegraphics[width=0.85\linewidth]{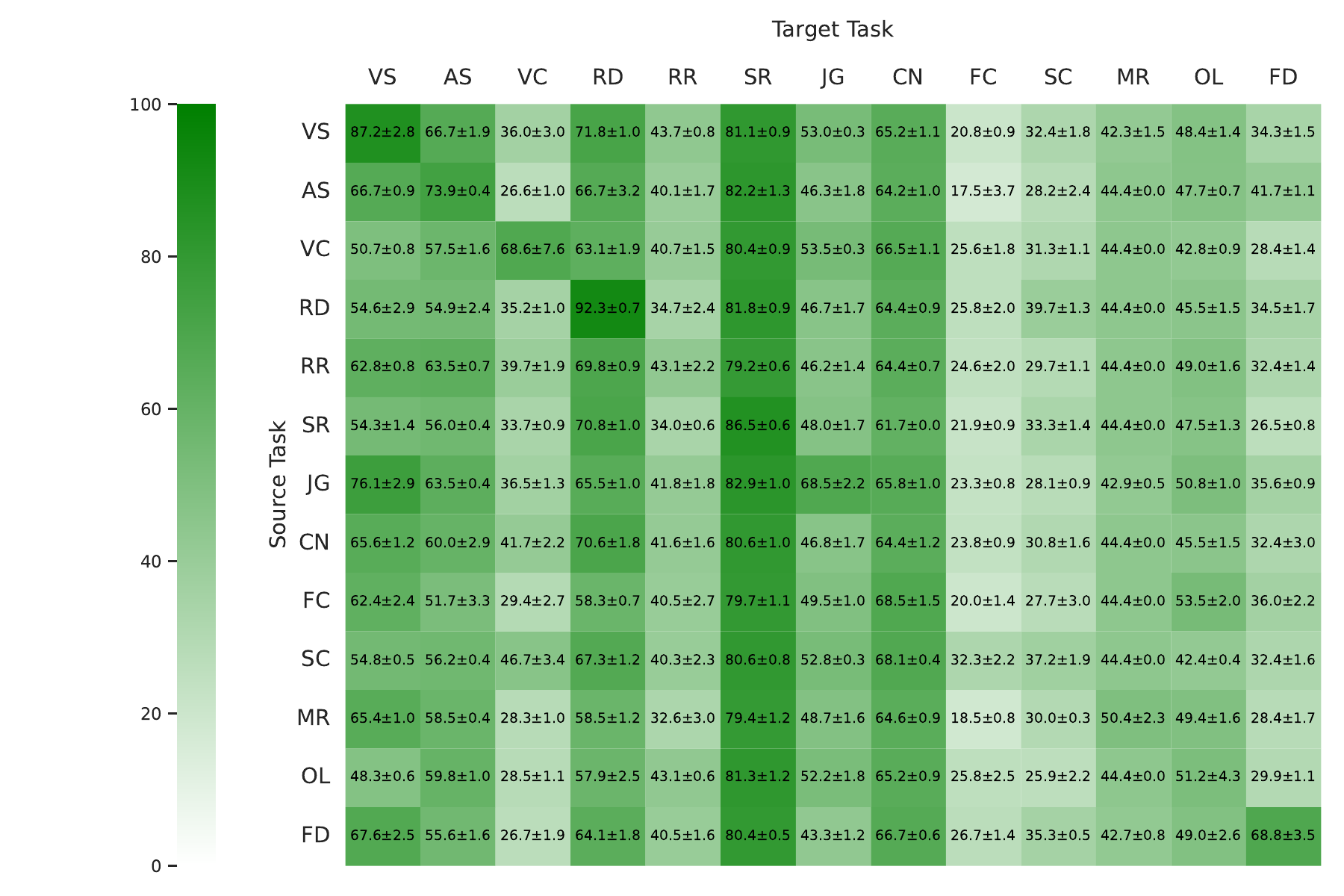}
    \caption{Accuracy Heatmap for Qwen-2.5-VL 3B.}
    \label{fig:acc_detail_3}
\end{figure*}

\begin{figure*}[!htbp]
    \centering
    \includegraphics[width=0.85\linewidth]{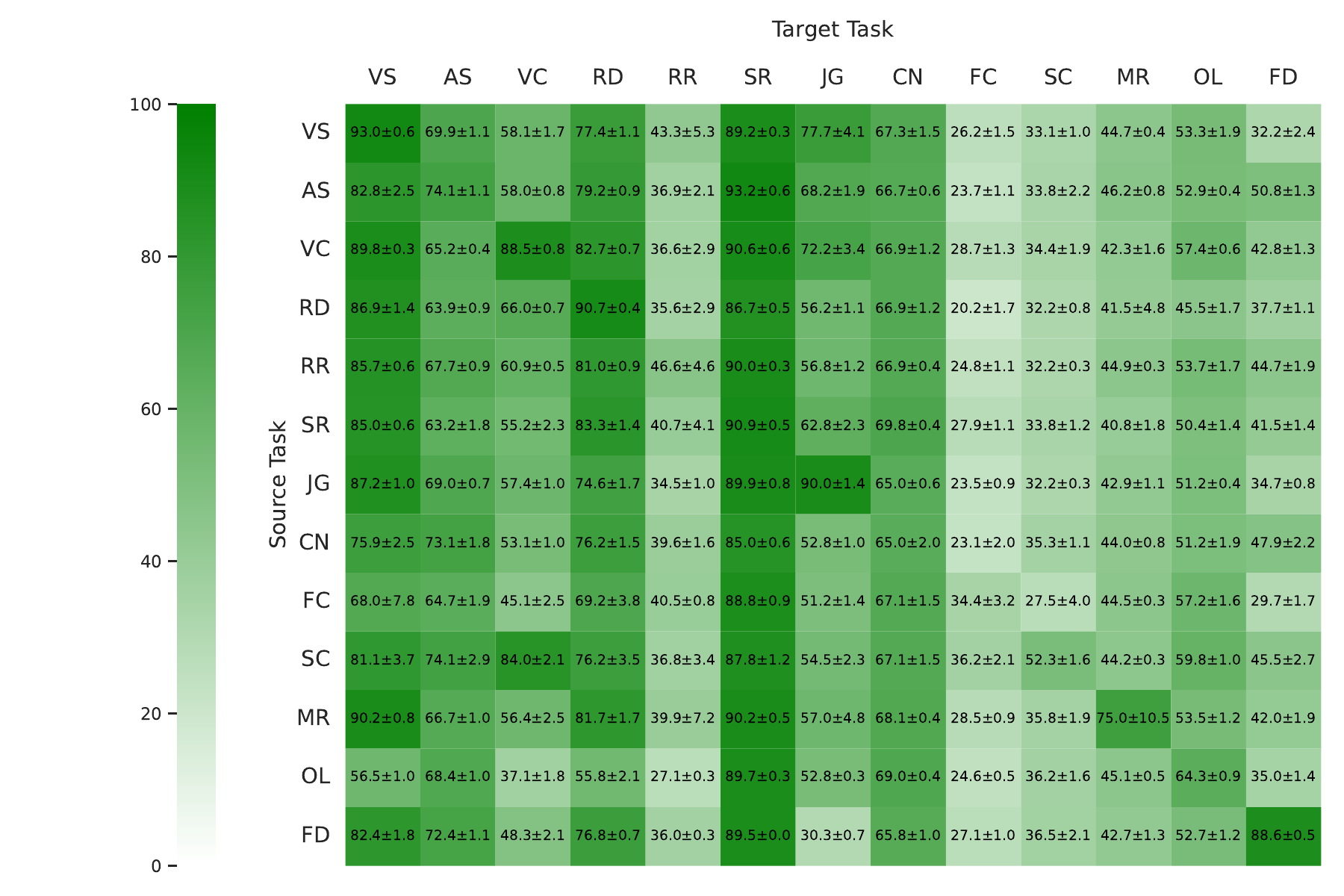}
    \caption{Accuracy Heatmap for Qwen-2.5-VL 7B.}
    \label{fig:acc_detail_7}
\end{figure*}

\begin{figure*}[!htbp]
    \centering
    \includegraphics[width=0.85\linewidth]{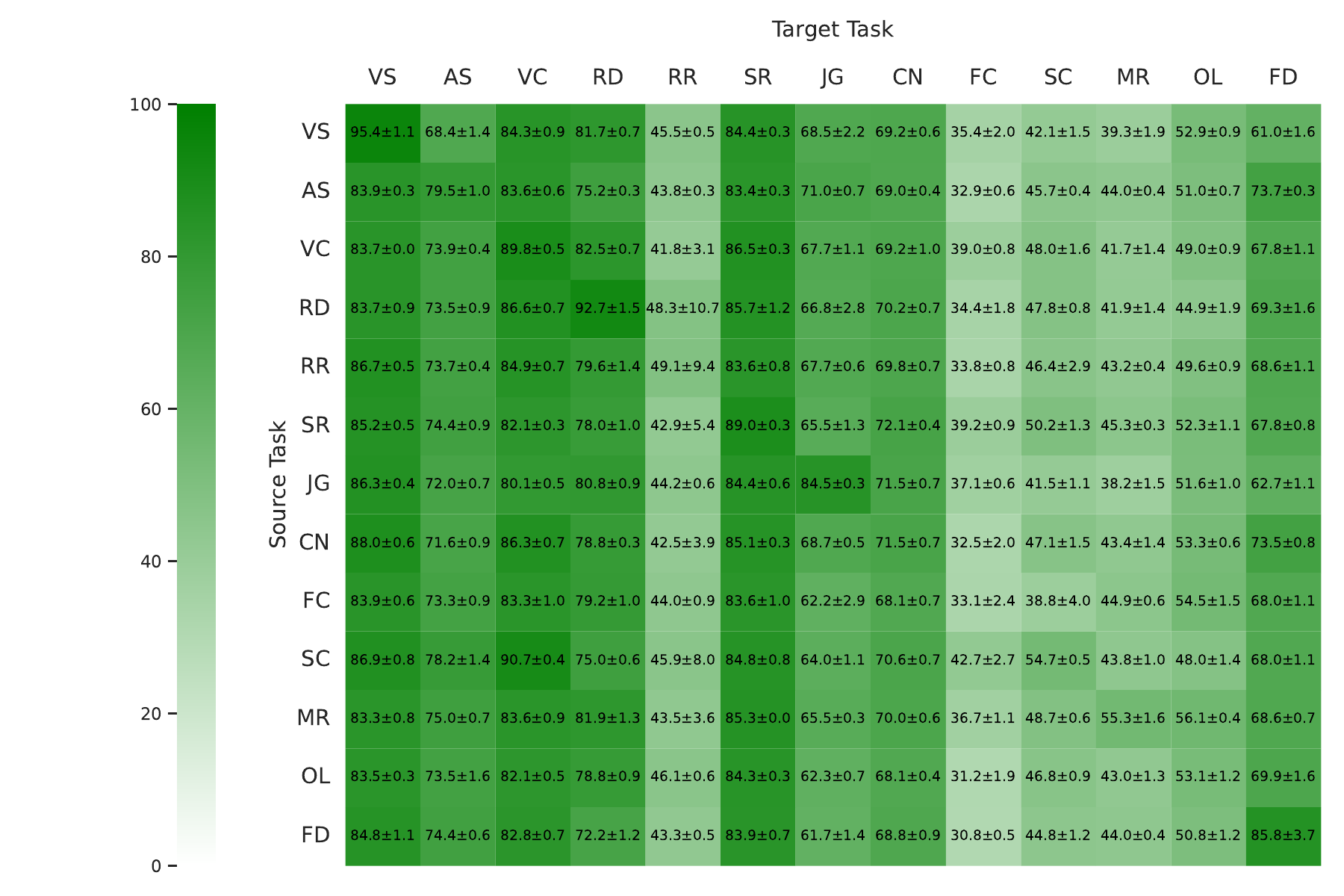}
    \caption{Accuracy Heatmap for Qwen-2.5-VL 32B.}
    \label{fig:acc_detail_32}
\end{figure*}

We also include accuracy heatmaps for all 13 tasks, reporting both the mean and standard deviation, in Figure~\ref{fig:acc_detail_3}, Figure~\ref{fig:acc_detail_7}, and Figure~\ref{fig:acc_detail_32} for the 3B, 7B, and 32B models, respectively. These summaries highlight how performance varies across tasks and model scales, and the accompanying standard deviations indicate the degree of variability in the underlying measurements.

\section{Task Category Trends}
\label{sec:task_cat}
\begin{figure*}[!htbp]
    \centering
    \includegraphics[width=\linewidth]{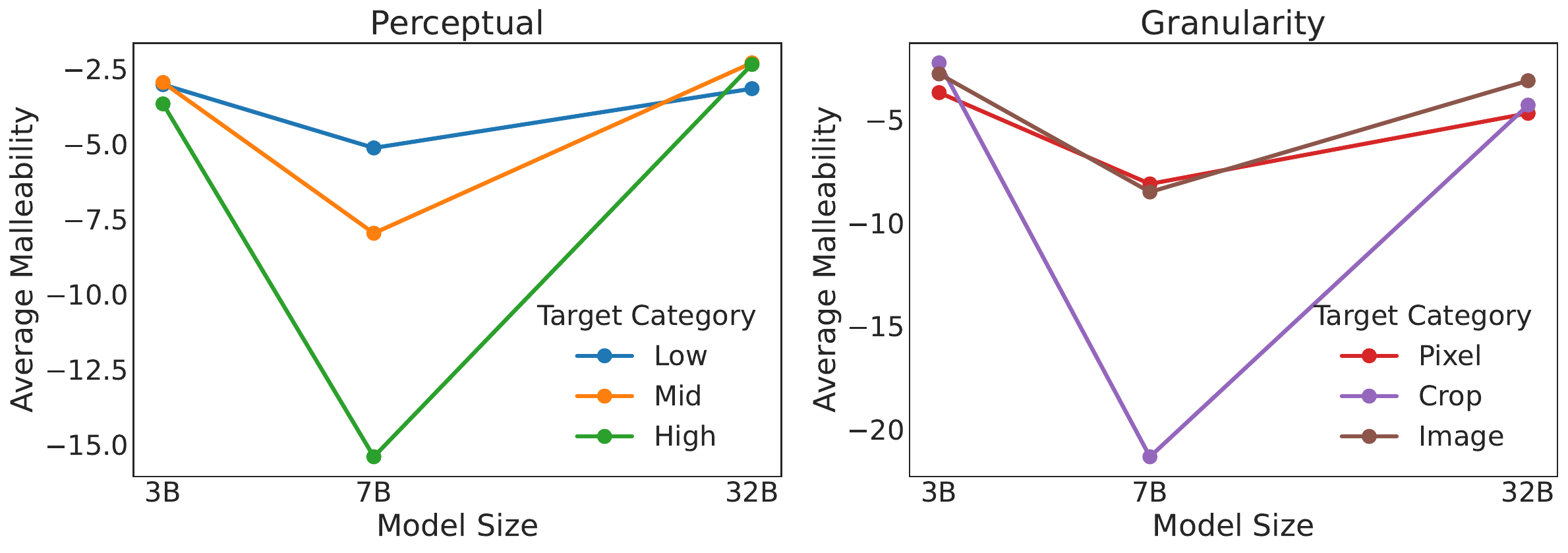}
    \caption{Average negative malleability trends across granular and perceptual levels.}
    \label{fig:group-malleability-neg}
\end{figure*}

\begin{figure*}[!htbp]
    \centering
    \includegraphics[width=\linewidth]{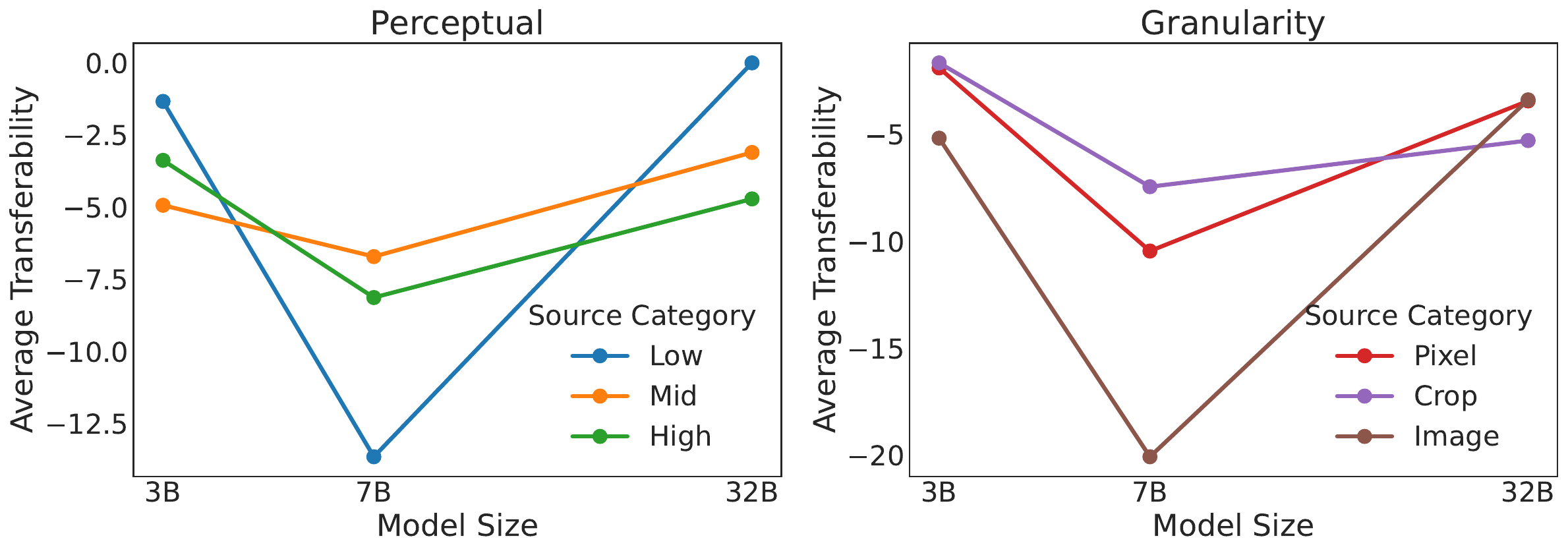}
    \caption{Average negative transferability trends across granular and perceptual levels.}
    \label{fig:group-transfer-neg}
\end{figure*}

\begin{figure*}[!htbp]
    \centering
    \includegraphics[width=0.8\linewidth]{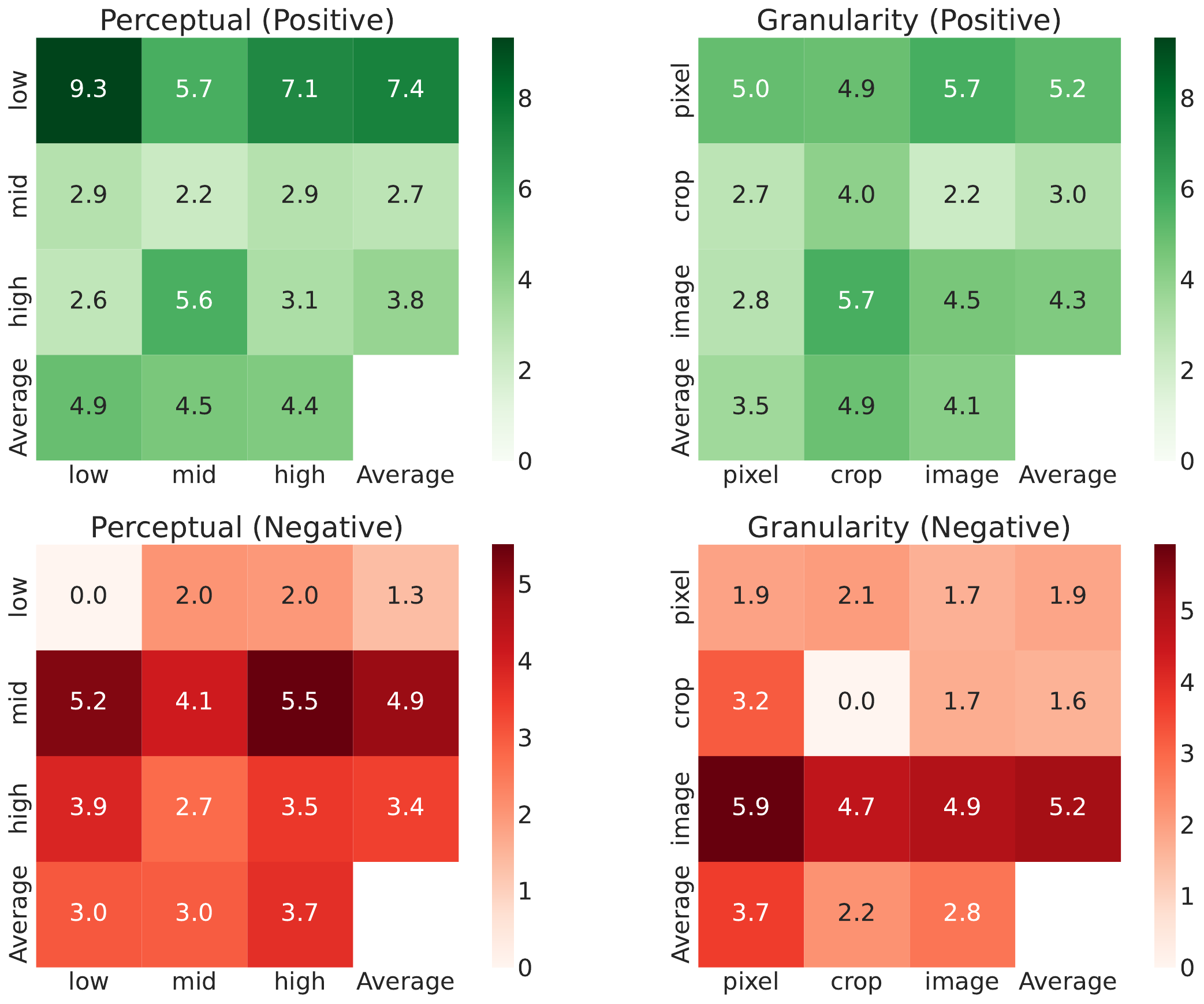}
    \caption{Qwen2.5-VL 3B category wise heatmaps}
    \label{fig:cat_3B}
\end{figure*}

\begin{figure*}[!htbp]
    \centering
    \includegraphics[width=0.8\linewidth]{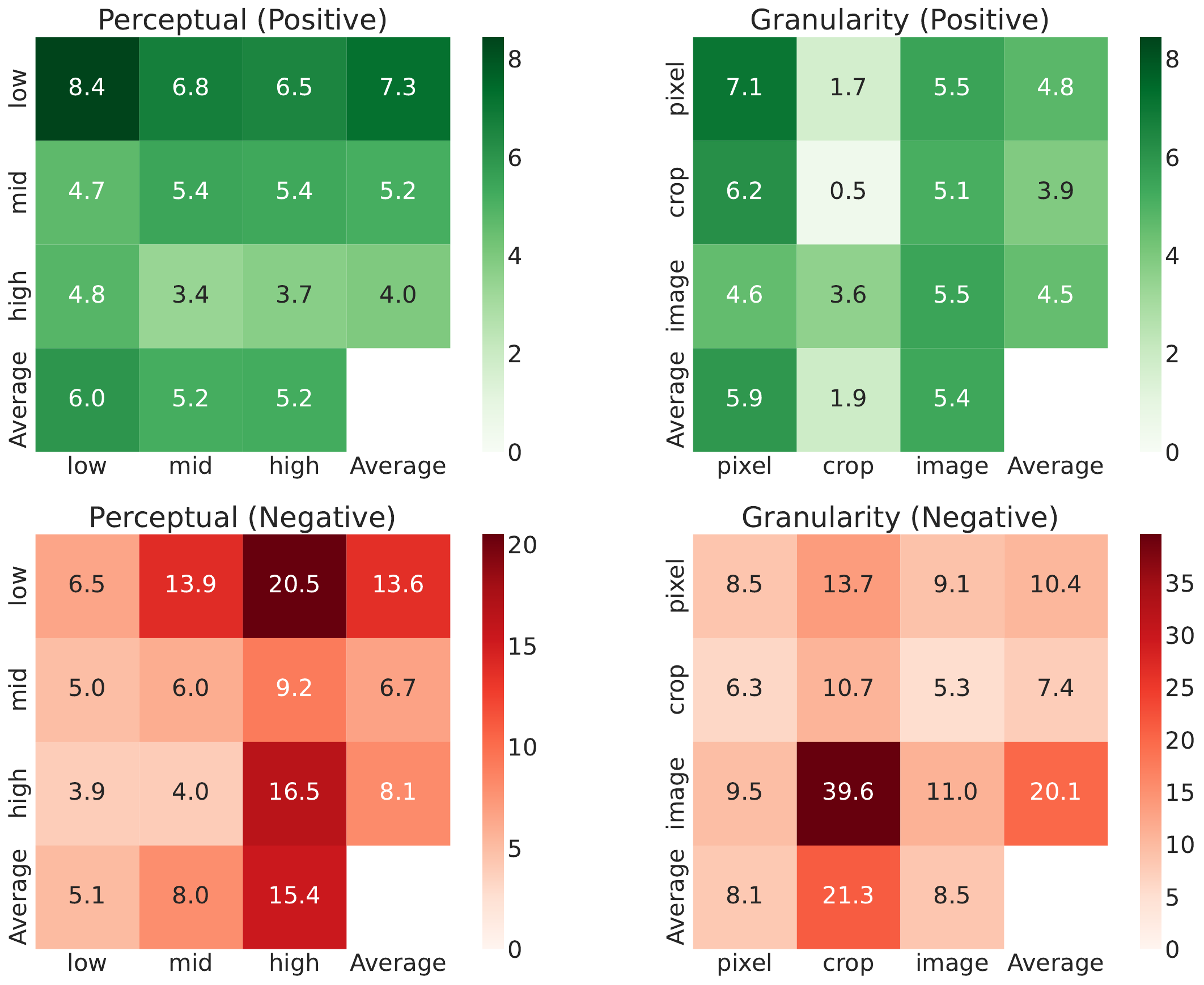}
    \caption{Qwen2.5-VL 7B category wise heatmaps}
    \label{fig:cat_7B}
\end{figure*}

\begin{figure*}[!htbp]
    \centering
    \includegraphics[width=0.8\linewidth]{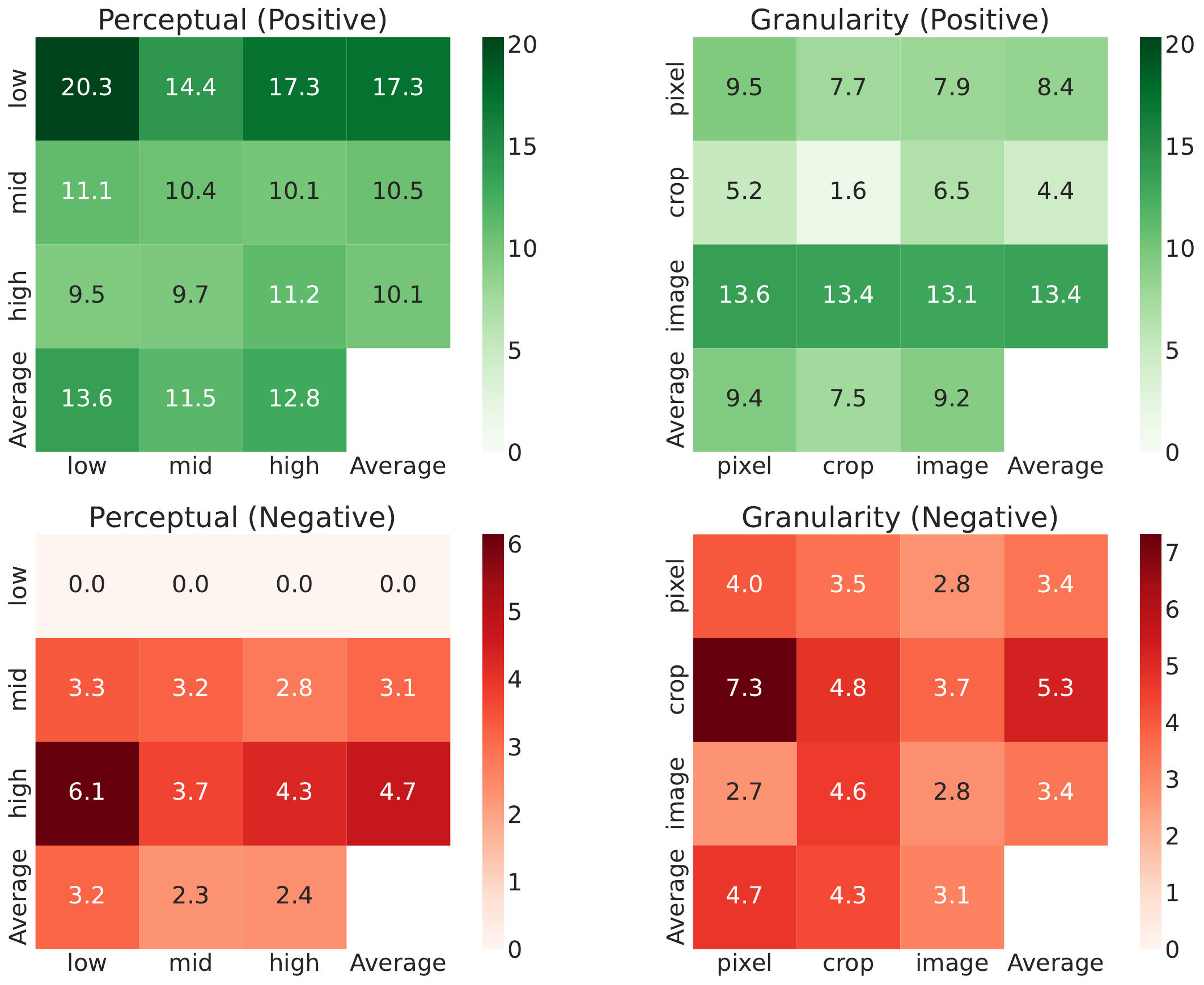}
    \caption{Qwen2.5-VL 32B category wise heatmaps}
    \label{fig:cat_32B}
\end{figure*}

In Figure~\ref{fig:group-transfer-neg} and Figure~\ref{fig:group-malleability-neg}, we plot the negative transferability and negative malleability respectively. Unlike the positive trends, we observe a sharp negative transferability and malleability in Qwen2.5-VL-7B model. On an average across models, low-level and image-level tasks exhibit the highest magnitude of negative transferability. High-level and crop-level tasks exhibit the highest magnitude of negative malleability. Additionally, we provide the heatmaps for transferability and malleability across all the task categories in Figure~\ref{fig:cat_3B}, Figure~\ref{fig:cat_7B} and Figure~\ref{fig:cat_32B}, for model sizes 3B, 7B and 32B respectively.

% In Figure~\ref{fig:cat_32B}, ~\ref{fig:cat_7B}, ~\ref{fig:cat_3B}, we provide detailed category-wise transferability heatmaps.

% Consistent with the positive trends in the main paper, we observe that average transferability and malleability increases with model size. Figure~\ref{fig:group-malleability-neg} demonstrates that low-level tasks are the most malleable on average. Additionally, we note that both pixel and image level tasks exhibit similar average malleability levels. Similarly, low-level and image-level tasks are highly transferable.

\section{Cliques across Model Sizes}
\label{sec:cliques}

\begin{figure*}[!htbp]
    \centering
    \begin{subfigure}{0.48\linewidth}
        \centering
        \includegraphics[width=\linewidth]{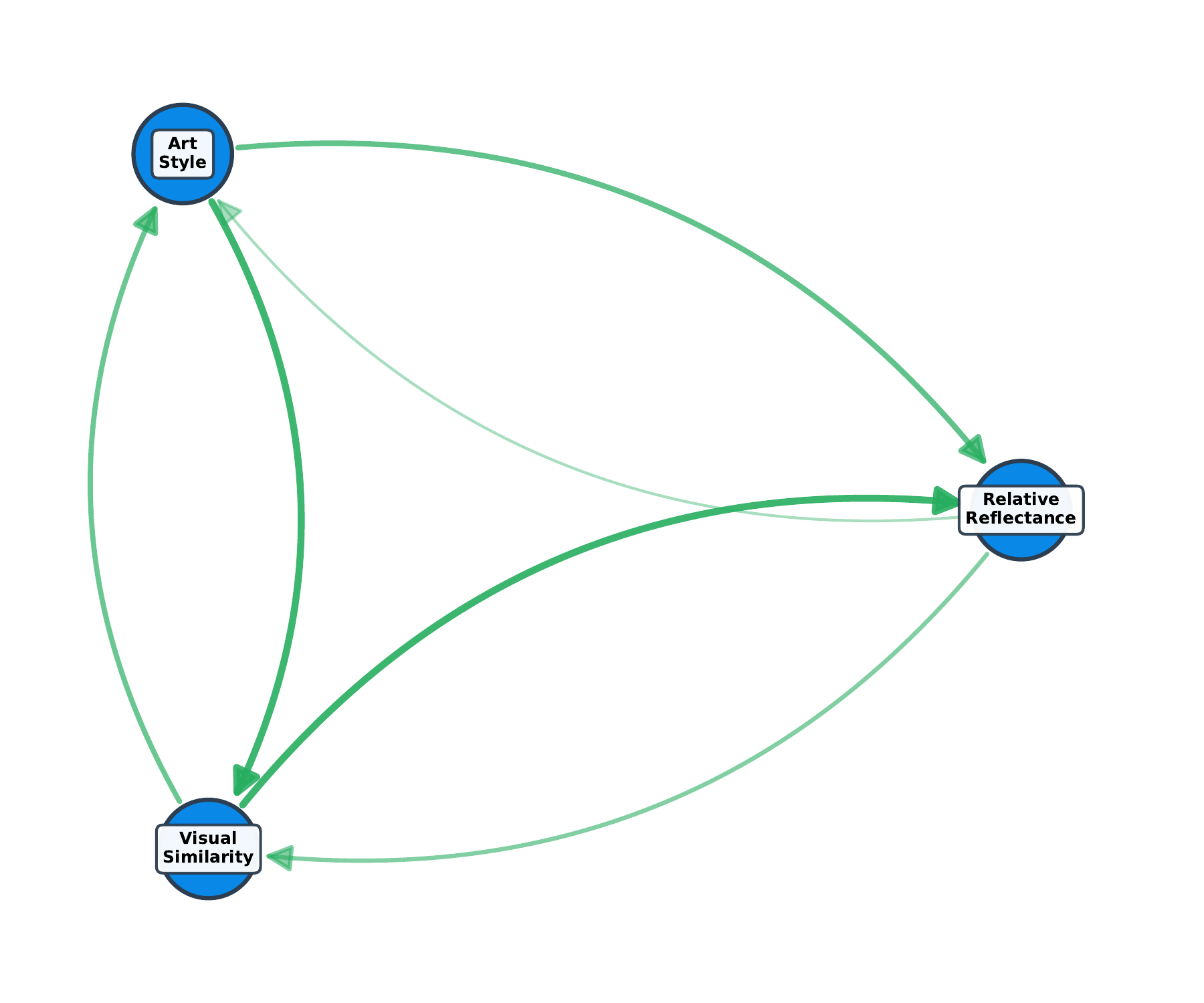}
        \caption{Positive Clique}
    \end{subfigure}
    \hfill
    \begin{subfigure}{0.48\linewidth}
        \centering
        \includegraphics[width=\linewidth]{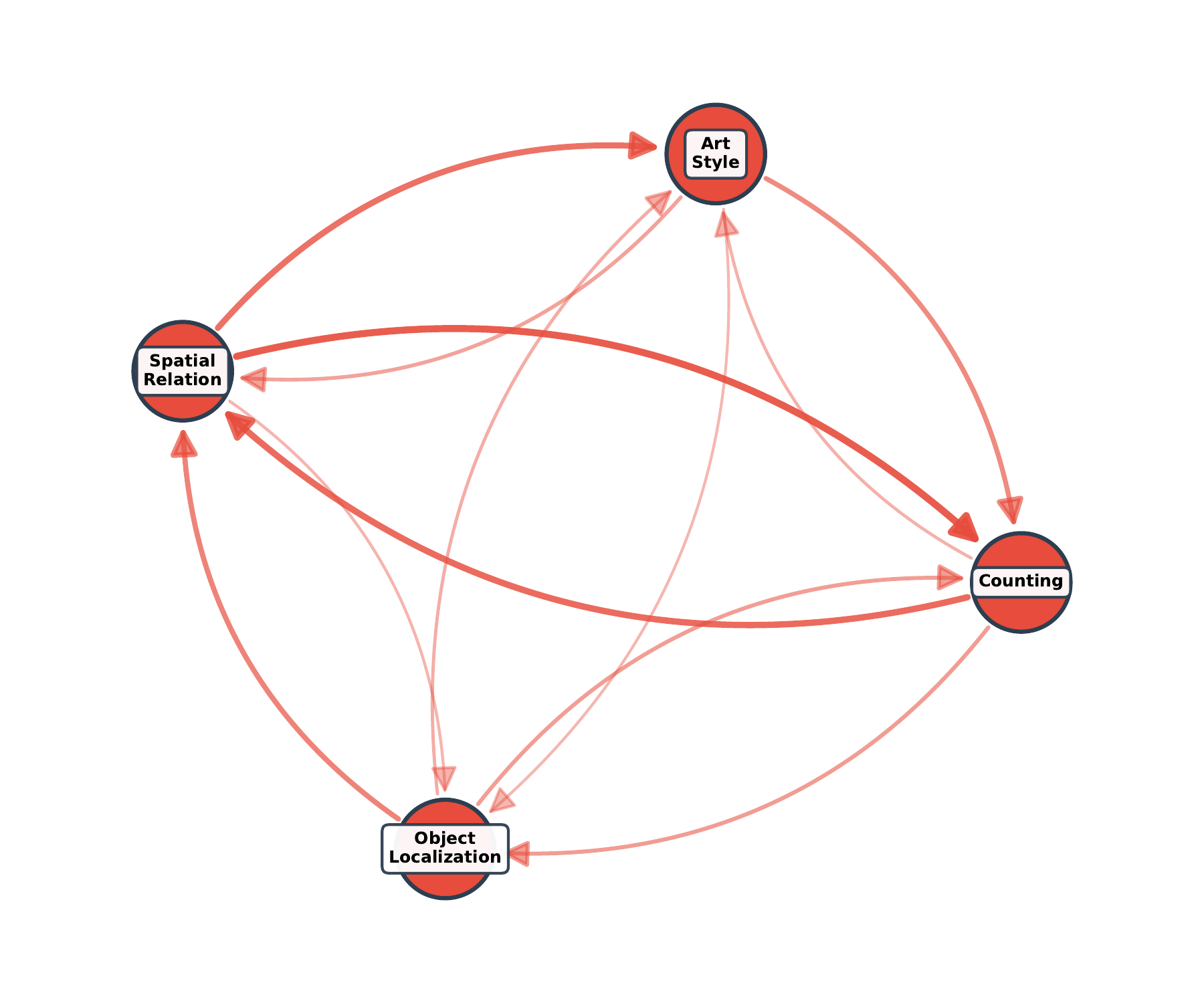}
        \caption{Negative Clique}
    \end{subfigure}
    \caption{Largest (a) positive and (b) negative clique for Qwen-2.5-VL 3B.}
    \label{fig:cliq_3}
\end{figure*}

\begin{figure*}[!htbp]
    \centering
    \begin{subfigure}{0.48\linewidth}
        \centering
        \includegraphics[width=\linewidth]{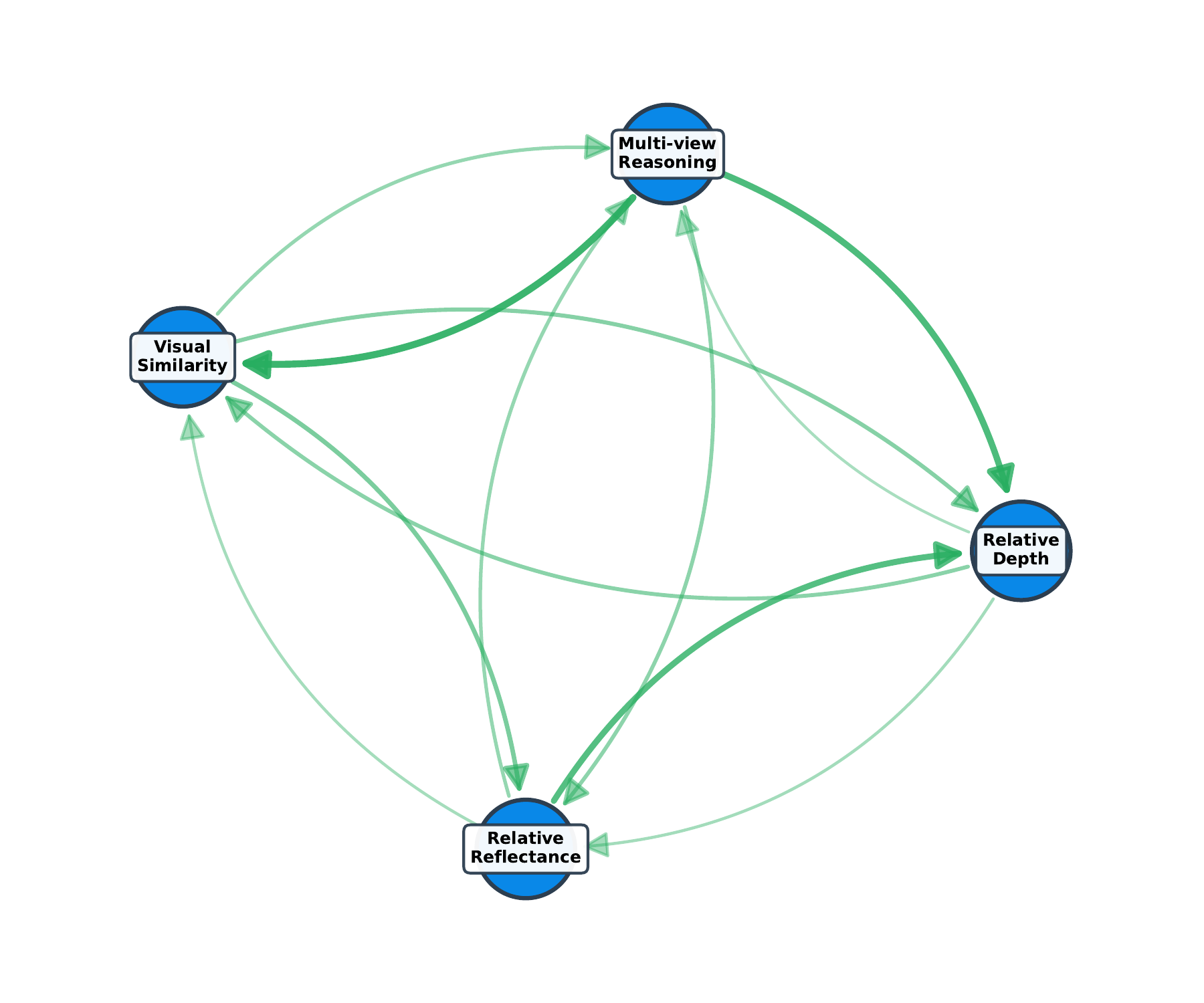}
        \caption{Positive Clique}
    \end{subfigure}
    \hfill
    \begin{subfigure}{0.48\linewidth}
        \centering
        \includegraphics[width=\linewidth]{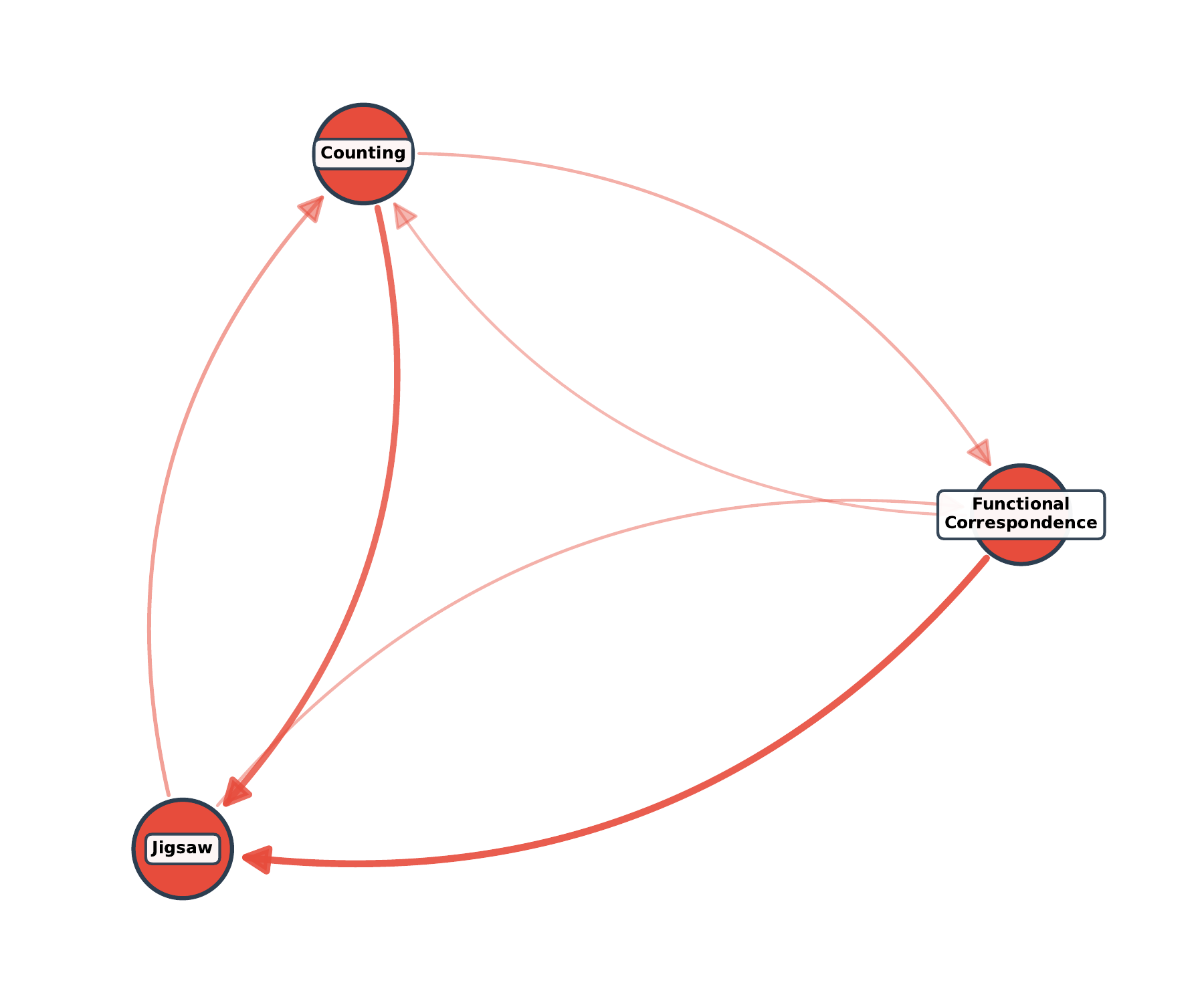}
        \caption{Negative Clique}
    \end{subfigure}
    \caption{Largest (a) positive and (b) negative clique for Qwen-2.5-VL 7B.}
    \label{fig:cliq_7}
\end{figure*}

\begin{figure*}[!htbp]
    \centering
    \begin{subfigure}{0.48\linewidth}
        \centering
        \includegraphics[width=\linewidth]{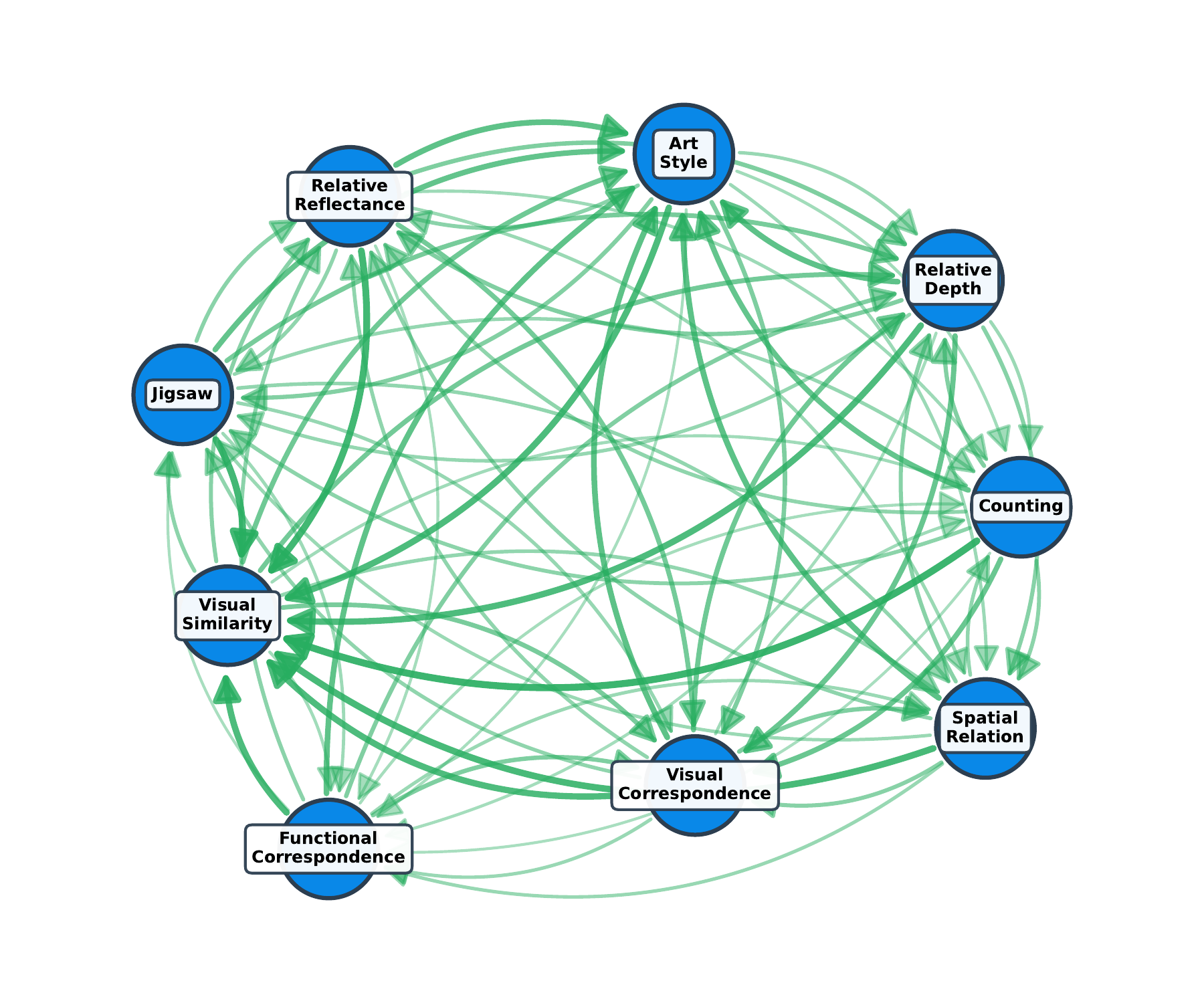}
        \caption{Positive Clique}
    \end{subfigure}
    \hfill
    \begin{subfigure}{0.48\linewidth}
        \centering
        \includegraphics[width=\linewidth]{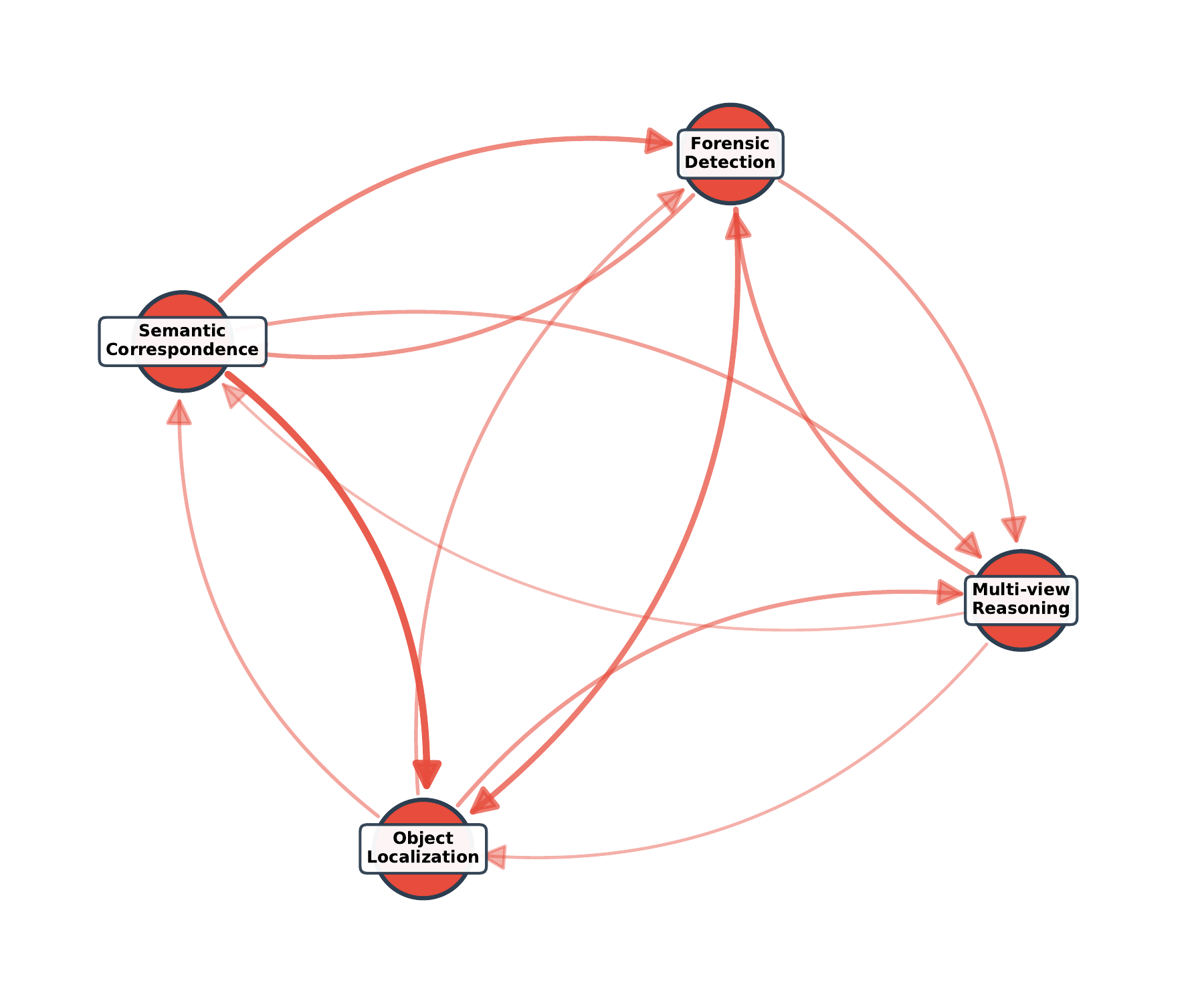}
        \caption{Negative Clique}
    \end{subfigure}
    \caption{Largest (a) positive and (b) negative clique for Qwen-2.5-VL 32B.}
    \label{fig:cliq_32}
\end{figure*}

Table~\ref{tab:all_cliques_list} lists the positive and negative cliques identified across all three model sizes. In addition, Figure~\ref{fig:cliq_3}, Figure~\ref{fig:cliq_7}, and Figure~\ref{fig:cliq_32} visualize the largest positive and negative clique for the 3B, 7B, and 32B models, respectively. We note that 32B variant has the largest positive clique of size 9.

\begin{table*}[!htbp]
\centering
\resizebox{\linewidth}{!}{%
\begin{tabular}{ccl}
\toprule
\textbf{Clique Type} & \textbf{Model Size} & \textbf{Cliques} \\
\midrule
\multirow{5}{*}{Positive} 
    & \multirow{2}{*}{3B} 
    & $\{$AS, RR, VS$\}$, $\{$MR, RR, VS$\}$, $\{$RD, RR, VC$\}$, $\{$FC, RD, RR$\}$, \\
    & 
    & $\{$MR, RD, RR$\}$, $\{$RD, SC, VC$\}$, $\{$FC, JG, OL$\}$ \\
\cmidrule{2-3}
    & \multirow{2}{*}{7B} 
    & $\{$MR, RD, RR, VS$\}$, $\{$MR, RR, SR$\}$, $\{$AS, MR, RR$\}$, \\
    &
    & $\{$AS, MR, OL$\}$, $\{$CN, MR, OL$\}$ \\

\cmidrule{2-3}
    & 32B
    & $\{$AS, CN, FC, JG, RD, RR, SR, VC, VS$\}$, $\{$AS, CN, FD$\}$ \\

\midrule

\multirow{5}{*}{Negative} 
    & \multirow{2}{*}{3B} 
    & $\{$AS, CN, OL, SR$\}$, $\{$CN, FD, OL, SR$\}$, $\{$CN, FD, JG, SR$\}$, $\{$OL, SR, VS$\}$, \\
    & 
    & $\{$AS, FC, SR$\}$, $\{$AS, OL, SC$\}$, $\{$AS, OL, VC$\}$, $\{$FD, OL, VC$\}$ \\
\cmidrule{2-3}
    & \multirow{2}{*}{7B} 
    & $\{$CN, FC, JG $\}$, $\{$FD, JG, SC $\}$, $\{$FD, SC, VS$\}$, \\
    &
    & $\{$CN, SC, VS$\}$, $\{$CN, JG, SC$\}$ \\

\cmidrule{2-3}
    & 32B
    & $\{$FD, MR, OL, SC$\}$ \\

\midrule
\bottomrule
\end{tabular}%
}
\caption{List of all positive and negatives cliques for all model sizes (3B, 7B, 32B) for Qwen-2.5-VL.}
\label{tab:all_cliques_list}
\end{table*}

\section{Implementation Details}
\label{sec:implement-det}

All finetuning experiments are performed on 8xA100 GPUs 40GB using the opensource Qwen repository\footnote{\url{https://github.com/QwenLM/Qwen3-VL}}. DeepSpeed~\citep{ds} ZeRO-2 is used for Qwen-2.5-VL 3B and 7B, while DeepSpeed~\citep{ds} ZeRO-3 is used for Qwen-2.5-VL 32B, all with mixed-precision. Batch size is set to 16, weight decay as 0 and warmup ratio of 0.03 with cosine decay learning rate scheduler. For finetuning, LoRa rank is set to 8 and $\alpha$ is set to 16 for all tasks. Task-wise training details are mentioned in Table \ref{tab:task-overview}.
We utilize the GPT-4.1 model for extracting responses from model responses and the evaluation is performed using the official code provided by the BLINK benchmark\footnote{\url{https://github.com/zeyofu/BLINK\_Benchmark}}.

\begin{table*}[!htbp]
\centering
\begin{tabular}{m{2.5cm} m{5cm} m{3.5cm} m{2.5cm}}
\toprule
\textbf{Task} & \textbf{Description} & \textbf{Source Dataset} & \textbf{Hyperparams} \\
\midrule
Visual Similarity & \textit{Given a reference image alongside two alternatives, identify the image most visually similar to the reference.} & DreamSim (Nights)~\citep{fu2023dreamsim} & 15,914 examples, \newline2500 steps, \newline1e-4 lr \\
\hline
Counting & \textit{Given an image, a counting-related question, and 4 options, choose the correct answer.} & TallyQA~\citep{acharya2018tallyqaansweringcomplexcounting} & 250k examples,\newline 1 epoch,\newline 1e-4 lr \\
\hline
Relative Depth & \textit{Decide which of two specified points is closer.} & Depth in the Wild \newline+ Human Annotations~\citep{chen2016single} & 210k examples,\newline 1 epoch,\newline 1e-4 lr \\
\hline
Jigsaw & \textit{Choose the image that completes the scene.} & TARA~\citep{FZCVR22} & 11,837 examples,\newline 920 steps,\newline 1e-4 lr \\
\hline
Art Style & \textit{Given a reference painting and two candidate paintings, identify which shares the same art style.} & WikiArt\footnotemark & 100k examples,\newline 1000 steps,\newline 1e-4 lr \\
\hline
Functional \newline Correspondence & \textit{Match a reference point in one image with the best corresponding point among 4 options in another image, based on functional affordances.} & FunKPoint~\citep{lai2021functional} & 100k examples,\newline 2000 steps,\newline 1e-4 lr \\
\hline
Semantic \newline Correspondence & \textit{Given a point in a reference image, choose the most semantically similar point among 4 options in another image.} & Spair-71k~\citep{min2019spair71klargescalebenchmarksemantic} & 36k examples,\newline 5 epochs,\newline 1e-4 lr \\
\hline
Spatial Relation &\textit{ Identify the spatial relationship between objects in an image.} & Visual Spatial Reasoning~\citep{Liu2022VisualSR} & 7k examples,\newline 5 epochs,\newline 1e-4 lr \\
\hline
Object \newline Localization & \textit{Given an image and two bounding boxes (one ground-truth, one perturbed), choose the correct bounding box.} & LVIS~\citep{gupta2019lvisdatasetlargevocabulary} & 18,912 examples,\newline 1480 steps,\newline 1e-4 lr \\
\hline
Visual \newline Correspondence & \textit{Identify the same point across two input images. One image has 1 point, the other has 4 candidate points. }& HPatches~\citep{balntas2017hpatches} & 6k examples,\newline 10 epochs,\newline 1e-4 lr \\
\hline
Multi-view\newline Reasoning & \textit{Predict the direction of camera motion from two views.} & Wild 6D~\cite{fucategory} & 4k examples,\newline 10 epochs,\newline 1e-4 lr \\
\hline
Relative\newline Reflectance & \textit{Decide which of two pixels is darker, or whether they have similar reflectance.} & Intrinsic Images in the Wild + Human Annotations~\citep{bell2014intrinsic} & 14k examples,\newline 10 epochs,\newline 1e-4 lr \\
\hline
Forensic\newline Detection & \textit{Identify synthetic images from a mixture of real and synthetic samples.} & Synthetic: COCO captions~\citep{lin2015microsoftcococommonobjects} + Stable Diffusion XL \newline Real: COCO captions + Web search & 60,518 examples,\newline 100 steps,\newline 1e-4 lr \\
\bottomrule
\end{tabular}
\caption{Overview of tasks used in our evaluation. Each task is paired with its source dataset and finetuning setup. The number of examples, epochs/steps, and lr are specified for each task.}
\label{tab:task-overview}
\end{table*}

\footnotetext{\url{https://huggingface.co/datasets/keremberke/painting-style-classification}}

\section{Effect of Training Steps on PGF}
\label{sec:step}

\begin{figure*}[!htbp]
    \centering
    \includegraphics[width=0.85\linewidth]{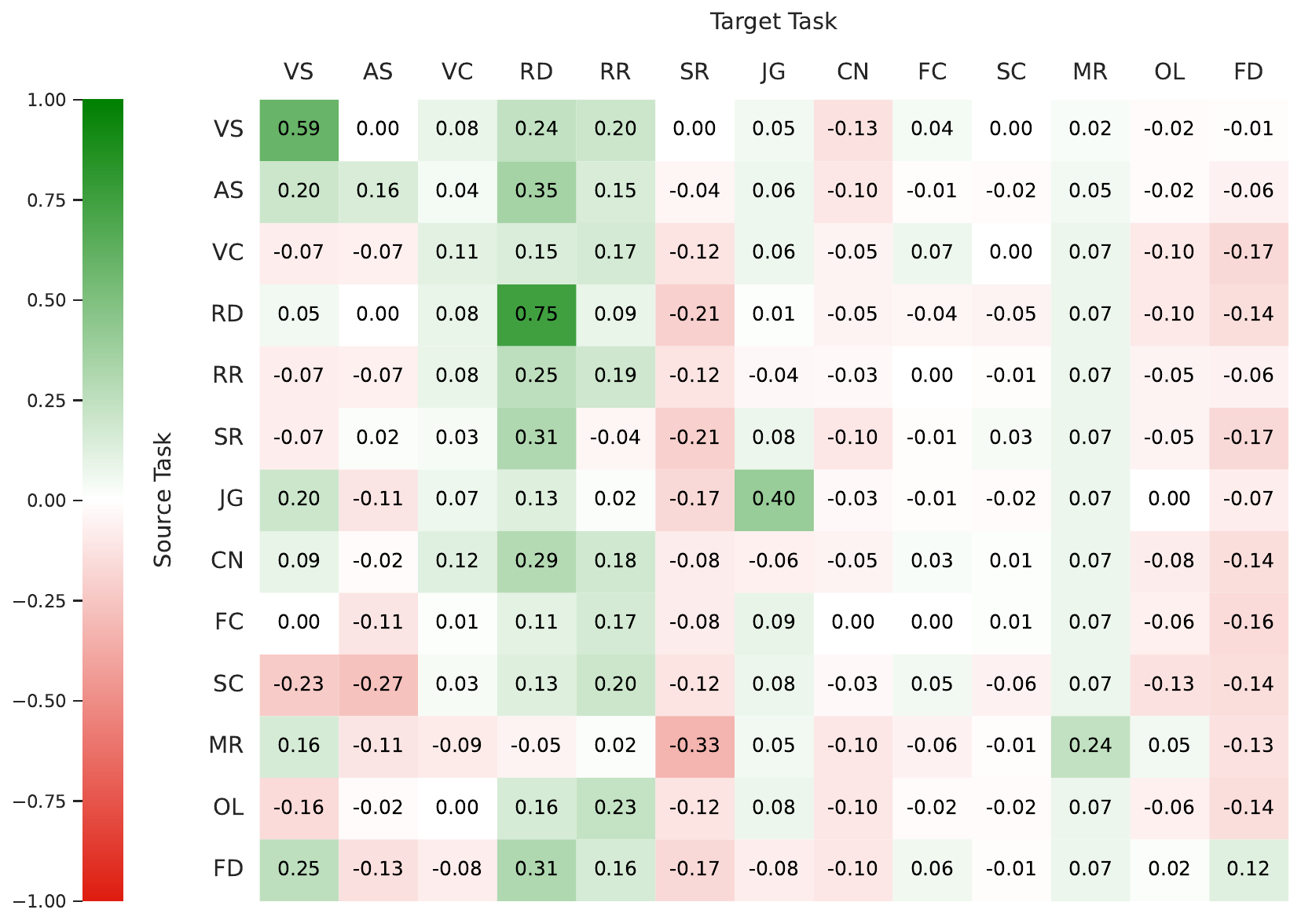}
    \caption{PGF Heatmap for Qwen-2.5-VL 3B trained on 25\% of the original training steps.}
    \label{fig:step-ablation-25}
\end{figure*}

\begin{figure*}[!htbp]
    \centering
    \includegraphics[width=0.82\linewidth]{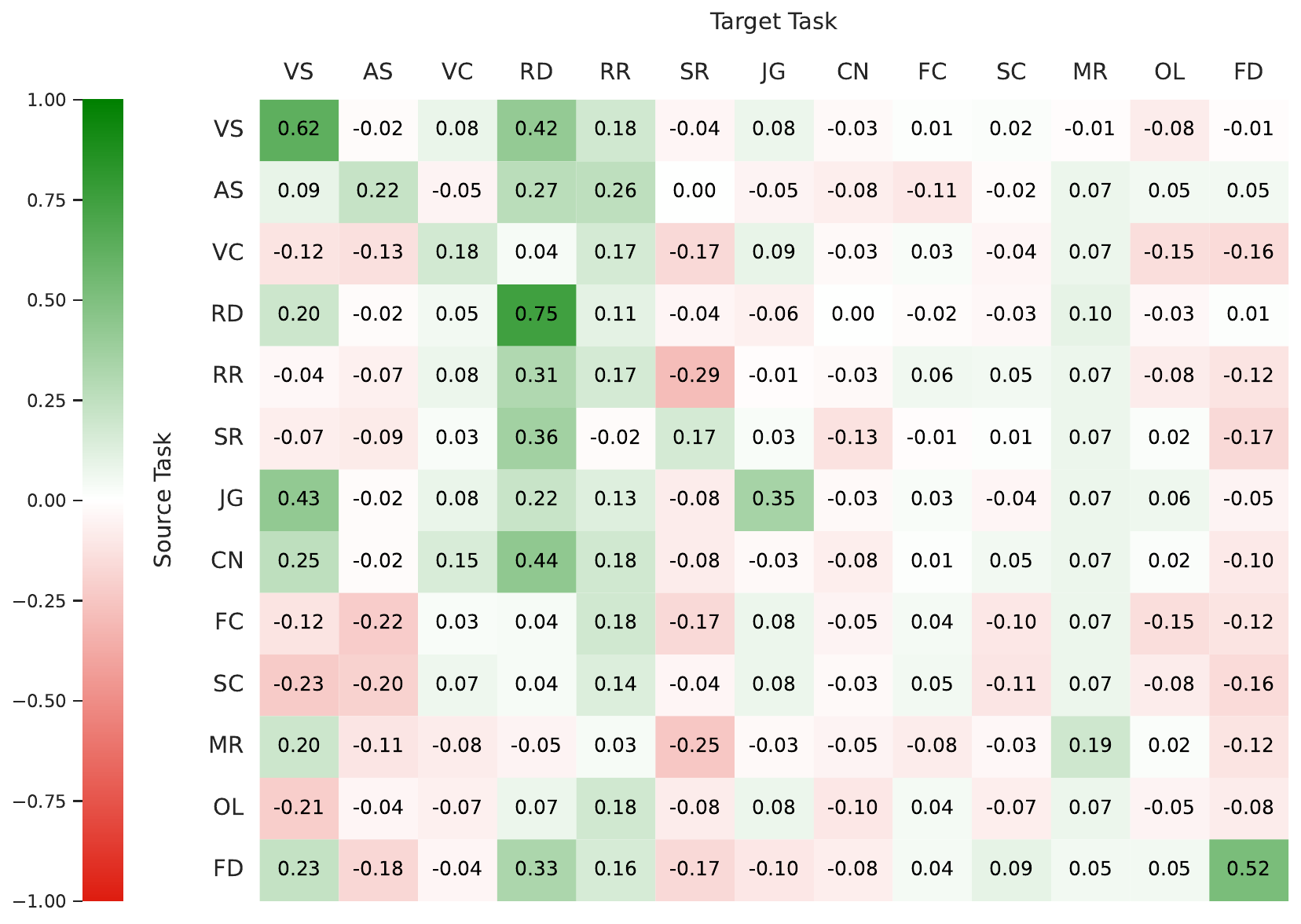}
    \caption{PGF Heatmap for Qwen-2.5-VL 3B trained on 50\% of the original training steps.}
    \label{fig:step-ablation-50}
\end{figure*}

\begin{figure*}[!htbp]
    \centering
    \includegraphics[width=0.82\linewidth]{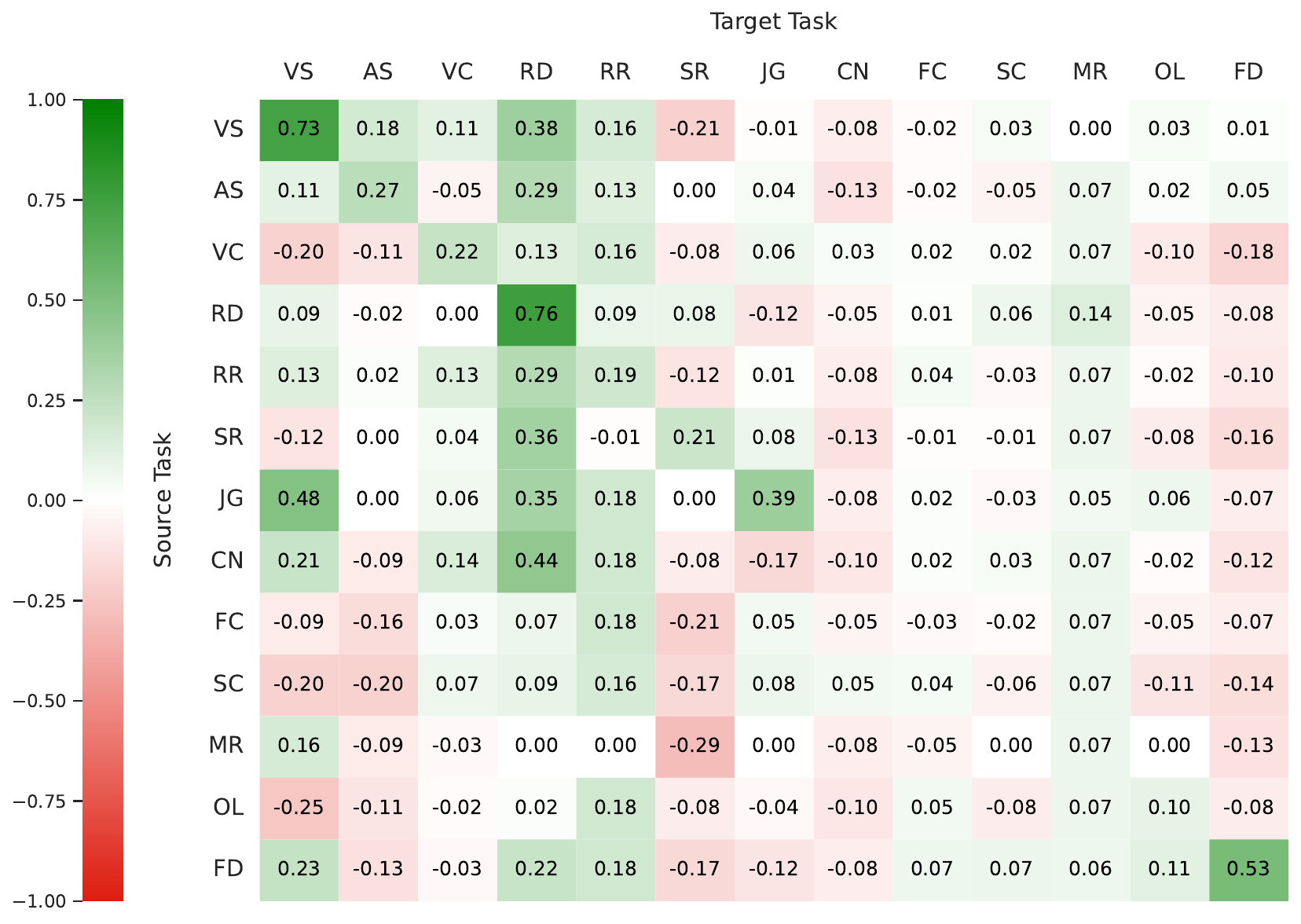}
    \caption{PGF Heatmap for Qwen-2.5-VL 3B trained on 75\% of the original training steps.}
    \label{fig:step-ablation-75}
\end{figure*}

In Figures~\ref{fig:step-ablation-25}, Figure~\ref{fig:step-ablation-50} and Figure~\ref{fig:step-ablation-75}, we examine the impact of finetuning steps on transferability using Qwen2.5-VL-3B. These heatmaps show that with increasing number of steps, average transferability increases monotonically. This behavior is expected as additional optimization amplifies the model’s deviation from the original checkpoint, strengthening transfer signals. While the absolute PGF values change with training duration, the qualitative structure of the transfer patterns remains stable across ablations, indicating that the relationships among tasks are largely preserved even under longer finetuning.

\section{LoRA Weights Analysis}
\label{sec:lora}

\begin{figure*}[!htbp]
    \centering
    \includegraphics[width=0.65\textwidth]{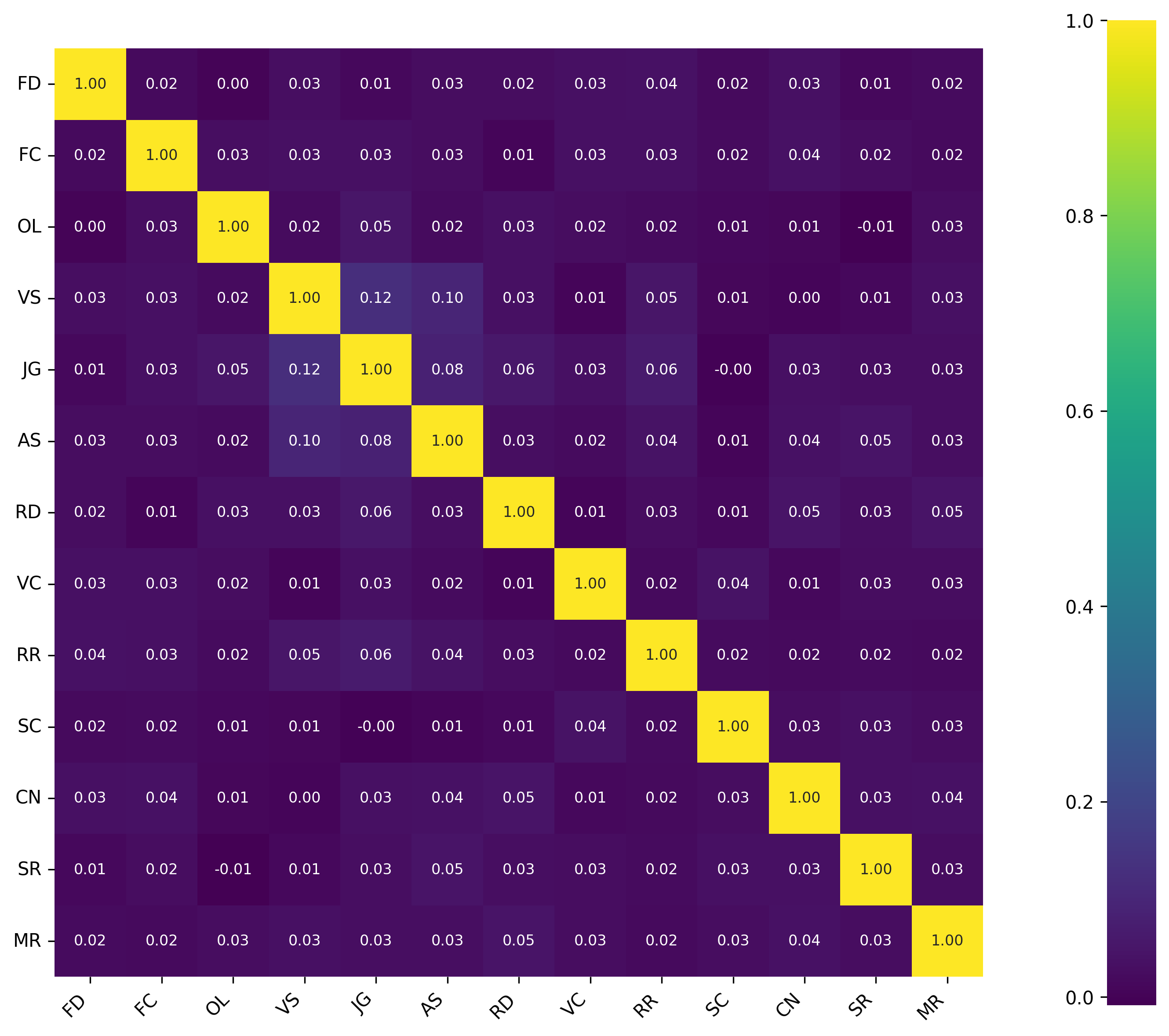}
    \caption{Cosine Similarity of LoRA weights of the output projection from layer 35 (last layer) after finetuning Qwen2.5VL-3B.}
    \label{fig:lora-3B}
\end{figure*}

\begin{figure*}[!htbp]
    \centering
    \includegraphics[width=0.65\textwidth]{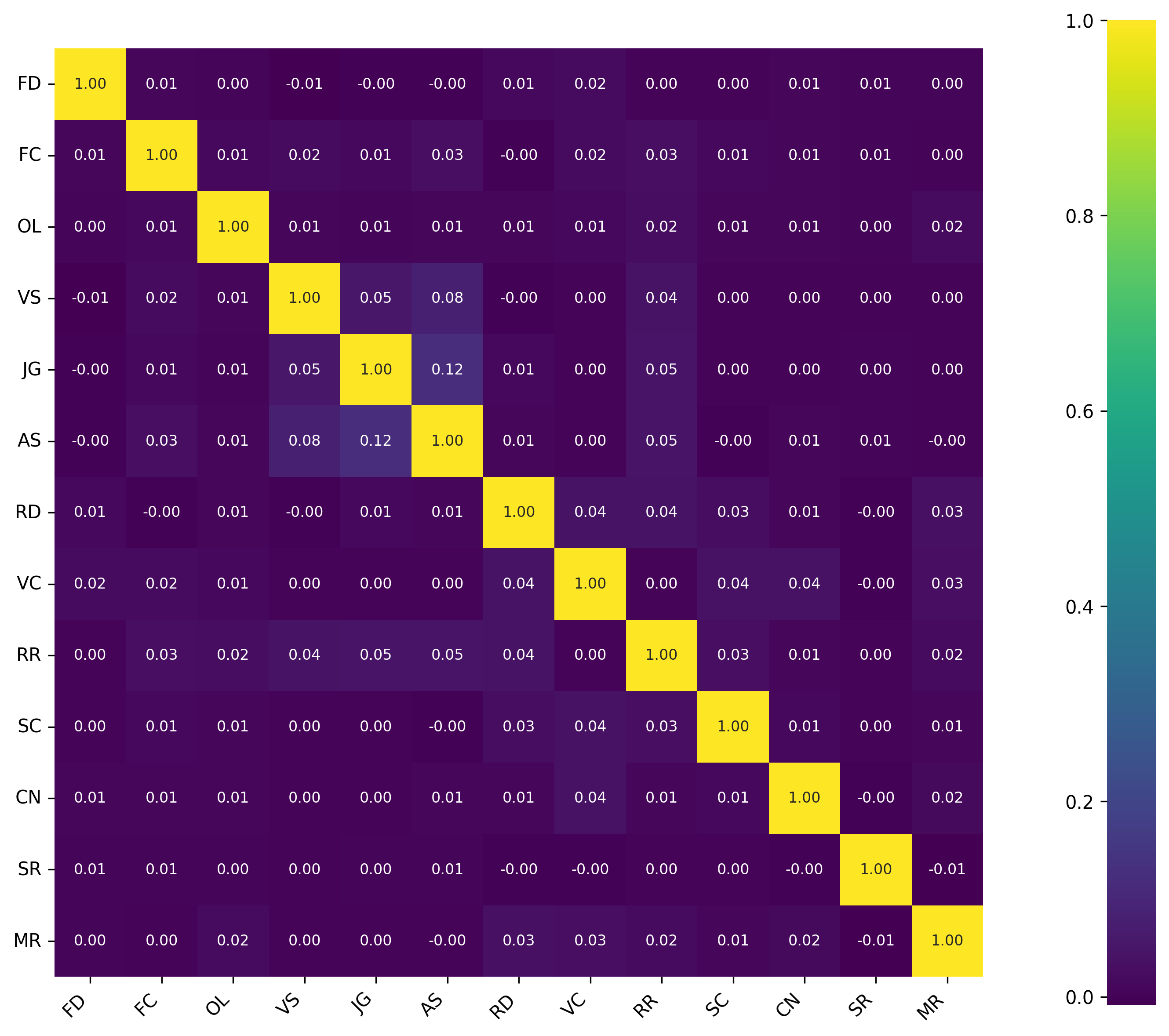}
    \caption{Cosine Similarity of LoRA weights of the output projection from layer 27 (last layer) after finetuning Qwen2.5VL-7B.}
    \label{fig:lora-7B}
\end{figure*}

\begin{figure*}[!htbp]
    \centering
    \includegraphics[width=0.65\textwidth]{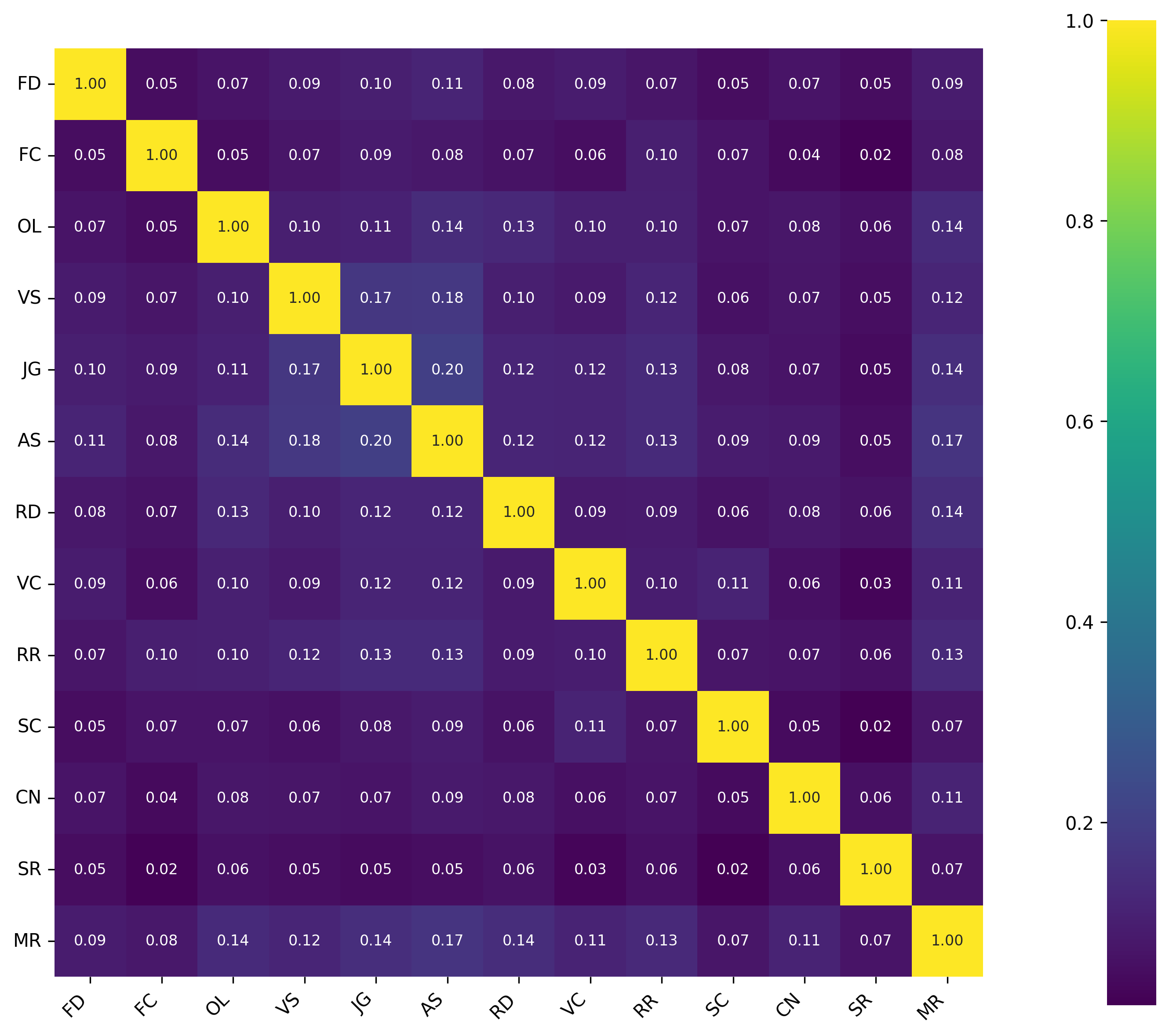}
    \caption{Cosine Similarity of LoRA weights of the output projection from layer 65 (last layer) after finetuning Qwen2.5VL-32B.}
    \label{fig:lora-32B}
\end{figure*}

In this section, we analyze the cosine similarity of LoRA-finetuned weights across tasks to assess whether certain tasks induce more similar parameter updates, thereby revealing shared structure or transferable representations. For this analysis, we focus on the output projection weights from the final layer, as they exhibited the highest variance across all the layers. Figure~\ref{fig:lora-3B}, Figure~\ref{fig:lora-7B} and Figure~\ref{fig:lora-32B}, show the resulting heatmaps for Qwen2.5-VL 3B, 7B, and 32B, respectively. Across all models, the strongest similarities appear among the Visual Similarity, Jigsaw, and Art Style tasks. We hypothesize that this arises because these are multi-image tasks, requiring comparable skills such as reasoning over pairs of images, assessing similarity, or aligning image composition. Consistent with the model-size trend, the 32B model exhibits the highest overall cosine similarity, suggesting stronger cross-task alignment in larger models. Interestingly, the 3B model shows higher similarities than the 7B model, which may be attributable to architectural differences: the 3B variant has 35 layers, whereas the 7B has 27 wider layers. A deeper interpretability analysis of these task-induced representations remains an avenue for future work.

\section{Generalization to Other Models}
\label{sec:llava}

\begin{figure*}[!htbp]
    \centering
    \includegraphics[width=0.85\linewidth]{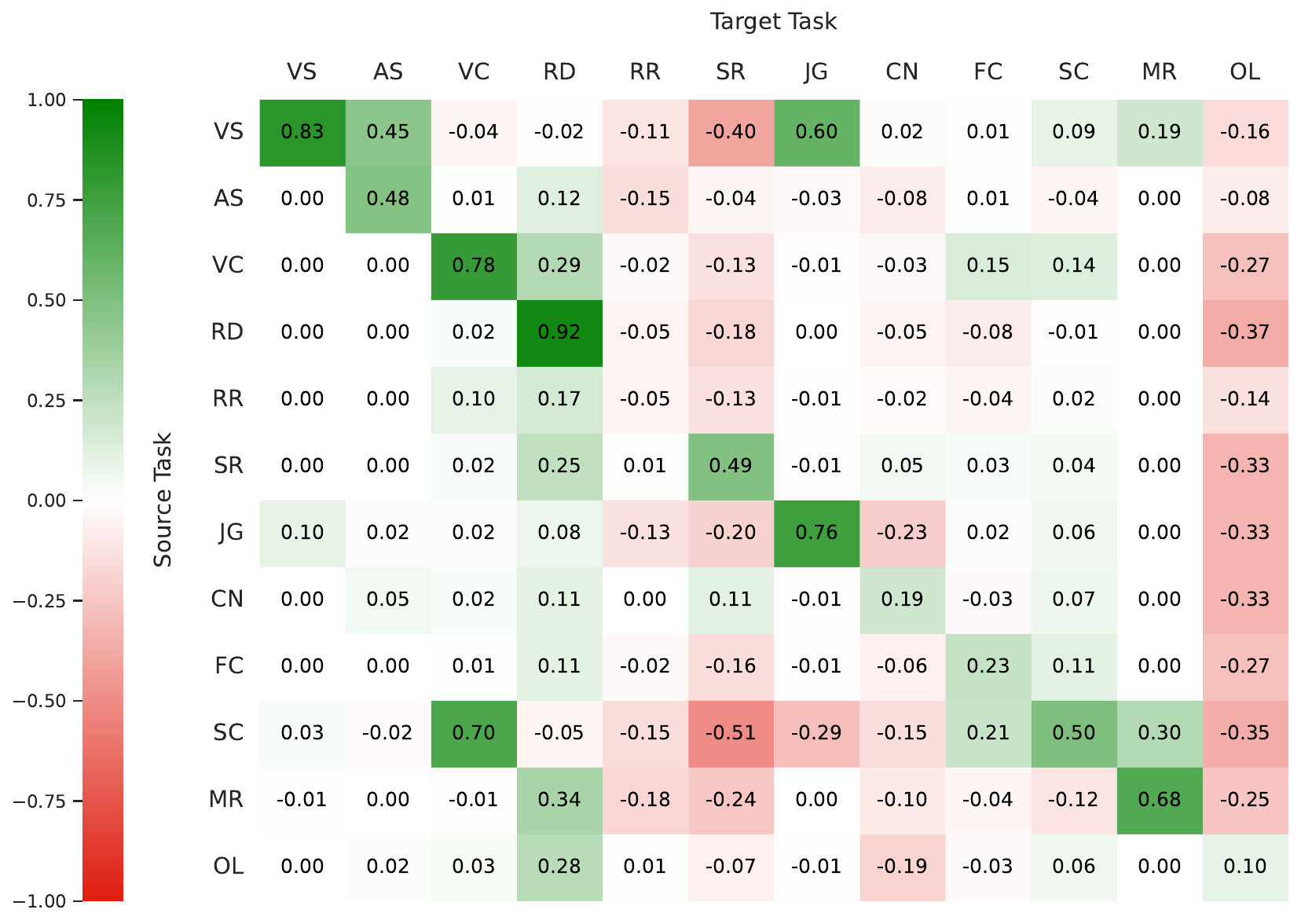}
    \caption{PGF Heatmap for the LLaVA V1.5 13B Model.}
    \label{fig:llava-heat}
\end{figure*}

We further assess whether the transfer patterns observed in Qwen2.5-VL models generalize to other VLM architectures by repeating our experiments on Llava1.5-13B. In Figure~\ref{fig:llava-heat}, we illustrate the PGF heatmap across BLINK tasks using the Llava1.5-13B model.  Qualitatively, we find that Visual Similarity, Art Style, and Jigsaw again form a coherent positive-transfer clique, aligning closely with the structure observed in Qwen2.5-VL. Likewise, Relative Depth consistently emerges as a sponge task, reinforcing its model-agnostic sensitivity to finetuning across architectures.

\section{Task Graph Visualizations}
\label{sec:task_graph}

\begin{figure*}[!htbp]
\centering
\begin{minipage}{0.24\linewidth}
    \centering
    \includegraphics[width=\linewidth]{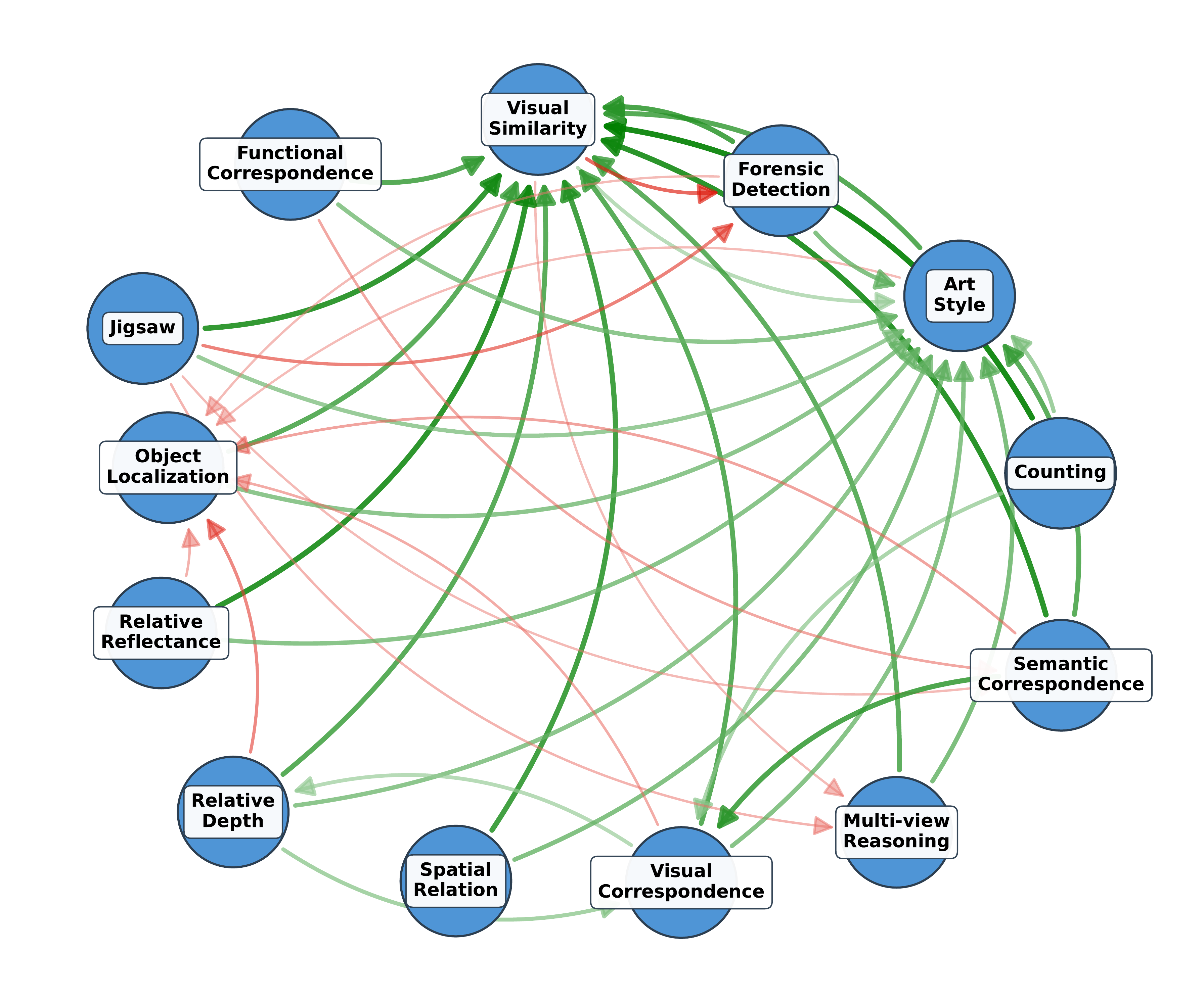}
    \caption*{(a) 25th percentile}
\end{minipage}
\hfill
\begin{minipage}{0.24\linewidth}
    \centering
    \includegraphics[width=\linewidth]{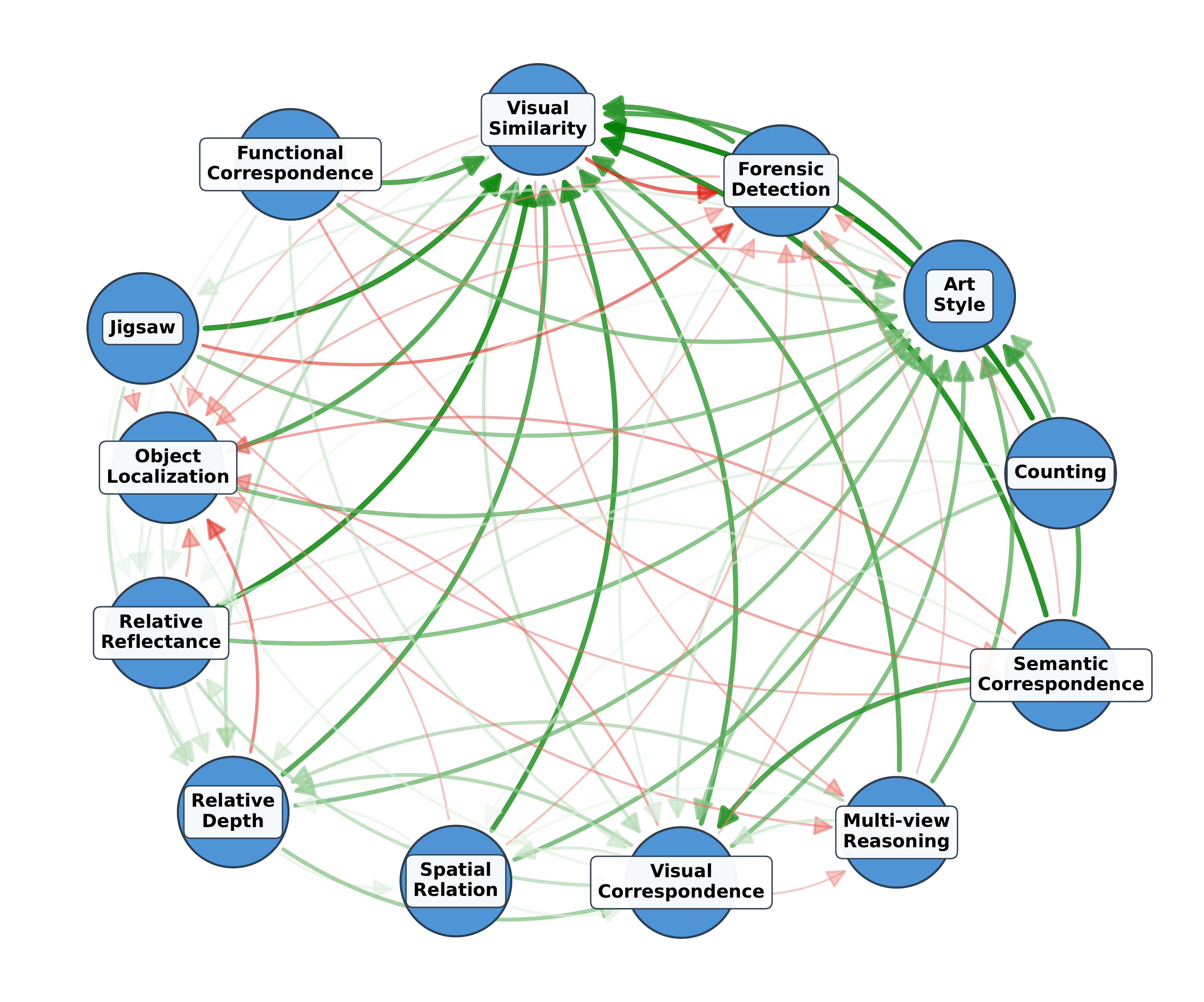}
    \caption*{(b) 50th percentile}
\end{minipage}
\hfill
\begin{minipage}{0.24\linewidth}
    \centering
    \includegraphics[width=\linewidth]{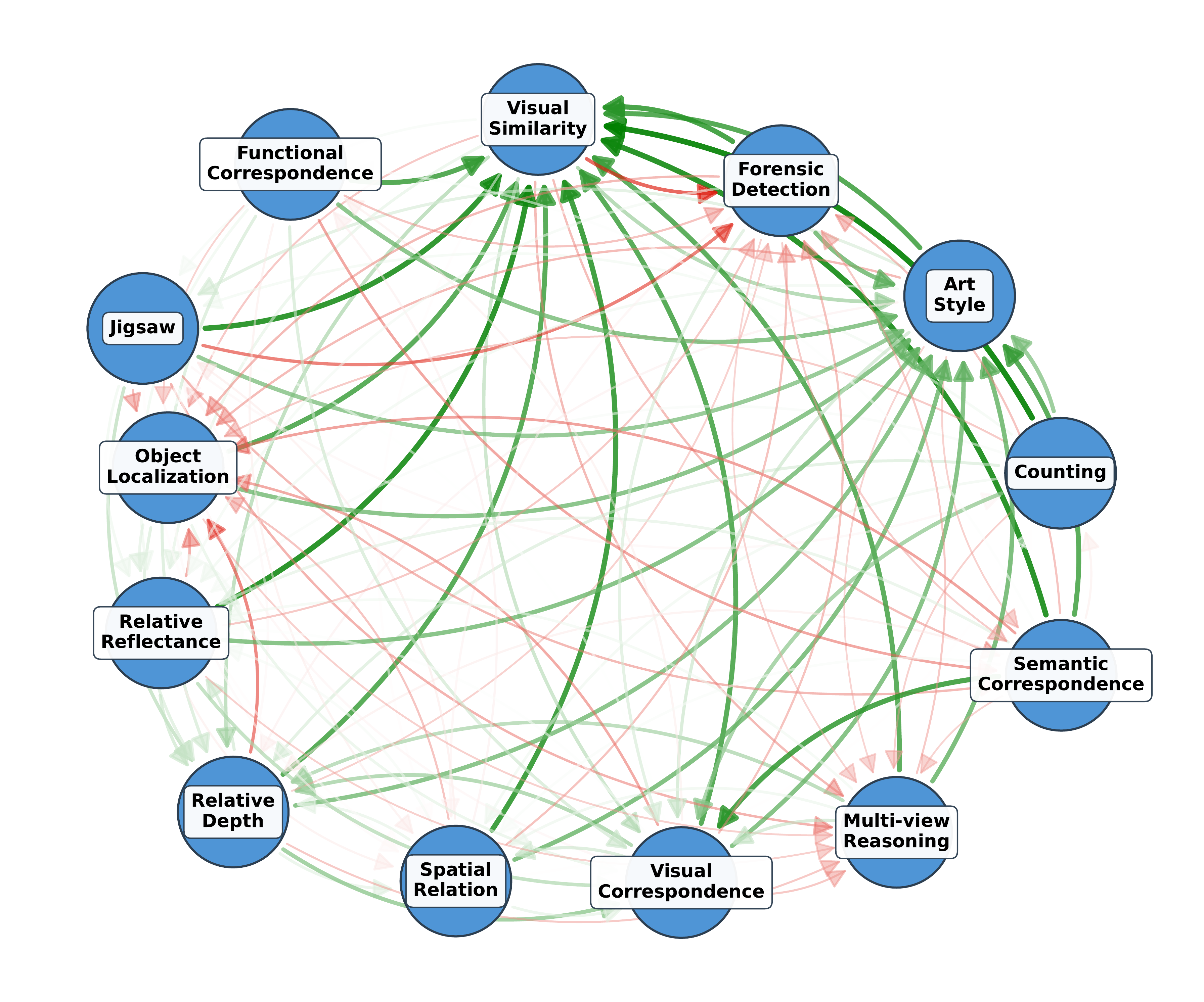}
    \caption*{(c) 75th percentile}
\end{minipage}
\hfill
\begin{minipage}{0.24\linewidth}
    \centering
    \includegraphics[width=\linewidth]{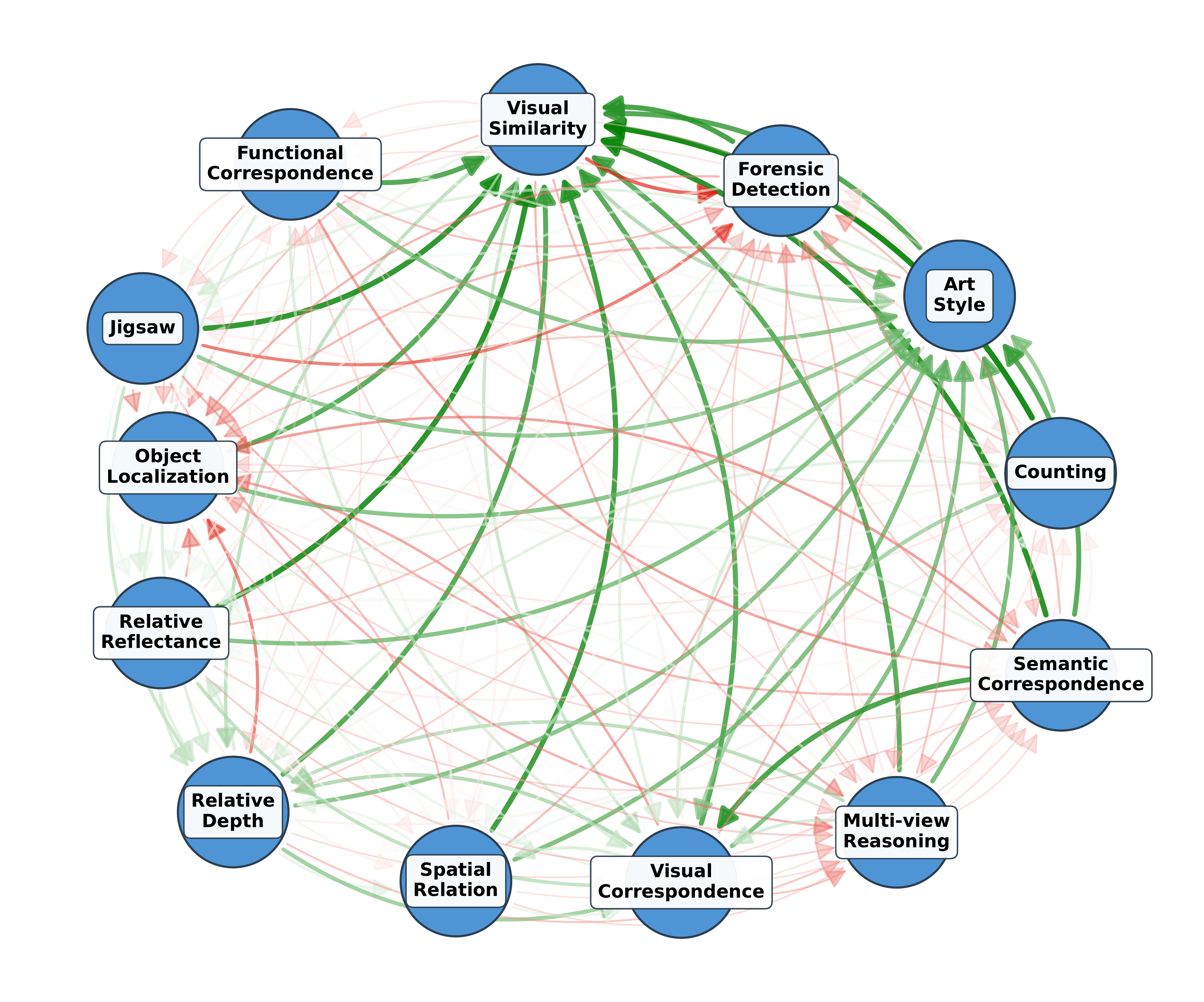}
    \caption*{(d) 100th percentile}
\end{minipage}
\caption{Visualization of Qwen-2.5-VL 32B task graph with varying percentile of edges shown.}
\label{fig:percent_graph}
\end{figure*}

We provide an ablation on the percentile of edges shown for visualization of the task graph in Figure~\ref{fig:percent_graph}. We ablate on the Qwen-2.5-VL 32B model and provide visualizations for 25th, 50th, 75th and 100th percentile of edges.

\section{PGF with Best Performance Ceiling}
\label{sec:best_perf}
\begin{figure*}[!htbp]
    \centering
    \includegraphics[width=0.82\linewidth]{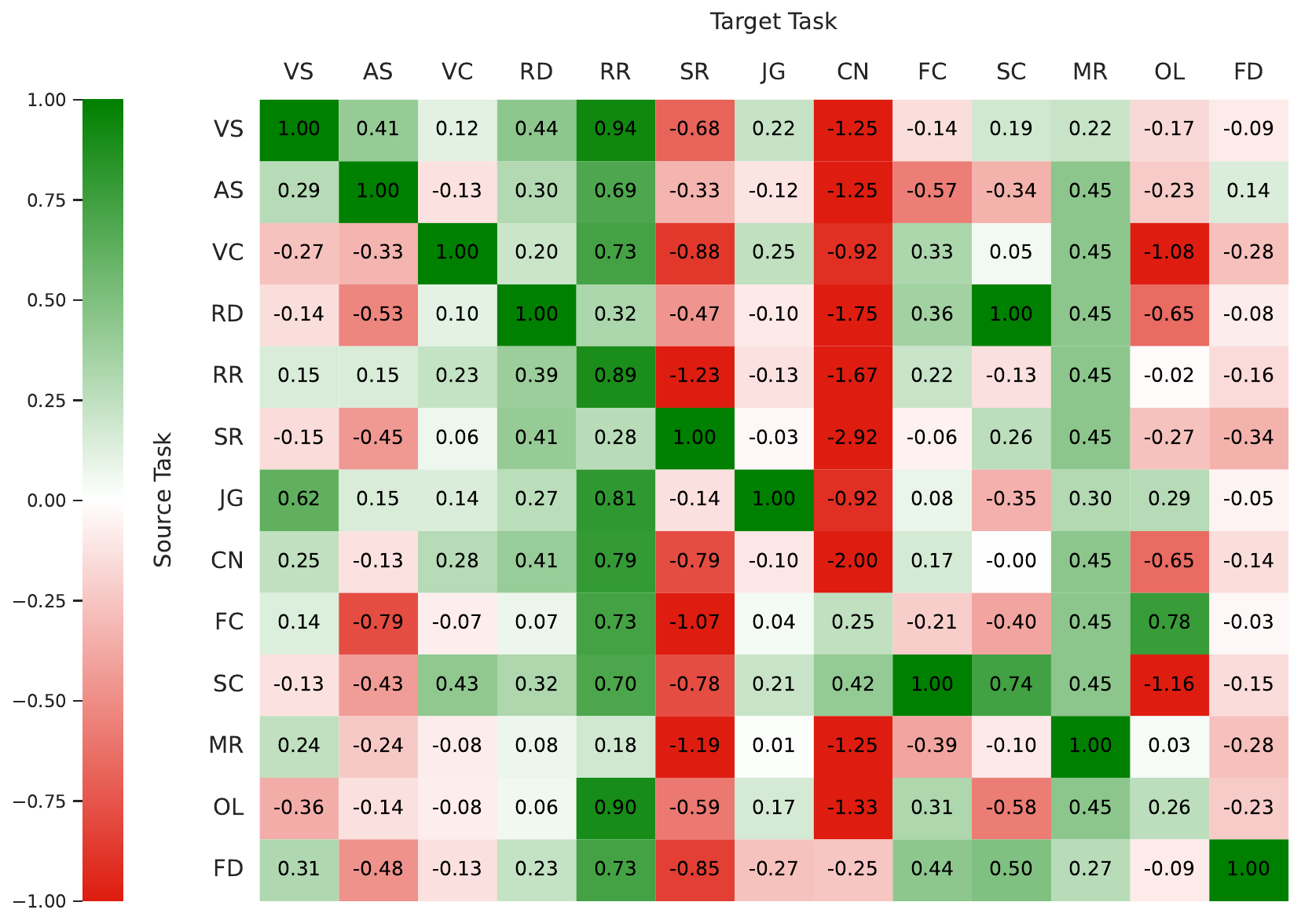}
    \caption{Best-bound PGF Heatmap for Qwen-2.5-VL 3B.}
    \label{fig:best_bound_pgf_3b}
\end{figure*}

\begin{figure*}[!htbp]
    \centering
    \includegraphics[width=0.82\linewidth]{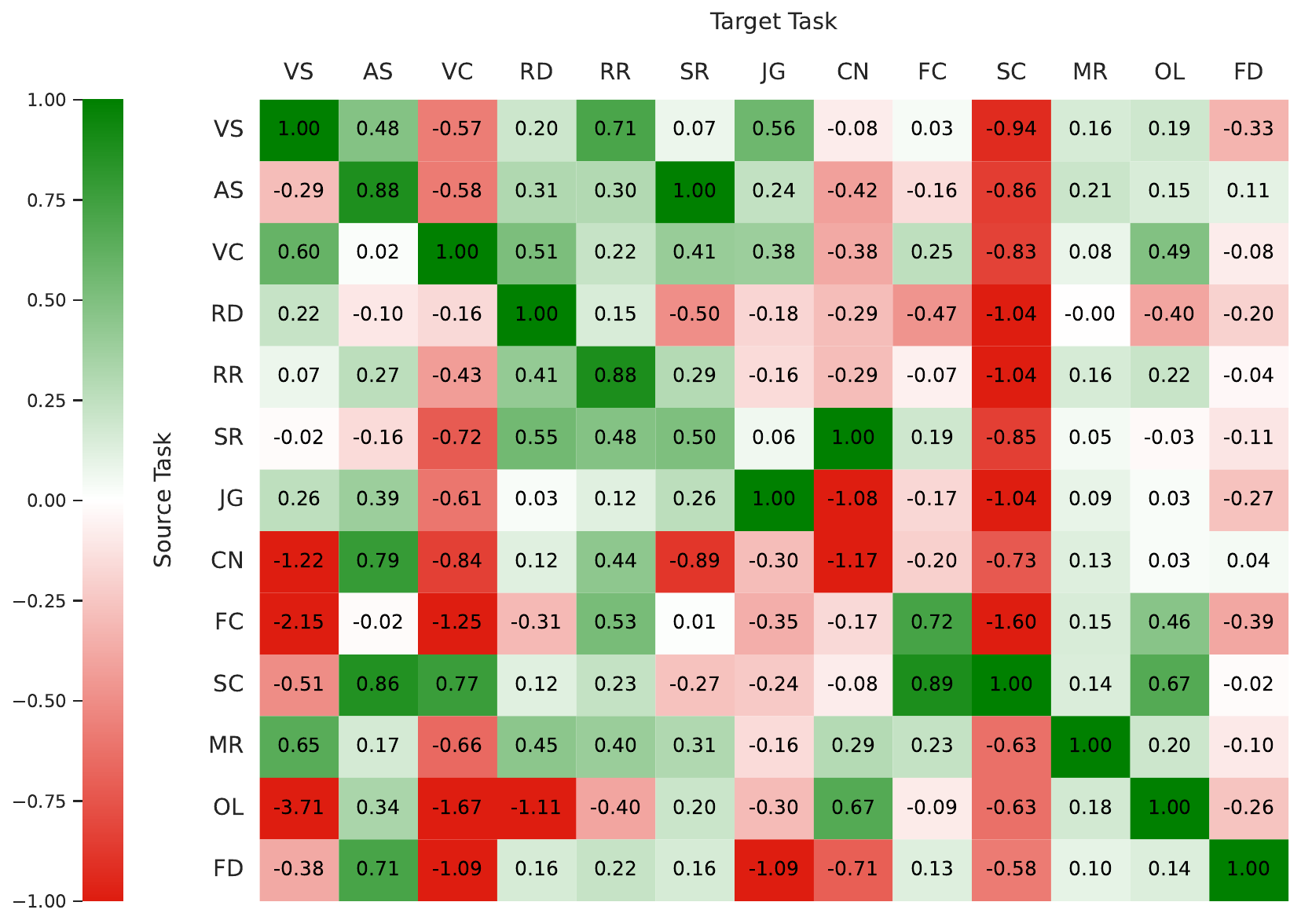}
    \caption{Best-bound PGF Heatmap for Qwen-2.5-VL 7B.}
    \label{fig:best_bound_pgf_7b}
\end{figure*}

\begin{figure*}[!htbp]
    \centering
    \includegraphics[width=0.82\linewidth]{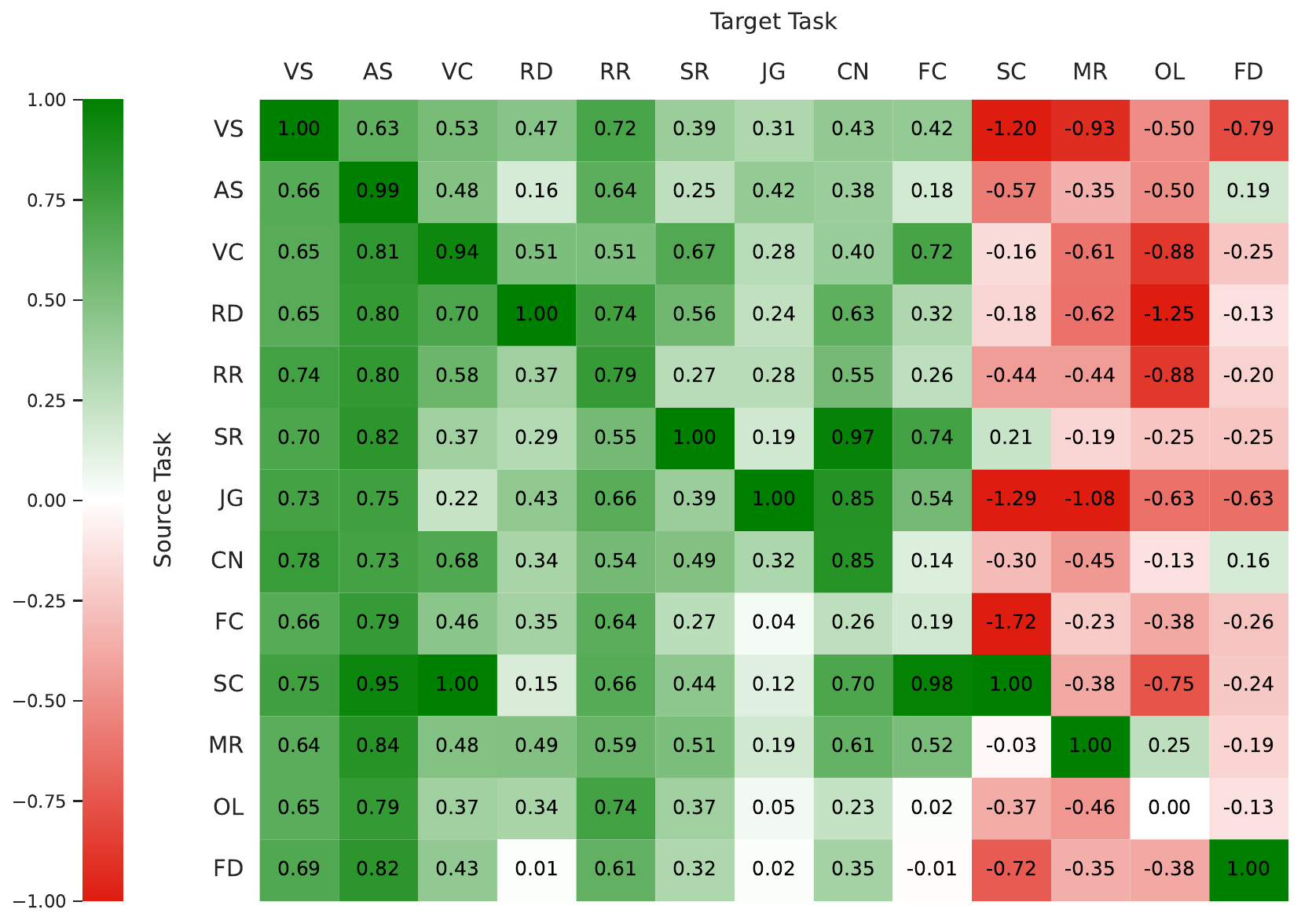}
    \caption{Best-bound PGF Heatmap for Qwen-2.5-VL 32B.}
    \label{fig:best_bound_pgf_32b}
\end{figure*}

To examine how the choice of ceiling $U$ influences PGF, we replace the original ceiling with the best observed performance on the target task. The resulting effects are demonstrated in Figure~\ref{fig:best_bound_pgf_3b}, Figure~\ref{fig:best_bound_pgf_7b}, and Figure~\ref{fig:best_bound_pgf_32b}. As expected, these plots exhibit a sequence of PGF scores equal to 1 along the diagonal, since direct supervision typically yields the highest performance.

% \section{Rationale}
% \label{sec:rationale}
% % 
% Having the supplementary compiled together with the main paper means that:
% % 
% \begin{itemize}
% \item The supplementary can back-reference sections of the main paper, for example, we can refer to \cref{sec:intro};
% \item The main paper can forward reference sub-sections within the supplementary explicitly (e.g. referring to a particular experiment); 
% \item When submitted to arXiv, the supplementary will already included at the end of the paper.
% \end{itemize}
% % 
% To split the supplementary pages from the main paper, you can use \href{https://support.apple.com/en-ca/guide/preview/prvw11793/mac#:~:text=Delete%20a%20page%20from%20a,or%20choose%20Edit%20%3E%20Delete).}{Preview (on macOS)}, \href{https://www.adobe.com/acrobat/how-to/delete-pages-from-pdf.html#:~:text=Choose%20%E2%80%9CTools%E2%80%9D%20%3E%20%E2%80%9COrganize,or%20pages%20from%20the%20file.}{Adobe Acrobat} (on all OSs), as well as \href{https://superuser.com/questions/517986/is-it-possible-to-delete-some-pages-of-a-pdf-document}{command line tools}.

\section{Broader Impact}
\label{sec:broader_impact}
Vision Language Models (VLMs) are increasingly being deployed in real-world systems like robotics, surveillance, autonomous vehicles, etc. Deploying VLMs in these critical domains requires a comprehensive understanding of the impact of finetuning on various tasks. Our findings demonstrate, for the first time, how finetuning on one task impacts performance across other tasks. This may help directly help practitioners design efficient and reliable finetuning pipelines. For example, identifying tasks that interfere with other tasks and ones that are highly transferable can reduce unexpected outcomes during deployment. Furthermore, PGF guided data selection can lower costs for users and democratize VLM finetuning. At the same time, our work highlights risks that arise from unintended transfer effects. Negative transfer between certain tasks indicates that naively finetuning VLMs for specialized capabilities can silently degrade other perception abilities, which may be consequential in safety-critical domains such as medical imaging or navigation. Although our benchmarks are standardized, real-world applications involve more diverse and noisy data distributions, where interference may be more severe. We encourage practitioners to apply transferability analyses before deploying VLMs in high-stakes settings. Overall, our analysis contributes to transparency by uncovering the patterns of task transfer. This underscrores the need for more comprehensive evaluation benchmarks for VLMs, ones that measure both performance and inter-task correlations. In future, we plan to extend this analysis to open-ended generation tasks, multiple languages, ensuring transfer behaviors generalize across diverse contexts.

% \clearpage
% {
%     \small
%     \bibliographystyle{ieeenat_fullname}
%     \bibliography{main}
% }

% \clearpage

% \begin{bibunit}[ieeenat_fullname]
% \input{sec/0_abstract}    
% \input{sec/1_intro}
% \input{sec/2_related_works}
% \input{sec/3_method}
% \input{sec/4_results}
% \input{sec/5_more_results}
% \input{sec/6_discussion}
% \input{sec/7_conclusion}

% \putbib[main] % This uses main.bib
% \end{bibunit}

% \clearpage

% \begin{bibunit}[ieeenat_fullname]

% \putbib[supp] % This uses supp.bib
% \end{bibunit}

\end{document}